\documentclass[runningheads]{llncs}

\usepackage{graphicx}
\usepackage{comment}
\usepackage{amsmath,amssymb}
\usepackage{color}
\usepackage{url}

\usepackage[dvipsnames]{xcolor}
\usepackage{cite}
\usepackage{wrapfig} 
\usepackage{booktabs}

\usepackage{enumitem}  
\usepackage[labelfont=bf]{caption}

\expandafter\def\expandafter\normalsize\expandafter{%
    \normalsize%
    \setlength\abovedisplayskip{0pt}%
    \setlength\belowdisplayskip{8pt}%
    \setlength\abovedisplayshortskip{-8pt}%
    \setlength\belowdisplayshortskip{2pt}%
}

\usepackage[labelfont=bf]{subcaption}
\usepackage{mathtools}
\usepackage[export]{adjustbox}
\usepackage{dsfont}
\usepackage{multirow}
\usepackage{amsmath}

\usepackage{colortbl}
\usepackage{tabulary}
\usepackage{etoolbox}
\usepackage{arydshln} 

\usepackage{tikz}
\usepackage{cancel}

\usepackage{amsmath,amsfonts,bm}

\def\eqref#1{equation~\ref{#1}}

\def\1{\bm{1}}

\DeclareMathAlphabet{\mathsfit}{\encodingdefault}{\sfdefault}{m}{sl}
\SetMathAlphabet{\mathsfit}{bold}{\encodingdefault}{\sfdefault}{bx}{n}

\newcommand{\std}[1]{\textcolor{gray}{$\pm$#1}}

\usepackage{hyperref}
\AtEndPreamble{
    \usepackage[capitalize]{cleveref}
    \crefname{section}{Sec.}{Secs.}
    \Crefname{section}{Section}{Sections}
    \crefname{table}{Tab.}{Tabs.}
    \Crefname{table}{Table}{Tables}
    \crefname{figure}{Fig.}{Figs.}
    \Crefname{figure}{Figure}{Figures}
    \crefname{appendix}{App.}{App.}
    \Crefname{appendix}{Appendix}{Appendix}
}

\newif\ifreview
\reviewfalse

\ifreview
	\usepackage{lineno}

	\linenumbers
\fi

\begin{document}


\def\SubNumber{30}

\def\GCPRTrack{Main Track}

\title{Examining Common Paradigms in \\Multi-Task Learning
\vspace{-0.2cm}
}

\ifreview
\else

    \author{Cathrin Elich\inst{\ddagger,1,2,3} \and
	Lukas Kirchdorfer\inst{\ddagger,1,4} \and
	\\ Jan M. Köhler\inst{*,1} \and
    Lukas Schott\inst{*,1}}
	
	\authorrunning{C. Elich et al.}
	
     \institute{
        Bosch Center for Artificial Intelligence \and
        Max Planck Institute for Intelligent Systems, Tübingen, Germany \and
        Max Planck ETH Center for Learning Systems \and
        University of Mannheim \\
        {\tt\small cathrin.elich@tuebingen.mpg.de, \tt\small \{jan.koehler,lukas.schott\}@bosch.com} \\
        { \small  $^\ddagger$Work done during an internship at Bosch.\; \; \; $^*$Joint senior authors.}
    }
\fi

\maketitle              

\begin{abstract}
While multi-task learning (MTL) has gained significant attention in recent years, its underlying mechanisms remain poorly understood. 
Recent methods did not yield consistent performance improvements over single task learning (STL) baselines, underscoring the importance of gaining more profound insights about challenges specific to MTL.
In our study, we investigate paradigms in MTL in the context of  STL:
First, the impact of the choice of optimizer has only been mildly investigated in MTL. 
We show the pivotal role of common STL tools such as the Adam optimizer in MTL empirically in various experiments.
To further investigate Adam's effectiveness, we theoretical derive a partial loss-scale invariance under mild assumptions. 
Second, the notion of gradient conflicts has often been phrased as a specific problem in MTL. 
We explore the role of gradient conflicts in MTL and compare it to STL. For angular gradient alignment we find no evidence that this is a unique problem in MTL. We emphasize differences in gradient magnitude as the main distinguishing factor.  
Overall, we find surprising similarities between STL and MTL suggesting to consider methods from both fields in a broader context. 
  \keywords{Multi-task learning \and Deep Learning \and Computer Vision}
\end{abstract}
%
%
%

\section{Introduction} 
\label{sec:intro}

Multi-task learning (MTL) is gaining significance in the deep learning literature and in industry applications. Especially, tasks like autonomous driving and robotics necessitate real-time execution of neural networks while obeying constraints of limited computational resources. Consequently, there is a demand for neural networks capable of simultaneously inferring multiple tasks \cite{ishihara2021multi,lee2021fast}.

In a seminal study, Caruana \cite{caruana1997multitask} highlights both advantages and challenges in MTL. 
On the one hand, certain tasks can exhibit a symbiotic relationship, resulting in a mutual performance enhancement when trained together. 
On the other hand, conflicts between tasks can arise and decrease the performance when trained jointly, also known as \textit{negative transfer}.

Several approaches have been suggested to mitigate the issue of negative transfer among tasks during network training. 
Our study focuses on two main branches in the literature: 
First, \emph{gradient magnitude} methods which incorporate weights to scale task-specific losses to achieve an adequate balance between tasks. 
Second, \emph{gradient alignment} methods which aim to resolve conflicts in gradient vectors that may arise between tasks within a shared network backbone.

The effectiveness of the proposed MTL methods remain uncertain in the literature. Upon comparing various studies, it becomes evident that there is no definitive approach that consistently performs well across different settings \cite{vandenhende_survey_2021}. 
This observation has been reinforced in more recent studies where competitive performance was achieved through plain unitary scaling in combination with common regularization methods~\cite{kurin_unitary-scalar_2022} or tuned task weighting~\cite{xin_mto-even-help_2022}.

The current understanding of MTL still lacks a deeper comprehension of its underlying mechanisms. 
To address this gap, our study aims to examine commonly held paradigms, such as the choice of optimizer, as well as the notion of gradient alignment and gradient magnitudes. 
Our \textbf{contributions} are:
\begin{itemize}[noitemsep,topsep=0pt,leftmargin=10pt]
    \item The impact of off-the-shelf optimizers has received little attention in MTL benchmarks. We evaluate the Adam~\cite{kingma__adam__2015} optimizer and demonstrate its favorable performance over  SGD+momentum in various experiments.
    \item We provide a potential explanation for Adam's effectiveness in MTL by theoretically demonstrating a partial invariance w.r.t.\ to different loss scalings. 
    Similarly, we derive a full invariance for an optimal variation of the well-established used method of uncertainty weighting \cite{kendall_uw_2018}. 
    \item So far gradient \emph{alignment} conflicts have mostly been considered between different tasks \cite{yu_pcgrad_2020, liu_cagrad_2021, javaloy_rotograd_2022, chen_graddrop_2020}. We present empirical evidence that conflicts arising from gradient alignment between tasks are not exclusive and can even be more pronounced between different samples within a task. 
    \item Corroborating the methods proposed to balance gradient \emph{magnitude} conflicts in MTL \cite{kendall_uw_2018, liu_mtan_2019, liu_imtl_2021, xin_mto-even-help_2022}, we confirm that gradient magnitudes pose a challenge between tasks and is less pronounced between samples within a task. 
    \item We examine the presumption of increased robustness on corrupted data as a result of MTL \cite{klingner2020improved,mao2020multitask}. We find light evidence that a higher number of tasks can result in improved transferability.
    Due to page limitations, we moved these results to \cref{sec:app_ood}, focusing the more compelling findings in the main text.
\end{itemize}
Overall, we provide a vast set of experiments and theoretical insights which contribute to a more comprehensive understanding of MTL in computer vision.

\section{Related Work}
\label{sec:rel_work}

Work in multi-task learning (MTL) can be roughly divided into three fields: 

\textit{Network architectures} focus on the question of how features should be shared across tasks, e.g. \cite{misra_crossstitch_2016, liu_mtan_2019, xu_padnet_2018, maninis_astmt_2019}.
\textit{Multi-task optimization (MTO)} aims to resolve imbalances and conflicts of tasks during MTL.
\textit{Task affinities} examine a grouping of tasks that should be learned together to benefit from the joint training \cite{standley_whichtasks_2020, fifty_taskgrouping_2021}.
A general overview of recent works in MTL can be found in \cite{vandenhende_survey_2021, ruder2017overview}.
Our work focuses on MTO, which we review more thoroughly in the following.

\textbf{\textit{Gradient magnitude methods}} 
prevent the dominance of individual tasks by balancing them with task-specific weights.
One line of works are loss-weighting methods.
Here, weights are determined before any (task-wise) gradient computation and are used for a weighted aggregation of the tasks' losses.
These methods consider either the task uncertainty (UW)~\cite{kendall_uw_2018}, rate of change of the losses (DWA, FAMO)~\cite{liu_mtan_2019,bofamo2023}, the tasks' difficulty (DTP)~\cite{guo_gtp_18}, validation performance by applying meta-learning (MOML, Auto-$\lambda$) \cite{ye21moml, liu_auto-lambda_2022}, or randomly chosen task weights (RLW)~\cite{lin_rlw_2022}.
In line with these, the geometric mean of task losses has been used to handle the different convergence rates of the tasks \cite{chennupati_gls_2019}. 
An advantage of theses methods is their computational efficiency as the gradient needs to be computed only once for the aggregated loss.
Alternatively, other methods consider the task-specific gradients directly, e.g., by normalizing them (GradNorm)~\cite{chen_gradnorm_18} 
 or propose a hybrid balancing between task-wise loss and gradient scaling (IMTL, DB-MTL)~\cite{liu_imtl_2021,lin2023dualbalancing}.
Furthermore, there are several adaptions for the multiple-gradient descent algorithm (MGDA)~\cite{desideri_mgda_2012}, e.g. for applying it efficiently in deep learning setups~\cite{sener_mgda_2018} or by introducing a stochastic gradient correction~\cite{fernando_moco_2023}.
Recently, task-wise gradient weights have been estimated by treating MTL as a bargaining problem (Nash-MTL)~\cite{navon_nashmtl_22}, or considering a stability criterion (Aligned-MTL)~\cite{senushkin_alignedmtl_2023}.
Crucially, all \emph{gradient magnitude} methods consider \textit{scalar} weightings of task-wise gradients within the backbone and/or heads. They do not modify the alignment of task-specific gradient vectors.

\textbf{\textit{Gradient alignment methods}} 
perform more profound vector manipulations on the task-wise gradients w.r.t.\ to the network weights of a shared backbone before aggregating them.
The underlying assumption indicates conflicting gradients as a major problem in MTL.
To address this, GradDrop \cite{chen_graddrop_2020} randomly drops gradient components in the case of opposing signs. 
PCGrad \cite{yu_pcgrad_2020} proposes to circumvent problems of conflicting gradients by projecting them onto each other's normal plane.
Following this idea, Liu et al.~\cite{liu_cagrad_2021} propose CAGrad to converge to a minimum of the average loss instead of any point on the Pareto front. 
RotoGrad \cite{javaloy_rotograd_2022} rotates gradients at the intersection of the heads and backbone to improve their alignment.
Shi et al.~\cite{shi_recon_2023} propose to alter the network architecture based on the occurrence of layer-wise gradient conflicts.
Lastly, \cite{pascal2021improved} use separate optimizers such as SGD and SGD+momentum per task. This is extended to AdaGrad, RMSProp and Adam in AdaTask \cite{AdaTask_AAAI2023}. 

Recent studies \textbf{\textit{question the effectiveness of optimization-based methods}} in MTL. 
Xin et al.~\cite{xin_mto-even-help_2022} execute an extensive hyperparameter search to show that simple scalar task-weighting performs equivalent or superior to many aforementioned multi-task optimization methods. 
Their hyperparameter search not only include the task-weights, but also common deep learning parameters such as the learning rate and regularization. 
Concurrently, Kurin et al.~\cite{kurin_unitary-scalar_2022} empirically show that fixed task-weights combined with regularization and stabilization techniques yield to equivalent performance compared to sophisticated multi-task optimization methods.
Following these, Royer et al.~\cite{royer2023scalarization} examine the role of model capacity for MTL performance as well as the occurrences of gradient conflicts.
We extend these critical studies. 
In particular, we theoretically and empirically demonstrate that the choice of optimizer is crucial and could potentially help to explain discrepancies found in prior studies (\ref{sec:exp_adam_sgd}).
We further specifically distinguish between gradient conflicts between tasks and samples (\ref{sec:exp_grad_conflicts}).

\section{Problem Statement}
\label{sec:preliminaries}

Multi-task learning addresses the problem of learning a set of $T$ tasks simultaneously (see e.g.~\cite{vandenhende_survey_2021,caruana1997multitask}).
It is noteworthy that this setup is occasionally also referred to as \textit{multi-label} or \textit{multi-target learning}~\cite{zhang_22_mtlSurvey}
Importantly, this study does not incorporate multi-input data. 
We consider a supervised learning setup, use a shared backbone architecture, and learn all tasks together.
Formally, given input data $\mathcal{X}$, the goal is to learn a function $f_{\boldsymbol{\theta}}(\mathbf{x})$ which maps a point $\mathbf{x} \in \mathcal{X}$ to each task label $y_t$ with $t=1, ..,T$.
The trainable parameters $\boldsymbol{\theta}=\{\phi, \psi_{1:T}\}$ consist of \textit{shared} parameters $\phi$ and \textit{task-specific} parameters $\psi_t$. 
Training a task $t$ is associated with the loss $\mathcal{L}_t(f_{\boldsymbol{\theta}}(x);\boldsymbol{\theta})$, e.g., a regression or classification loss. We denote respective gradients on the shared and task-specific parameters with $\mathbf{g}^{\phi}_t=\nabla_{\phi}\mathcal{L}_t$, and $\mathbf{g}^{\psi}_t=\nabla_{\psi}\mathcal{L}_t$.
%
When training on multiple tasks, the shared parameters $\phi$ needs to be updated w.r.t.\ all task-wise gradients $\mathbf{g}^{\phi}_t$ which requires an appropriate aggregation.
A simple solution is to uniformly sum up the task losses $\mathcal{L} = \sum_t \mathcal{L}_t$ which is referred to as \textit{Equal Weighting} (EW).
However, as tasks might be competing against each other, this can result in negative transfer and thus sub-optimal solutions.
One way to deal with this difficulty is to adapt the \textit{magnitude} of task-specific gradients.
This can be achieved by weighting tasks during training, e.g., by scaling different losses $\mathcal{L} = \sum_t \alpha_t \mathcal{L}_t$, where  $\alpha_t \geq 0$. 
Note that the $\alpha_t$ can change during training. Furthermore, the weighing can also be performed on gradient level to distinguish between shared and task-specific gradients. We refer to those approaches as \emph{gradient magnitude} methods. 
Interestingly, the relationship between loss weights, network updates and learning rate also depends on the optimizer. We show a derivation for SGD and Adam in \Cref{sec:appx_lr_task_weight}.
Additionally to adapting the gradient magnitude, one can directly adapt the \textit{alignment} of task-wise gradient vectors within the shared backbone 
$\tilde{\mathbf{g}}^{\phi}=\mathbf{h}(\mathbf{g_1}^{\phi}, ..., \mathbf{g_T}^{\phi})$.

In practice, an optimum for $\boldsymbol{\theta}$ that yields best performance on all tasks often does not exist. 
Instead, improving performance on some task often yields a performance decrease in another task.
To still enable a comparison across network instances in MTL, an instance $\boldsymbol{\theta}^\ast$ is called to be \textit{Pareto optimal}, if there is no other $\boldsymbol{\theta}'$ such that $ \mathcal{L}_t(\boldsymbol{\theta}') \leq \mathcal{L}_t(\boldsymbol{\theta}^\ast) \; \forall t$ with strict inequality in at least one task. The \textit{Pareto front} consists of the Pareto optimal solutions.

\section{Experiments and results}

\label{sec:exp}

In this section we perform several experiments to gain a more profound understanding of multi-task learning (MTL) in computer vision by questioning common paradigms.
We compare the impact of Adam and SGD in MTL in \cref{sec:exp_adam_sgd} and examine the process of gradient similarity in different settings in \cref{sec:exp_grad_conflicts}.
Throughout this evaluation, we repeatedly make use of common setups, which we will specify as follows and in more detail in \cref{sec:appx_impl_details}.

\textbf{Datasets:}~
For our experiments, we consider three different datasets that are commonly used for evaluating MTL in computer vision: 
\textit{Cityscapes}~\cite{cordts_cityscapes_2016} contains images of urban street scenes.
In line with previous work, we consider the tasks of semantic segmentation (7 classes) and depth estimation.
\textit{NYUv2}~\cite{silberman_nyuv2_2012} is an indoor dataset for scene understanding which was recorded over 464 different scenes across three different cities.
Besides semantic segmentation (13-class) and depth estimation, it also contains the task of surface normal prediction.
\textit{CelebA}~\cite{liu_celeba_2015} consists of 200K face images which are labeled with 40 binary attributes.

\textbf{Networks:}~
We use network architectures with hard-parameter sharing which consist of a shared backbone and task-specific heads. 
For the dense prediction tasks on Cityscapes and NYUv2, we compare SegNet\cite{badrinarayanan_segnet_2017} and DeepLabV3+\cite{chen_deeplabv3_2018}.
Experiments on CelebA are performed on a ResNet-18~ \cite{he_resnet_16} with an additional single linear layer for each head.

\textbf{Training:}~
For each method, we follow the loss or gradient aggregation as described in the related work, e.g., for equal weighting all task-specific losses are simply summed up to compute the joint network gradients. 
The learning rate is tuned separately for each approach.
We use the validation set performance of the $\Delta_m$~metric as early stopping criteria.
The $\Delta_m$~metric~\cite{maninis_astmt_2019} measures the average relative task performance drop of a method $m$ compared to the single-task baseline $b$ using the same backbone and is computed as
$
    \Delta_m = \frac{1}{T}\sum_{t=1}^T (-1)^{l_t}(M_{m, t}-M_{b, t})/M_{b, t}
$
where $l_t =1$ if a higher value means better for measure $M_{\cdot,t}$ of some task metric $t$, and 0 otherwise.

\subsection{Effectiveness of Adam in multi-task learning}
\label{sec:exp_adam_sgd}
\textit{Examined paradigm:} The impact of the choice of standard optimizer is often disregarded and varies across studies (overview in \cref{tab:rel_work}) when comparing MTL methods. 
For instance, Adam \cite{kingma__adam__2015} was successfully used to show that random/constant weighting of tasks' losses performs competitive compared to MTO methods \cite{lin_rlw_2022, kurin_unitary-scalar_2022, xin_mto-even-help_2022}. 
In contrast, many methods proposing adaptive, task-specific weighting methods \cite{liu_imtl_2021,kendall_uw_2018} use stochastic gradient descent with momentum (SGD+mom).
In recent works, the optimizer choice converged to Adam and a fixed learning rate schedule \cite{liu_mtan_2019, yu_pcgrad_2020, liu_cagrad_2021, senushkin_alignedmtl_2023} without a comparison to SGD+mom.

In this part of our study, we investigate the impact of Adam and SGD+mom in conjunction with common MTO methods. We identify the choice of optimizer as a crucial confounder in the experimental setup.
Compared to SGD+mom, we find that the Adam optimizer itself is a quite effective baseline in MTL and can be regarded as a loss weighting method from a theoretical viewpoint.

\subsubsection{Toy Task Experiment}

To get a first impression of the impact of the optimizer and common hyperparameters such as the learning rate, we investigate the impact of Adam and plain gradient descent (GD) in a simple toy task.  

\begin{figure}[tb]
\centering
\scriptsize
\newcommand{\gcs}{\hspace{4pt}}
\newcommand{\vgcs}{\hspace{8pt}}
\begin{tabular}{p{1.6cm}c@{\gcs}ccc c@{\vgcs} cc@{\gcs}cccp{1.cm}}
     && \multicolumn{3}{c}{\textbf{GD}} && && \multicolumn{3}{c}{\textbf{Adam}}\\
     \cmidrule(){3-5} \cmidrule(){9-11} 
    learning rate && 1.0 & 0.05 & 0.001 && && 1.0 & 0.05 & 0.001 \\[-0.75ex]
    ~ \newline ~ \newline EW &&
    \includegraphics[trim = 15mm 2mm 22mm 2mm, clip, valign=t, width=0.1\textwidth]{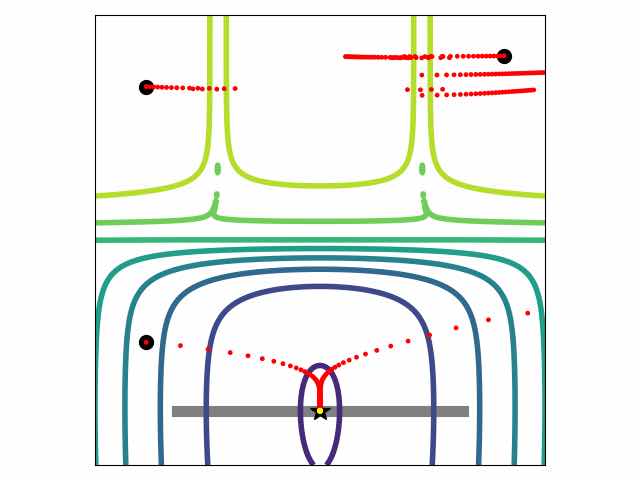} & 
    \includegraphics[trim = 22mm 2mm 22mm 2mm, clip, valign=t, width=0.1\textwidth]{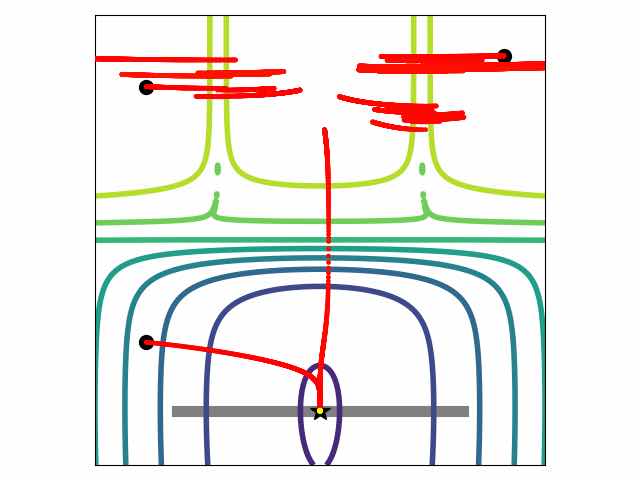} & 
    \includegraphics[trim = 22mm 2mm 22mm 2mm, clip, valign=t, width=0.1\textwidth]{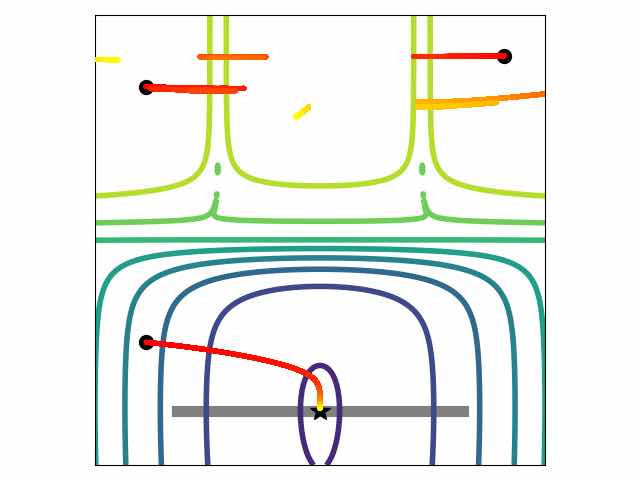} 
    && &&
    \includegraphics[trim = 22mm 2mm 22mm 2mm, clip, valign=t, width=0.1\textwidth]{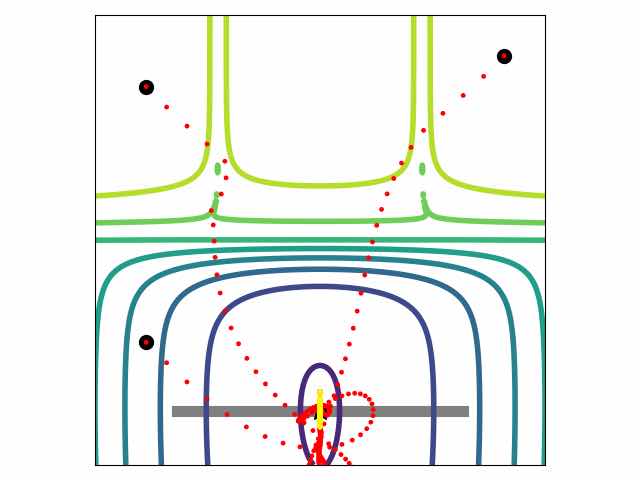} & 
    \includegraphics[trim = 22mm 2mm 22mm 2mm, clip, valign=t, width=0.1\textwidth]{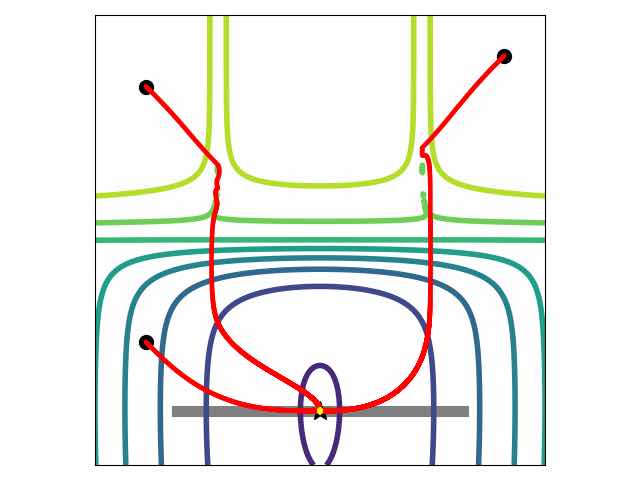} & 
    \includegraphics[trim = 22mm 2mm 22mm 2mm, clip, valign=t, width=0.1\textwidth]{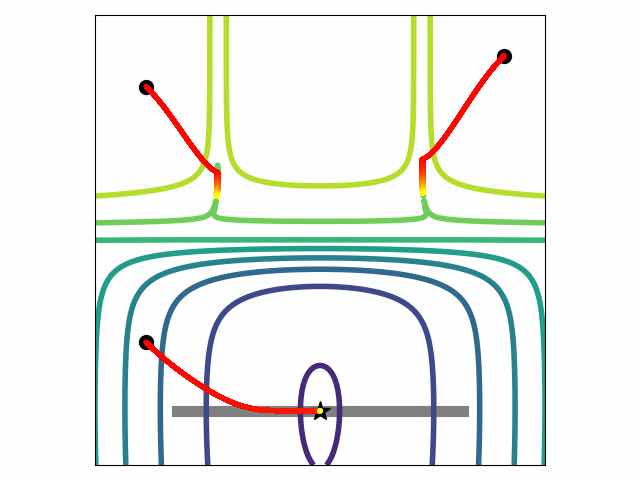} &
    \rotatebox{90}{\hspace{-60pt}\color{Orchid}\textit{setup from \cite{liu_cagrad_2021}}}
    \\
    ~ \newline ~ \newline CAGrad &&  
    \includegraphics[trim = 15mm 2mm 22mm 2mm, clip, valign=t, width=0.112\textwidth]{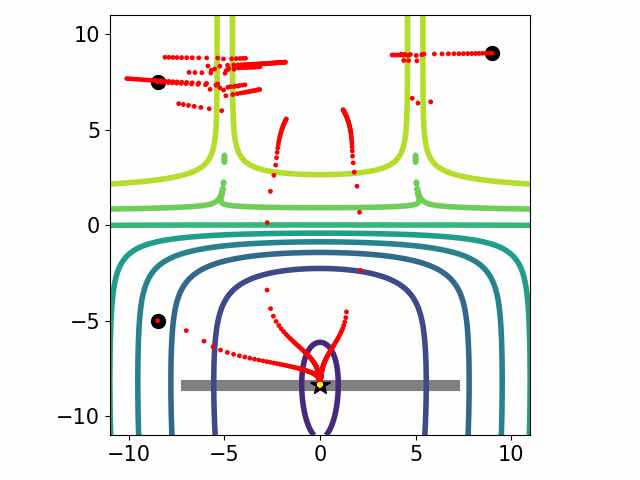} & 
    \includegraphics[trim = 22mm 2mm 22mm 2mm, clip, valign=t, width=0.1\textwidth]{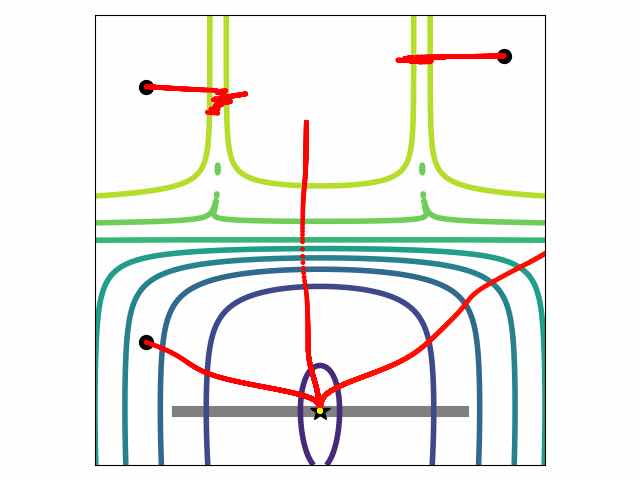} & 
    \includegraphics[trim = 22mm 2mm 22mm 2mm, clip, valign=t, width=0.1\textwidth]{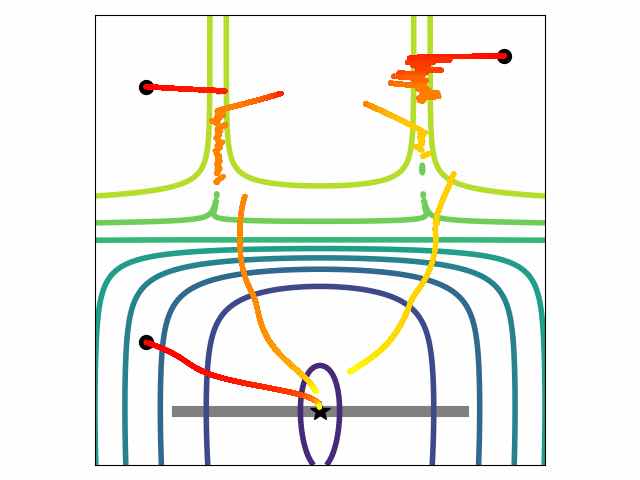} 
    && &&
    \includegraphics[trim = 22mm 2mm 22mm 2mm, clip, valign=t, width=0.1\textwidth]{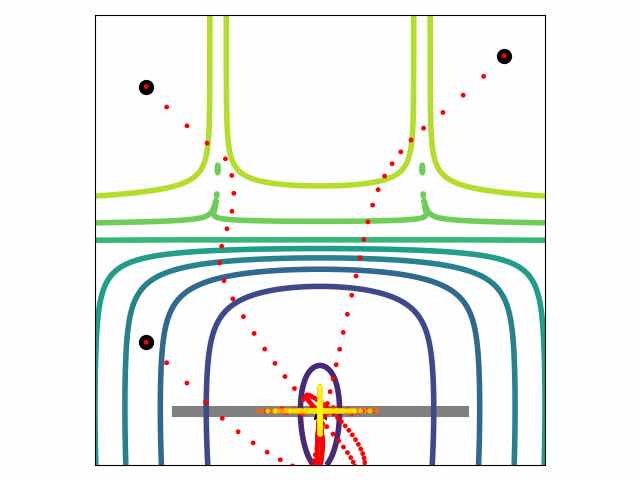} & 
    \includegraphics[trim = 22mm 2mm 22mm 2mm, clip, valign=t, width=0.1\textwidth]{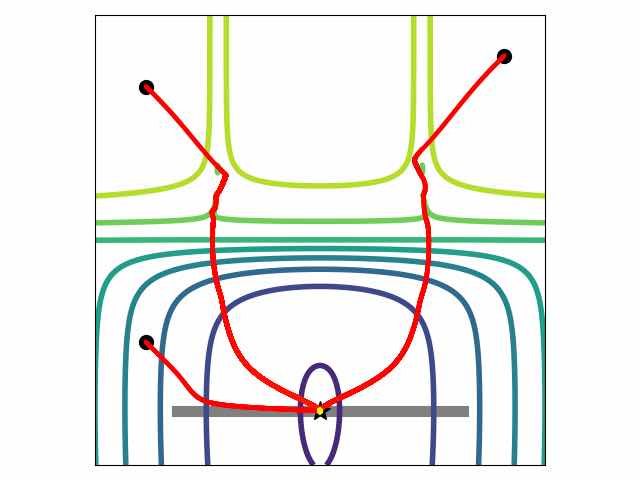} & 
    \includegraphics[trim = 22mm 2mm 22mm 2mm, clip, valign=t, width=0.1\textwidth]{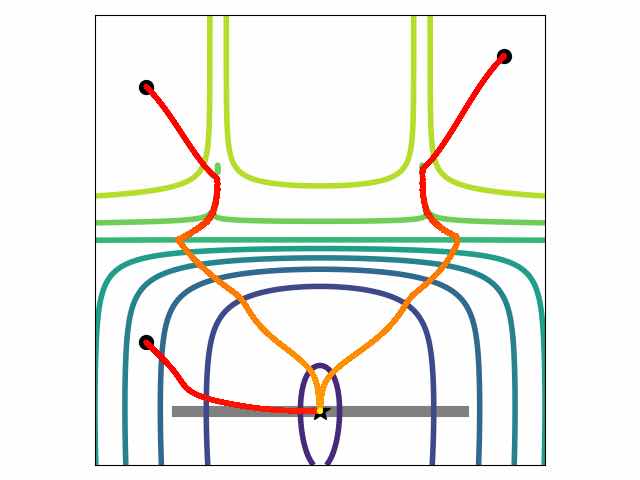} \\
    \begin{tikzpicture}[overlay]
      \draw[Orchid, ultra thick] (9.1,0.2) rectangle (10.35,2.75);
     \end{tikzpicture}
     \\[-2.75ex]
\end{tabular}
\caption{\textbf{Toy task experiment from CAGrad~\cite{liu_cagrad_2021} for different learning rates and optimizers.}
Consistent with results from \cite{xin_mto-even-help_2022}, we observe that the choice of the learning rate is crucial even for this toy optimization problem.
Moreover, it becomes apparent, that selecting Adam  over simple gradient decent (GD) yields superior results. 
The contour lines depict the 2D loss landscape; the optimization trajectories are colored from {\color{red}red} to {\color{Dandelion}yellow} for 100k iteration steps from three different starting points (seeds).
} %
\label{fig:cagrad_toytask}
\end{figure}
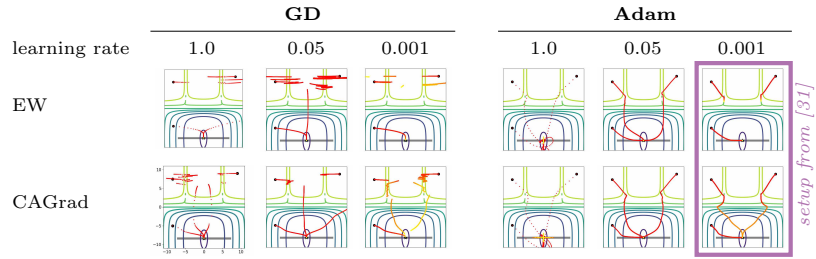

\textit{Approach:} We repeat the experiment of  Liu et al.~\cite{liu_cagrad_2021} using their original implementation but further test different learning rates and optimizers.
They motivate their gradient alignment method CAGrad with a simple toy optimization problem in which their method reliably converges to the minimum of the average loss, while other MTO approaches would either get stuck (e.g., EW) or only converge to any point on the Pareto front (e.g., PCGrad~\cite{yu_pcgrad_2020}, MGDA~\cite{sener_mgda_2018}).

\begin{table}[t]
\vspace{-.5cm}
\centering
\scriptsize
\setlength{\tabcolsep}{2pt}  
\newcommand{\gcs}{\hspace{6pt}}
\newcommand{\dgcs}{\hspace{16pt}}
\caption{
Maximum number of iterations for all seeds in the toy task experiment from~\cite{liu_cagrad_2021} to reach the global minimum 
for varying MTO method, learning rate, and optimizer combination. 
EW+Adam often shows the fastest convergence to the global minimum.
'-' denotes that not all seeds converged within 100k iterations.
As reported in \cite{liu_cagrad_2021}, PCGrad often only converges to a point on the Pareto Front. 
We highlight the \textbf{best} run for each learning rate over all MTO methods. 
}
\begin{tabular}{cc@{\gcs}cc@{\gcs}ccccc c@{\dgcs} cc@{\gcs}cc@{\gcs}ccccc}
    \toprule
   &&&& \multicolumn{5}{c}{learning rate} && && && \multicolumn{5}{c}{learning rate}\\
   \cmidrule{5-9}\cmidrule{15-19}
    && method && 10.0 & 1.0 & 0.1 & 0.01 & 0.001* && && method && 10.0 & 1.0 & 0.1 & 0.01 & 0.001*  \\
    \cmidrule{1-9}\cmidrule{11-19}
    {\multirow{3}{*}{\rotatebox{90}{\hspace{-5pt} GD}}} && EW && - & - & - & - & - 
    &&{\multirow{3}{*}{\rotatebox{90}{\hspace{-5pt}\vspace{-15pt} Adam}}} && EW && 26 & \textbf{22} & \textbf{709} & \textbf{9,015} & - \\
    && PCGrad && - & - & - & - & - 
    && && PCGrad &&  \textbf{25} & 56 & 34,175 & - & - \\
    && CAGrad && 644 & 213 & 8,069 & 20,418 & - 
    && && CAGrad && 27 & 32 & 802 & 11,239 & 57,700 \\
    \bottomrule
\end{tabular}
\label{tab:cagrad_toytask_niter}
\end{table}

\textit{Result:}  
For higher learning rates with Adam optimizer, even equal weighting (EW) reaches the global optimum 
(cf. \cref{fig:cagrad_toytask}, e.g., EW+Adam, lr=0.05) and often converges even faster than dedicated MTO methods (\cref{tab:cagrad_toytask_niter}).
Note, original results were shown for learning rate 0.001 using Adam and were, thus, in favor of CAGrad. 
Results for additional learning rates are reported in \cref{tab:cagrad_toytask_niter_full}. 

\textit{Conclusion:} 
The choice of optimizer appears to be more important on the success of the outcome of this experiment than the choice of MTO method, as Adam converges considerably faster and more reliably than GD. Also, tuning the learning rate is a relevant factor, however, especially in MTL with differently scaled losses, a single suitable learning rate for all tasks does often not exist.

\subsubsection{Experiments on Cityscapes and NYUv2}

We test the effectiveness of Adam and its role as a confounder in common MTL datasets for various MTO methods.

\textit{Approach:}
We compare Adam and SGD+mom in combination with any MTO method from equal weighting (EW), uncertainty weighting (UW)~\cite{kendall_uw_2018}, random loss weighting (RLW)~\cite{lin_rlw_2022}, PCGrad~\cite{yu_pcgrad_2020}, CAGrad~\cite{liu_cagrad_2021}, IMTL~\cite{liu_imtl_2021} and Aligned-MTL~\cite{senushkin_alignedmtl_2023}, for which we used the implementation from ~\cite{lin_libmtl_2022}, as well as MTL-IO~\cite{pascal2021improved} and  AdaTask~\cite{AdaTask_AAAI2023}. 
We distinguish between any combination of dataset \{Cityscapes~\cite{cordts_cityscapes_2016}, NYUv2~\cite{silberman_nyuv2_2012}\} and network architecture \{SegNet~\cite{badrinarayanan_segnet_2017}, DeepLabV3~\cite{chen_deeplabv3_2018}\}.
We run experiments for ten different initial learning rates from $[0.5, 0.1, 0.05, ... , 0.00001]$ and select the best one w.r.t. to the validation performance. More details are described in \cref{sec:appx_impl_details_adamsgd}. 
As different models and parameter setups can show preference towards different tasks and metrics, we are interested in those models which are Pareto optimal (PO). 

\begin{table}[tb]
\centering
\scriptsize
\setlength{\tabcolsep}{4pt}  
\newcommand{\gcs}{\hspace{8pt}}
\caption{
    \textbf{Number of Pareto optimal (PO) experiments using either Adam or SGD+mom. as optimizer.}
    Models trained with Adam are consistently more often on the Pareto front compared to those trained with SGD+mom.
    The number of Adam-based runs that are not dominated by any SGD-based run (PO w.r.t.\ SGD) is even higher, 
    while the reverse does not apply.
}
\begin{tabular}{llc@{\gcs}ccc@{\gcs}cc}
    \toprule
    &&& \multicolumn{2}{c}{Adam} && \multicolumn{2}{c}{SGD+mom.} \\
    \cmidrule(lr){4-5} \cmidrule(lr){7-8}
    & && PO (full) & PO w.r.t. SGD && PO (full) & PO w.r.t. Adam
    \\
    \midrule
    Cityscapes &SegNet && 5 & 24 && 0 & 0 \\ 
    Cityscapes &DeepLabV3 && 10 & 24 && 0 & 0 \\
    \midrule
    NYUv2 &SegNet && 11 & 21 && 1 & 1 \\
    NYUv2 &DeepLabV3 && 16 & 21 && 6 & 6 \\
    \bottomrule
\end{tabular}
\label{tab:adam_vs_sgd_pareto}
\end{table}

\begin{figure}[tb]
    \centering
    \begin{subfigure}{0.49\linewidth}
        \setlength{\belowcaptionskip}{0pt}
    \includegraphics[trim = 10mm 20mm 10mm 10mm, clip, width=1.\columnwidth]{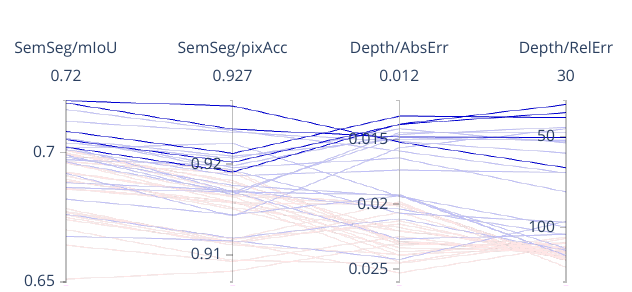}
        \caption{SegNet}
    \end{subfigure}
    \begin{subfigure}{0.49\linewidth}
        \setlength{\belowcaptionskip}{0pt}
    \includegraphics[trim = 10mm 20mm 10mm 10mm, clip, width=1.\columnwidth]{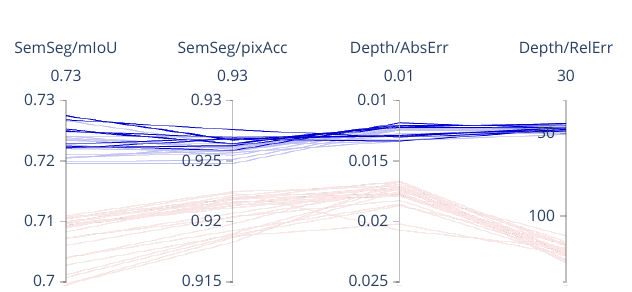}
        \caption{DeepLabV3}
    \end{subfigure}
    \caption{ 
    \textbf{Parallel coordinate plot over all experiments on Cityscapes.}
    We distinguish between experiments using {\color{red}SGD+mom} and {\color{blue}Adam} optimizer. 
    Experiments that reached Pareto front performance are drawn with higher saturation.
    We observe that Adam clearly outperforms the usage of SGD+mom.
    }
    \label{fig:adam_vs_sgd_pcp}
\end{figure}

\textit{Results:} 
We observe over all experimental setups that Adam performs favorably over SGD+mom (\cref{tab:adam_vs_sgd_pareto}). This especially holds true for experiments on Cityscapes where the Pareto front for both network architectures only consists of Adam-based models.
Moreover, an even larger number of Adam-based models is not dominated by any model trained with SGD+mom (PO w.r.t.\ SGD). 
For NYUv2,  Adam still performs stronger but SGD+mom.\ also occasionally delivers a PO result.
For the individual metrics, the predominance of Adam is further visualized in a parallel coordinate plot in \cref{fig:adam_vs_sgd_pcp,fig:adam_vs_sgd_pcp_v2}. 
Bold lines indicate the overall Pareto optimal experiments (PO full). 

In \cref{sec:additional_adam_sgd}, we further report best $\Delta_m$ results for common MTO methods in combination with Adam or SGD+mom (\cref{tab:res_cs_segnet,tab:res_cs_deeplab,tab:res_nyu_segnet,tab:res_nyu_deeplab}). 
Again, Adam boosts the overall performance across methods. 
Furthermore, when comparing the ranking of MTO methods w.r.t.\ the $\Delta_m$ metric, we see that the order can change based on the choice of optimizer, e.g., for Cityscapes with SegNet the best method with Adam is UW but with SGD+mom it is CAGrad. This underlines the importance of the choice of optimizer as a confounder in the experimental setup. 
Noteworthy, EW with Adam yields Pareto optimal results in three of the four setups (cf. \cref{tab:adam_vs_sgd_po_exp_count}) and is not dominated by any specialized MTO method trained in combination with SGD+mom for all dataset and network combinations. This supports claims questioning the effectiveness of specific MTO methods \cite{kurin_unitary-scalar_2022,xin_mto-even-help_2022}.
Nonetheless, looking at the $\Delta_m$ metric and individual metrics, we see that sometimes with a small relative performance drop on one metric, significant gains on another metric can be achieved (e.g., Cityscapes+sem.seg. and depth for UW vs EW).

\textit{Conclusion:} Not only a well-tuned learning rate but also the optimizer is crucial for MTL performance.
In a fair and extensive experimental comparison, we were able to show that Adam shows superior performance in MTL setup compared to SGD+mom.

\subsubsection{The reasonable effectiveness of Adam in the context of uncertainty weighting}

We show that Adam's mechanism to estimate a parameter-specific learning rate is partially loss-scale invariant and hypothesize that this could contribute to Adams effectiveness in MTL. We demonstrate this partial invariance theoretically and empirically. 
Furthermore, a full loss-scale invariance can also be shown under mild assumptions for UW \cite{kendall_uw_2018}, which is among the most prevalent loss weighting method in the literature, and related similar variant~\cite{lin2023dualbalancing}. 

The loss-scale invariance of UW can be shown by assuming an optimal solution for the $\sigma$ values similar to \cite{kirchdorfer2023analytical}. 
This assumption is mild as this is a 1-dimensional convex optimization problem for each $\sigma$. 
The invariance can be demonstrated by inserting the analytical solution starting from UW. For example, assuming a Laplacian distribution  (this can be shown for other distributions as well), we have
\begin{equation}
    \min_{\sigma_t} \frac{1}{\sigma_t} \mathcal{L}_t + \log \sigma_t \; \Rightarrow \; \sigma_t = \mathcal{L}_t
\end{equation}
The left hand side shows the typical form of UW, as shown for a Gaussian in \cite[eq.(5)]{kendall_uw_2018}.
Here, $\mathcal{L}_t$ is a task-specific loss and $\sigma_t$ is a scalar parameter that is usually learned. Plugging back the optimal solution for $\sigma_t$, we get
\begin{equation}
    \mathcal{L} = \sum_t \frac{\mathcal{L}_t}{sg[\mathcal{L}_t]} + c,
\end{equation}
where $sg$ is the stop-gradient operator and $c$ is a constant that can be omitted during optimization. Given this, we directly see the invariance w.r.t.\ loss-scalings. For instance, with $\mathcal{L}_1 \rightarrow \alpha_1 \mathcal{L}_1$ and $\mathcal{L}_2 \rightarrow \alpha_2 \mathcal{L}_2$, the derivative of the total loss $\mathcal{L}$ remains unchanged. 
As this invariance is shown on the loss-level, it holds for all gradient updates w.r.t.\ the head and backbone. 
Intuitively, this could explain why UW performs strongly in the context of various loss scalings such as measuring depth in centimeters or meters. 
Further details, are in \cref{sec:appx_adam_uwo}.

Similarly, for Adam, we can prove a partial scale invariance of losses in MTL that holds for the parameters of network heads. As before, we assume a hydra-like network architecture with a shared backbone and task-specific heads. 
We start with the parameter-update rule from Adam and scale the corresponding losses $\mathcal{L}_t \rightarrow \alpha_t \mathcal{L}_t$. 
When only considering the parameters of the corresponding heads $\psi_t$, the scalings $\alpha_t$ cancel out
\begin{equation}
        \psi_{t,i} = \psi_{t,i-1} - \frac{\gamma}{\sqrt{\alpha_t^2 \hat{v'_t}}} \alpha_t \hat{m'_t}. 
\end{equation}
Thus, for the network heads, we see a similar effect as for optimal UW that different scalings do not impact the network update. However, this does not hold for the backbone. 
The full derivation is shown in \cref{sec:appx_adam_uwo}. 
We confirm empirically in a handcrafted loss-scaling experiment in \cref{sec:app_invariance_empirical} and \cref{fig:scale_invariance_fixed,fig:scale_invariance_free} that SGD does not offer any scaling invariance, whereas Adam involves the invariance property for the heads. The optimal UW demonstrates a scaling invariance for the heads and the backbone. 

We would like to note that our derivation for Adam is only valid for constant $\alpha_t$, e.g., measuring depth in different units or unitary weightings \cite{xin_mto-even-help_2022, kurin_unitary-scalar_2022}. 
In case of dynamic loss weights that are not constant (e.g., UW), the weights do not cancel out fully due to the accumulation of gradient histories within Adam. 
Nonetheless, this has profound implications for loss weighting methods that are used in conjunction with Adam. 
For instance, when turning off the history within Adam  (by setting $\beta_{1,2}=0$) and having a fixed backbone, all loss weighting methods, such as UW, RLW, and others, become equivalent to equal weighting.

Additional ablations to our previous experiments suppport the relevance of invariance in MTL (cf. \cref{tab:res_nyu_segnet,tab:res_nyu_deeplab,tab:res_cs_segnet,tab:res_cs_deeplab}).
First, we compare to signSGD+mom~\cite{bernstein18signsgd} which only updates on the sign of gradients and is therefore trivially scale-invariant in the heads. 
In a direct comparison with SGD+mom, we observe a superiority of signSGD for a majority of tested setups. 
Next, we applied task-specific Adam optimizers as in AdaTask~\cite{AdaTask_AAAI2023} for a full loss-scale invariance and to allow an estimate of task-specific momentum and squared gradient accumulation. This is Pareto dominant over plain Adam+EW in almost all cases and significantly improves the $\Delta_m$ metric.

\textit{Conclusion:} In the context of MTL, we derive and measure a full loss-scale invariance for an optimal UW and a partial invariance for Adam. This partial invariance does not hold for SGD+mom and could explain, among other properties, the effectiveness of Adam in MTL. Furthermore, when comparing different loss weighting methods, it is crucial to be aware of the influence of the optimizer.

\subsection{Investigating gradient conflicts between tasks and samples}
\label{sec:exp_grad_conflicts}

\textit{Examined paradigm:} The field of MTL strongly focuses on resolving conflicts between tasks, especially from a perspective of gradient conflicts \cite{liu_cagrad_2021,yu_pcgrad_2020,javaloy_rotograd_2022}. In computer vision, tasks are often defined on a conceptional level such as segmentation and depth (Cityscapes), or recognizing multiple attributes (CelebA). However, in principle, conflicts can not only occur between tasks but also between samples within a task.

\begin{wrapfigure}{r}{0.29\textwidth}
    \vspace{-0.8cm}
  \centering
  \includegraphics[width=0.23\textwidth]{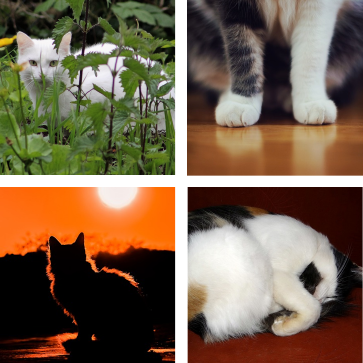} 
  \caption{High intra-task diversity can mimic MTL.}
  \vspace{-0.8cm}
\end{wrapfigure}
We argue that in an extreme case, even recognizing a single cat in multiple images could be considered MTL. 
For instance, in one image, the cat could be hiding behind a plant and only revealing its eyes, requiring a neural network to recognize the cat solely based on the eyes. 
In other images, the cat might only reveal its paws, front of a bright window, or might be tired and curled up into a furry ball because we took so many pictures. This would require a paw, shape or fur classifier. 
Thus, a neural network is required to recognize multiple attributes to reliably recognize our cat. We note that this is conceptually similar to the commonly \cite{sener_mgda_2018,yu_pcgrad_2020} considered MTL dataset CelebA~\cite{liu_celeba_2015} which requires attribute detection such as wavy hair, mustache or hat, but within one image.

Motivated by this example, we would like to quantify inter-task and inter-sample conflicts in common datasets from a perspective of the MTO literature, which inspects gradient conflicts in neural networks. 
In particular, we challenge the sole focus on \emph{inter-task} gradients conflicts in MTL. 
While several works follow the idea of overcoming gradient conflicts in MTL ~\cite{liu_cagrad_2021, javaloy_rotograd_2022, shi_recon_2023}, their appearance has only been mildly investigated so far.

\textit{Prerequisite:} 
We compare gradients w.r.t.\ network weights for different tasks $t$ and samples $\mathbf{x}_i$. 
The alignment of two gradients $\mathbf{g}, \mathbf{g'}$ on the shared parameters, e.g., of task $a$ and task $b$, is compared with the cosine similarity
\begin{equation}
S_{cos}(\mathbf{g},\mathbf{g'}) = \cos(\phi) = \frac{\mathbf{g}\cdot\mathbf{g'}}{\|\mathbf{g}\| \|\mathbf{g'}\|}.    
\end{equation}
Thus, two gradients are in conflict, if their cosine similarity is smaller than zero \cite{yu_pcgrad_2020}. In particular, $S_{cos}$ is $1$/$-1$ if gradients point in the same/opposite direction and $0$ in case of orthogonal directions.
The gradient magnitude similarity
\begin{equation}
    S_{mag}(\mathbf{g},\mathbf{g'}) = \frac{2\|\mathbf{g}\|_2\cdot\|\mathbf{g'}\|_2}{\|\mathbf{g}\|_2^2 + \|\mathbf{g'}\|_2^2}
\end{equation}
as defined in \cite{yu_pcgrad_2020}, yields values close to 1 for gradients of similar magnitude, or close to 0 for large discrepancies in magnitude.
High dissimilarity in both gradient direction and magnitude is presumed to be a common MTL problem.

\begin{figure}[tb]
    \centering
    \tiny
    \begin{subfigure}{0.49\linewidth}
        \setlength{\belowcaptionskip}{0pt}
        \begin{tabular}{cc}
        \rotatebox{90}{\hspace{10pt}SegNet} &
        \includegraphics[trim = 10mm 16mm 35mm 15mm, clip, width=0.93\columnwidth]{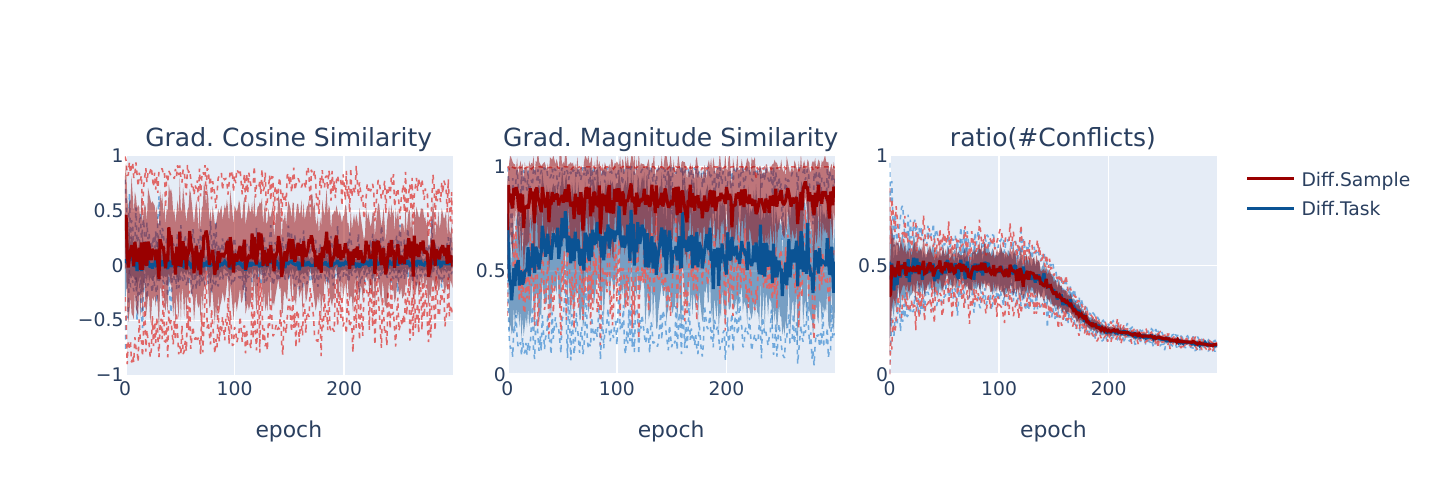}
        \\
        \rotatebox{90}{\hspace{2pt}DeepLabV3} &
        \includegraphics[trim = 10mm 16mm 35mm 15mm, clip, width=0.93\columnwidth]{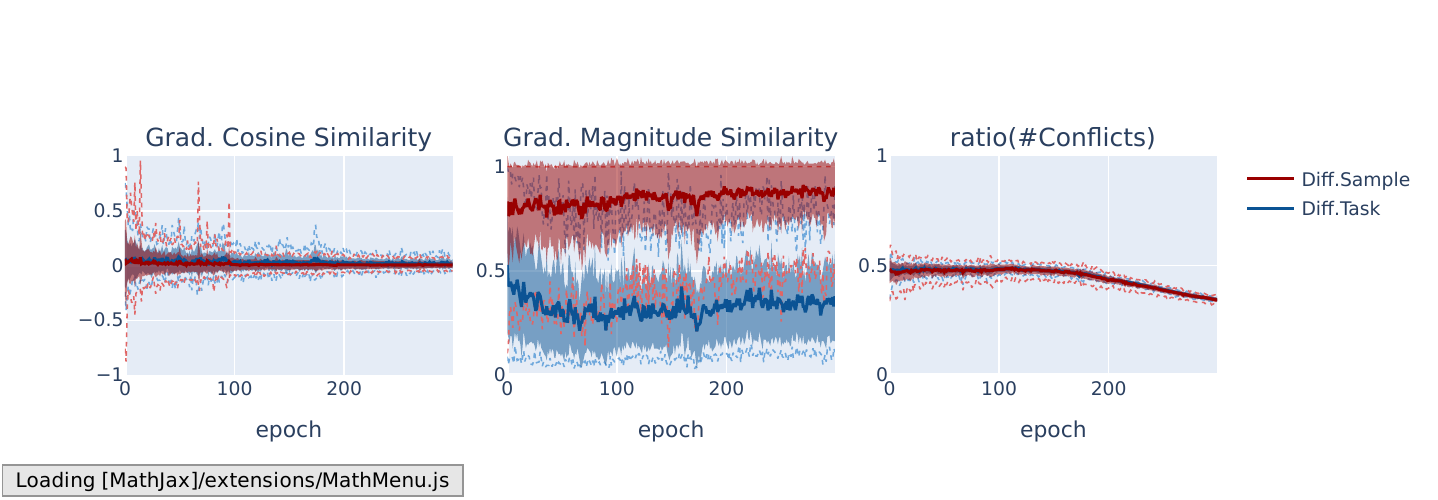} 
        \end{tabular}
        \caption{Cityscapes}
    \end{subfigure}
    \hfill
    \begin{subfigure}{0.49\linewidth}
        \setlength{\belowcaptionskip}{0pt}
        \begin{tabular}{cc}
        \rotatebox{90}{\hspace{10pt}SegNet}&
        \includegraphics[trim = 10mm 16mm 35mm 15mm, clip, width=0.93\columnwidth]{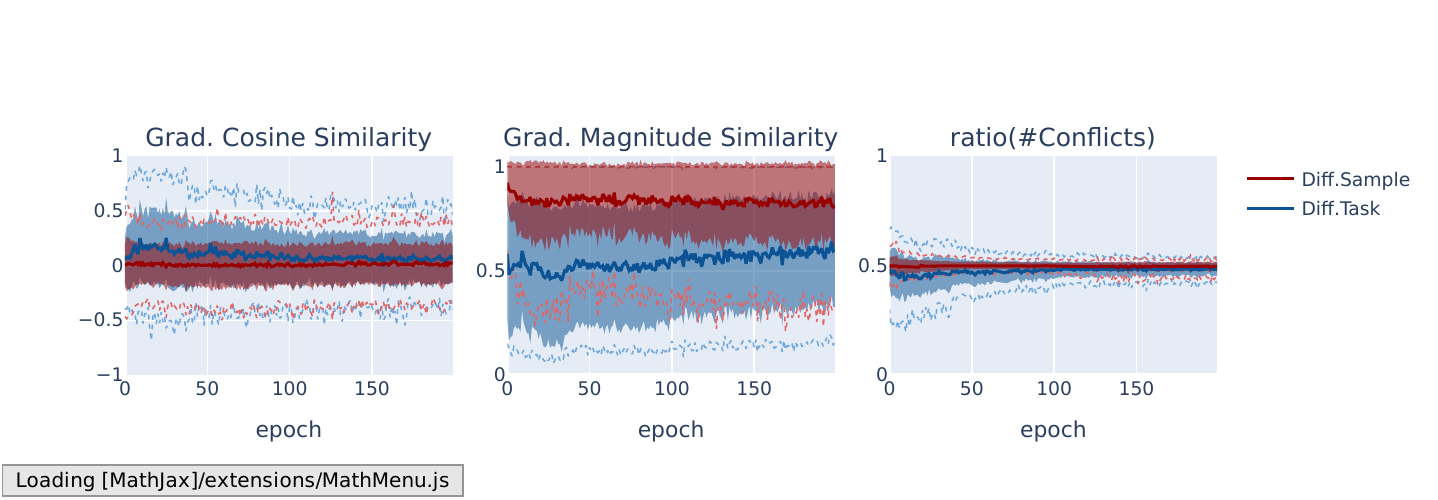} 
        \\
        \rotatebox{90}{\hspace{2pt}DeepLabV3}&
        \includegraphics[trim = 10mm 16mm 35mm 15mm, clip, width=0.93\columnwidth]{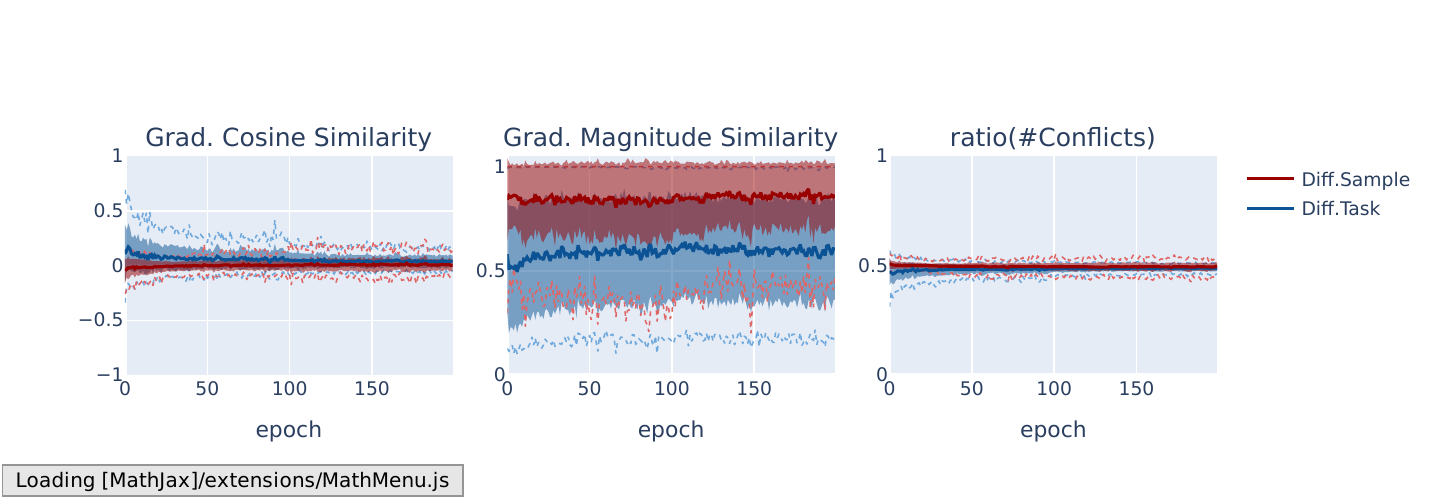} 
        \end{tabular}
        \caption{NYUv2}
    \end{subfigure}\\
    \begin{subfigure}{0.49\linewidth}
        \setlength{\belowcaptionskip}{0pt}
        \begin{tabular}{cc}
        \rotatebox{90}{\hspace{2pt}ResNet50}&
        \includegraphics[trim = 10mm 16mm 35mm 15mm, clip, width=0.93\columnwidth]{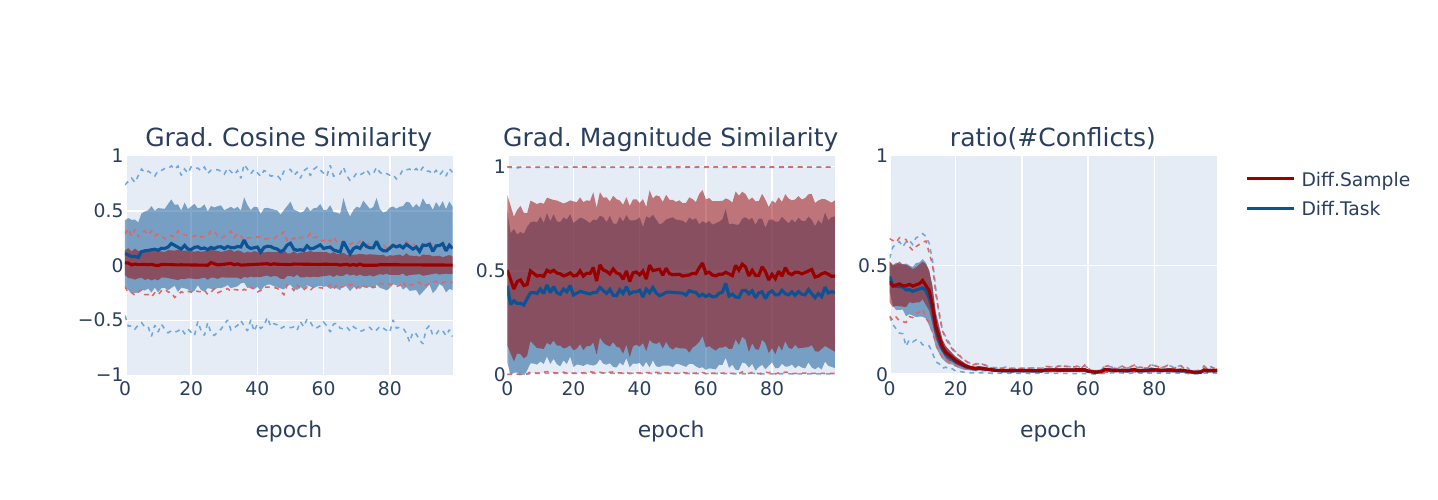} 
        \end{tabular}
        \caption{CelebA}
    \end{subfigure}
    \vspace{-0.2cm}
    \caption{\textbf{Gradient similarities and conflicts} for different datasets and network architectures over training epochs. 
    For each dataset and network combination, we report (from left to right) gradient cosine similarity, gradient magnitude similarity, and the ratio of conflicting gradient parameters w.r.t. gradient pairs corresponding to either {\color{red}inter-samples (fixed task)} or {\color{blue}inter-tasks (fixed sample)}.
    We report mean (solid line), standard deviation (shaded area), upper ($97.5\%$) and lower ($2.5\%$) percentile (dotted line) within an epoch. 
    Overall, the direction conflicts are similar (first / last column), whereas the magnitude differences are more pronounced in MTL (middle column). 
    }
    \label{fig:gradsim}
\end{figure}

\textit{Approach:} 
During the training on aforementioned datasets, we examine gradient similarity across two different setups: (1) between gradients of different tasks with respect to a single sample (\textbf{inter-task}), e.g., $\mathbf{g} = \nabla_\phi L_0 \left( f_{\boldsymbol{\theta}}(\mathbf{x}_i) \right)$ and $\mathbf{g'} = \nabla_\phi L_1 \left( f_{\boldsymbol{\theta}}(\mathbf{x}_i) \right)$; and (2) between gradients corresponding to the same task but different samples within a batch (\textbf{inter-sample}), e.g., $\mathbf{g} = \nabla_\phi L_t \left( f_{\boldsymbol{\theta}}(\mathbf{x}_0) \right)$ and $\mathbf{g'} = \nabla_\phi L_t \left( f_{\boldsymbol{\theta}}(\mathbf{x}_1) \right)$.
For both setups, we compute the gradient cosine similarity and gradient magnitude similarity as well as the ratio of conflicting gradient parameters.
We are aware that our comparison between samples and tasks is not direct. Nonetheless, it serves as a coarse indicator to estimate their impact during network training. 
Implementation details are in \cref{sec:appx_impl_details}.

\textit{Results:} 
We show the evolvement of the gradient similarity measures over epochs in \cref{fig:gradsim}.
Surprisingly, when comparing inter-sample (red line) and inter-task (blue line), we find no consistent evidence for gradient alignment conflicts (left column) to be an exclusive problem of having multiple tasks. 
For instance, for Cityscapes, the variation of gradient alignment is fully encapsulated within the spread we observe in inter-sample variation (task is fixed). For CelebA, the converse seems to be mostly the case. 
Furthermore, the choice of network architecture and distribution of task-specific and shared parameters (SegNet vs.\ DeepLabV3) can have a large influence on the spread of the cosine-similarity. Both architectures have roughly a similar number of shared-parameters. However, DeepLabV3 has a higher number of task-specific parameters which seems to reduce the variance in conflicts for both inter-sample and inter-task (row one vs.\ two).
In line with these observations, we found a similar number of conflicting gradient parameters (third column) for both inter-sample and inter-task comparisons among all experiments.

For gradient magnitude similarities (middle column), we observe a clearer pattern. The similarity in magnitudes are continuously (in the mean) less pronounced for the inter-task setup compared to inter-samples (blue line is below red one in all settings). 
Interestingly, the relative difference between the two setups remains similar over training which justifies the choice of fixed scalar task weightings as done in \cite{xin_mto-even-help_2022}.
Further measures can be found in \Cref{fig:gradsim_posneg,fig:gradsim_batch,fig:gradsim_scalarproduct}.

\textit{Conclusion:} We find that the difficulty of MTL (inter-task and inter-sample) as opposed to STL (inter-sample only) is predominantly due to differences in gradient \emph{magnitudes}. Balancing different magnitudes is tackled in the literature, e.g.,  \cite{kendall_uw_2018,xin_mto-even-help_2022}
The problem of conflicting gradients has been typically associated with task-specific conflicts \cite{yu_pcgrad_2020, liu_cagrad_2021, javaloy_rotograd_2022}, here, we find that gradient \emph{alignment} conflicts can actually be even more pronounced between samples.
On the one hand, these observations are along the same lines as findings by Royer et al.~\cite{royer2023scalarization} who reason that 'correcting conflicting gradients [between tasks] at every training iteration can be superfluous'.
On the other hand, gradient-alignment methods in MTL could be considered not only in the context of task-specific conflicts but also for conflicts between samples. 
Interestingly, previous work has explored the potential benefit of not only learning weights per task but also per sample in the dataset~\cite{vasu_21_insttaskparams}. While our experiments show a relatively high similarity in gradient magnitude across samples and, thus, don't motivate a sample-wise loss weighting, this could, however, be due to only little disruptive noise within data samples which has been the main motivation of \cite{vasu_21_insttaskparams}.

\section{Conclusion and outlook}
\label{sec:exp_conclusion}

This study aims to enhance our understanding of multi-task learning (MTL) in computer vision, providing valuable insights for future research as well as guidance for implementations of real-world applications.

We show that common optimization methods from single task learning (STL) like the Adam optimizer are effective in MTL problems. 
Next, we compare gradient conflicts during training between tasks and samples. 
While gradient \emph{magnitudes} are a specific problem between tasks (MTL) and thus justify the need for multi-task specific methods for automatic loss weighting, we find the variability in gradient \emph{alignment} to be similar between samples and tasks.  
Thus, we encourage a more unified viewpoint in which specific MTO methods are also considered in single-task problems and vice versa.

Beyond our work, we encourage to improve the understanding of challenges and paradigms specific to MTL.
For instance, our understanding of task (and sample) specific capacity allocation within a network and how best to tune it to custom requirements, is still not thoroughly understood. 
Often task-weights are increased to assign more importance to a task which is in contrast to tuning the learning rate per task where a smaller learning rate can be beneficial. Thus, we require further investigations and disentanglement of these two concepts.

\section*{Acknowledgments}
We thank Claudia Blaiotta, Martin Rapp, Frank R. Schmidt, Leonhard Hennicke, and Bastian Bischoff for their feedback and valuable discussions. 
Cathrin Elich thanks her supervisors, Jörg Stückler and Marc Pollefeys, for enabling the opportunity to pursue an internship during her Ph.D. studies.

The Bosch Group is carbon neutral. Administration, manufacturing and research activities do no longer leave a carbon footprint. This also includes GPU clusters on which the experiments have been performed. 

%
%
%
\bibliographystyle{splncs04}
\bibliography{ms}

\appendix

\clearpage
\newpage
\onecolumn

\setlength{\belowcaptionskip}{0pt} 


\begin{center}
    \large  
    \textbf{Examining Common Paradigms in Multi-Task Learning \\-Supplementary Material-} 
\end{center}

\renewcommand{\thetable}{A\arabic{table}}
\renewcommand{\thefigure}{A\arabic{figure}}
\renewcommand{\thesection}{A\arabic{section}}

\setcounter{table}{0}
\setcounter{figure}{0}
\setcounter{section}{0}
\setcounter{page}{1} 

\section{Theoretical insights into multi-task learning dynamics}
\label{sec:appx_adam_uwo}

In this section, we aim to explain the success of the Adam optimizer \cite{kingma__adam__2015} by relating it to uncertainty weighting \cite{kendall_uw_2018}. 
We show partial invariances w.r.t.\ prior task-weights for the Adam optimizer and full invariances for the uncertainty weighting under mild assumptions. We further show that for SGD + momentum no invariance can be observed. Instead, the loss-weight can be seen as a task-specific learning rate which is not the case for the Adam optimizer. 
Previous literature on weighting methods in MTL did not explicitly show how task-weighting methods are affected by different optimizers.

\subsection{Uncertainty weighting (UW): Full loss-scale invariance}

In UW~\cite{kendall_uw_2018}, the homoscedastic uncertainty\footnote{In Kendall et al., this is termed the aleatoric homoscedastic uncertainty. However, as the task weights vary over the course of training and also with respect to the model capacity, it is technically not only the aleatoric uncertainty but also encapsulates further components such as model capacity and amount of data seen.} 
$\sigma_t$ to weight task $t$ is learned by gradient descent. However, we can also analytically compute the optimal uncertainty weights in each iteration instead of learning them using gradient descent as done in \cite{kirchdorfer2023analytical}. 
The minimization objective depends on the underlying loss function and likelihood. For simplicity, we show the derivation exemplary for the $L_1$ loss. It is straight-forward to derive the same for a Gaussian and other distributions.
The objective of uncertainty weighting is given as

\begin{equation}
    \min_{\sigma_t} \frac{1}{\sigma_t} \mathcal{L}_t + \log \sigma_t
\end{equation}

with $\mathcal{L}_t=|y-f^W(x)|$ which can be derived from a log likelihood of a Laplace distribution $p(y|f^W(x), \sigma) = \frac{1}{2\sigma} exp(-\frac{|y - f^W(x)|}{\sigma})$. Taking the derivative and solving for $\sigma_t$ results in an analytically optimal solution:

\begin{equation}
    \frac{\partial}{\partial \sigma_t} \frac{1}{\sigma_t} \mathcal{L}_t + \log \sigma_t = -\frac{1}{\sigma_t^2}\mathcal{L}_t + \frac{1}{\sigma_t}
\end{equation}

\begin{equation}
    -\frac{1}{\sigma_t^2}\mathcal{L}_t + \frac{1}{\sigma_t} \overset{!}{=} 0
    ~ ~ ~ ~ \Rightarrow ~ ~ ~ ~
    \sigma_t = \mathcal{L}_t
    \end{equation}


with $\sigma_t>0$. 
As the optimization problem is convex and just one dimensional, assuming an optimal log-sigma is a mild assumption.  
Plugging the optimal solution back into the original uncertainty weighting, we get

\begin{equation}
    \mathcal{L} = \sum_t \frac{1}{sg[\mathcal{L}_t]} \mathcal{L}_t + \log \sqrt{sg[\mathcal{L}_t]},
\end{equation}

where we denote $sg$ as the stopgradient operator. 

Since there is no gradient for the second part of the loss, it can be simplified such that

\begin{equation}
    \mathcal{L} = \sum_t \frac{\mathcal{L}_t}{sg[\mathcal{L}_t]}.
\end{equation}

Assuming task-specific weights $\alpha_t$, we get

\begin{equation}
\begin{split}
    \mathcal{L} & = \sum_t \frac{\alpha_t \mathcal{L}_t}{\alpha_t sg[\mathcal{L}_t]} \\
    & = \sum_t \frac{\mathcal{L}_t}{sg[\mathcal{L}_t]}
\label{eq:uw_o_invariance}
\end{split}
\end{equation}

Therefore, the optimal uncertainty weighting is invariant w.r.t.\ task-specific loss-scalings, as each scaling cancels out.

\subsection{SGD: No loss-scale invariance and relationship of learning rate and task weights on a gradient level}
\label{sec:appx_lr_task_weight}
Unlike for optimal UW, we show that the SGD update rule does not show any invariances and that task-weights are essentially task-specific learning rates. 
Instead, task-weights and learning rate are interacting hyperparameters and thus cannot be viewed in isolation. 

The parameter update rule in neural networks optimized with SGD is
\begin{equation}
    \theta_i = \theta_{i-1} - \gamma \frac{\partial}{\partial \theta_{i-1}} \mathcal{L},
\label{Formula:SGD}
\end{equation}

where the network parameters in iteration $i$ are defined as $\theta_i$, $\gamma$ is the learning rate and $\mathcal{L} = \sum_t \alpha_i \mathcal{L}_t$ . 

In the case of uniform task weights (EW), $\alpha =\alpha_i \forall i$, we have

\begin{equation}
    \begin{split}
        \theta_i & = \theta_{i-1} - \gamma \frac{\partial}{\partial \theta_{i-1}} \sum_i \alpha \mathcal{L}_t \\
        & = \theta_{i-1} - \gamma \alpha \frac{\partial}{\partial \theta_{i-1}} \sum_i \mathcal{L}_t\\
    \end{split}
    \end{equation}

Here, task weight and learning rate are interchangeable. 
In particular, increasing the weight $\alpha$ by a constant factor $c$ has the same effect as increasing the learning rate by a factor $c$. 

In the case of non-uniform task weights $\alpha_i$, the parameter update is 
\begin{equation}
    \begin{split}
        \theta_i & = \theta_{i-1} - \gamma \frac{\partial}{\partial \theta_{i-1}} \sum_i \alpha_t \mathcal{L}_t \\
        & =  \theta_{i-1} - \frac{\partial}{\partial \theta_{i-1}} \sum_i \gamma \alpha_t \mathcal{L}_t \\
    \end{split}
    \end{equation}

As the learning rate can be included in the task-specific weight, it follows that task weighting is interchangeable to assigning \textit{task-specific} learning rates. Tasks with a higher weight $\alpha_i$ have a proportionally higher parameter update step and vice versa.

While this holds for SGD and SGD + momentum, it does not apply to optimizers such as Adam, Adagrad, or RMSProp. We demonstrate this for Adam in the following subsection.

\subsection{Adam: Partial loss-scale invariance}
Similarly to the invariance demonstrated for optimal UW, we derive a partial invariance for Adam. 
In their work, Kingma and Ba \cite{kingma__adam__2015} have already shown that the magnitudes of the parameter updates using Adam are invariant to rescaling the gradients. Our novelty lies in demonstrating this invariance property in the context of MTL and its impact on different MTO methods. For Adam, we claim that the magnitude of task-specific weights only affects the backbone and cancels out for the heads.

We consider the standard MTL model setting with a shared backbone and task-specific heads. 
In this analysis, we assume a frozen backbone and only look at the task-specific parameters $\psi_t$ of task $t$ whose loss $\mathcal{L}_t$ is scaled by $\alpha_t$, such that $\mathcal{L}_t \rightarrow \alpha_t \mathcal{L}_t$. 
The parameter update of one head is independent of the other heads as the derivative of the losses w.r.t.\ the other tasks is $0$:

\begin{equation}
     \frac{\partial}{\partial \psi_{t,i-1}} \mathcal{L}_j = 0 \; \; \mathrm{for \: t \neq j}.
\end{equation}

The general update rule for parameters $\psi$ at time step $i$ using Adam is 

\begin{equation}
    \psi_{i} = \psi_{i-1} - \frac{\gamma}{\sqrt{\hat{v_i}} + \epsilon} \hat{m_i},
\end{equation}

where $m_i = \beta_1m_{i-1} + (1-\beta_1)g_i$ and $v_i = \beta_2v_{i-1} + (1-\beta_2)g_i^2$. To counteract the bias towards 0, the moments are corrected as $\hat{m_i} = \frac{m_i}{1-\beta_1^i}$ and $\hat{v_i} = \frac{v_i}{1-\beta_2^i}$.

For task-specific parameters $\psi_t$, task weights $\alpha_t$ linearly scale the first moment $m_{t,i}$

\begin{equation}
    \begin{split}
        m_{t,i} & = \beta_1m_{t,i-1} + (1-\beta_1)g_{t,i} \\
        & = \beta_1m_{t,i-1} + (1-\beta_1) \frac{\partial}{\partial \psi_{t,i-1}} \alpha_t \mathcal{L}_t \\
        & = \beta_1m_{t,i-1} + (1-\beta_1) \alpha_t \frac{\partial}{\partial \psi_{t,i-1}} \mathcal{L}_t \\
        & = \beta_1m_{t,i-1} + (1-\beta_1) \alpha_t g'_{t,i} \\
    \end{split}
    \end{equation}

and quadratically scale the second moment $v_{t,i}$

\begin{equation}
    \begin{split}
        v_{t,i} & = \beta_2 v_{t,i-1} + (1-\beta_2)g_{{t,i}}^2 \\
        & = \beta_2 v_{t,i-1} + (1-\beta_2) (\frac{\partial}{\partial \psi_{t,i-1}} \alpha_t \mathcal{L}_t)^2 \\
        & = \beta_2 v_{t,i-1} + (1-\beta_2) \alpha_t^2 (\frac{\partial}{\partial \psi_{t,i-1}} \mathcal{L}_t)^2 \\
        & = \beta_2 v_{t,i-1} + (1-\beta_2) \alpha_t^2 {g'}_{t,i}^2, \\
    \end{split}
    \label{eq:v_i}
    \end{equation}
where ${g'}_{t,i}$ is the gradient of the unscaled loss $\mathcal{L}_t$ w.r.t. the task-specific parameters for task $t$.
As this holds for iteration $i$ and because we have $m_{t,1} = \alpha {g'}_{t,1} + 0$ respectively $ v_{t,1} = \alpha_t^2 {g'}^2_{t,1} + 0$  with $m_{t,0} = 0$, $v_{t,0} = 0$ at the first iteration, this holds for any iteration step. 
We can thus rewrite $\hat{m}_{t,i} = \alpha_t \hat{m}'_{t,i}$ and $\hat{v}_{t,i} = \alpha_t^2 \hat{v'}_{t,i}$.

Plugging this back into the update rule, we get

\begin{equation}
    \begin{split}
        \psi_{t,i} & = \psi_{t, i-1} - \frac{\gamma}{\sqrt{\hat{v}_{t}i}} \hat{m}_{t,i} \\
        & = \psi_{t,i-1} - \frac{\gamma}{\sqrt{\cancel{\alpha_t^2} \hat{v'}_{t,i}}} \cancel{\alpha_t} \hat{m'}_{t,i} 
        \label{eq:adam_hist_invariance}
    \end{split}
    \end{equation}

where the loss-scaling $\alpha_t$ cancels out. Therefore, the parameters of the task-specific heads are invariant to loss-scalings using Adam. 

This partial invariance is a highly desired property as there is a fundamental trade-off between tuning the learning rate and manual task weights. Given Adams invariance for the head, the weighting only affects the backbone. Thus the learning rate can be set for the parameters of the head independent of the loss weights. 
With the loss weights, we can prioritize tasks in the backbone and therefore walk along the Pareto front as empirically shown by \cite{xin_mto-even-help_2022}. 

The invariance, however, does not hold anymore when the backbone parameters $\theta$ are updated as well.
As we have
\begin{equation}
    \begin{split}
    m_i &= \beta_1 m_{i-1} +(1-\beta_1)\frac{\partial}{\partial \theta_{i-1}} \sum_t \alpha_t \mathcal{L}_t\\
    &= \beta_1 m_{i-1} +(1-\beta_1)\sum_t \alpha_t  g'_{t,i}
    \end{split}
\end{equation}
and
\begin{equation}
    \begin{split}
    v_i &= \beta_1 v_{i-1} +(1-\beta_1)(\frac{\partial}{\partial \theta_{i-1}} \sum_t \alpha_t \mathcal{L}_t)^2\\
    &= \beta_1 v_{i-1} +(1-\beta_1) (\sum_t \alpha_t g'_{t,i})^2
    \end{split}
\end{equation}
we conclude that the task weights $a_t$ linearly affect the first moment $m_i$, while  having a quadratic effect on the update of the second moment $v_i$.

Note that for both task-heads only as well as the backbone, we have a full invariance in case of independent optimizers, e.g., one Adam optimizer per task similar to \cite{pascal2021improved,AdaTask_AAAI2023}. However, naive implementations scale poorly (in terms of computational complexity) with the number of tasks here. \\

In the following experiments, we provide empirical evidence for our finding that a) Adam offers loss-scale invariance for the parameters of the task-specific heads, and b) Adam offers loss-scale invariance for all network parameters (backbone and heads) if $\beta_{1,2} = 0$.

\section{Empirical Confirmation of scale invariances in Adam and Optimal Uncertainty Weighting}
\label{sec:app_invariance_empirical}

In the prior section, we derived theoretical results for loss-scale (partial) invariance within multi-task learning for the Adam optimizer and uncertainty weighting. In this section, we confirm this invariance empirically with a toy task. 

\paragraph{Experimental Setup}
We consider a two-task toy experiment in which we look at the gradient magnitudes with different combinations of Adam, SGD, EW, optimal uncertainty weighting (UW-O), and loss-scalings. 
To generate the data, we sample scalar input values from a uniform distribution; the outputs are just scalings of the input. We apply a simple neural network which consists of a shared backbone (two layers with LeakyReLU as non-linearity and 20 neurons per hidden layer) and two heads for the two tasks, each consisting again of two layers. 
Both task measure the depth but in different units using the $L_1$-loss. 

We provide two settings: In the first one, depth is measured on the same scale. In the second setting, one depth loss is scaled by 10x (e.g., measured in cm instead of deci-meters) and one other loss is scaled by 0.1 (e.g., measured in meters instead of deci-meters). 
For each setting, we test various combinations of loss weighting and optimizer combinations. 

The 8 different experiments are:

\begin{itemize}[noitemsep]
    \item EW using SGD
    \item EW using SGD with scalings $10 \cdot L_{seg}$, $0.01 \cdot L_{dep}$\\
    \item EW using Adam
    \item EW using Adam with scalings $10 \cdot L_{seg}$ and $0.01 \cdot L_{dep}$ \\
    \item UW-O) using SGD
    \item UW-O) using SGD with scalings $10 \cdot L_{seg}$ and $0.01 \cdot L_{dep}$ \\
    \item EW using separate Adam  optimizers per task 
    \item EW using separate Adam  optimizers per task with scalings $10 \cdot L_{seg},0.01 \cdot L_{dep}$
\end{itemize}

To better control for different factors of influence, we first perform the first 6 of the listed experiments with a fixed backbone, i.e., we do not update the parameters in the backbone but only in the heads. Afterward, we show all 8 experiments trained with a network where all parameters (including the backbone) are updated. This allows us to verify if our theoretical derivations regarding the (partial) loss-scaling invariance of Adam and UW-O also hold in practice, and compare this to the SGD optimizer.

Note that we only care about the invariance and did not tune any hyperparameters for performance. 

\begin{figure}
    \centering
    \includegraphics[trim = 0mm 0mm 0mm 0mm, clip, width=\columnwidth]{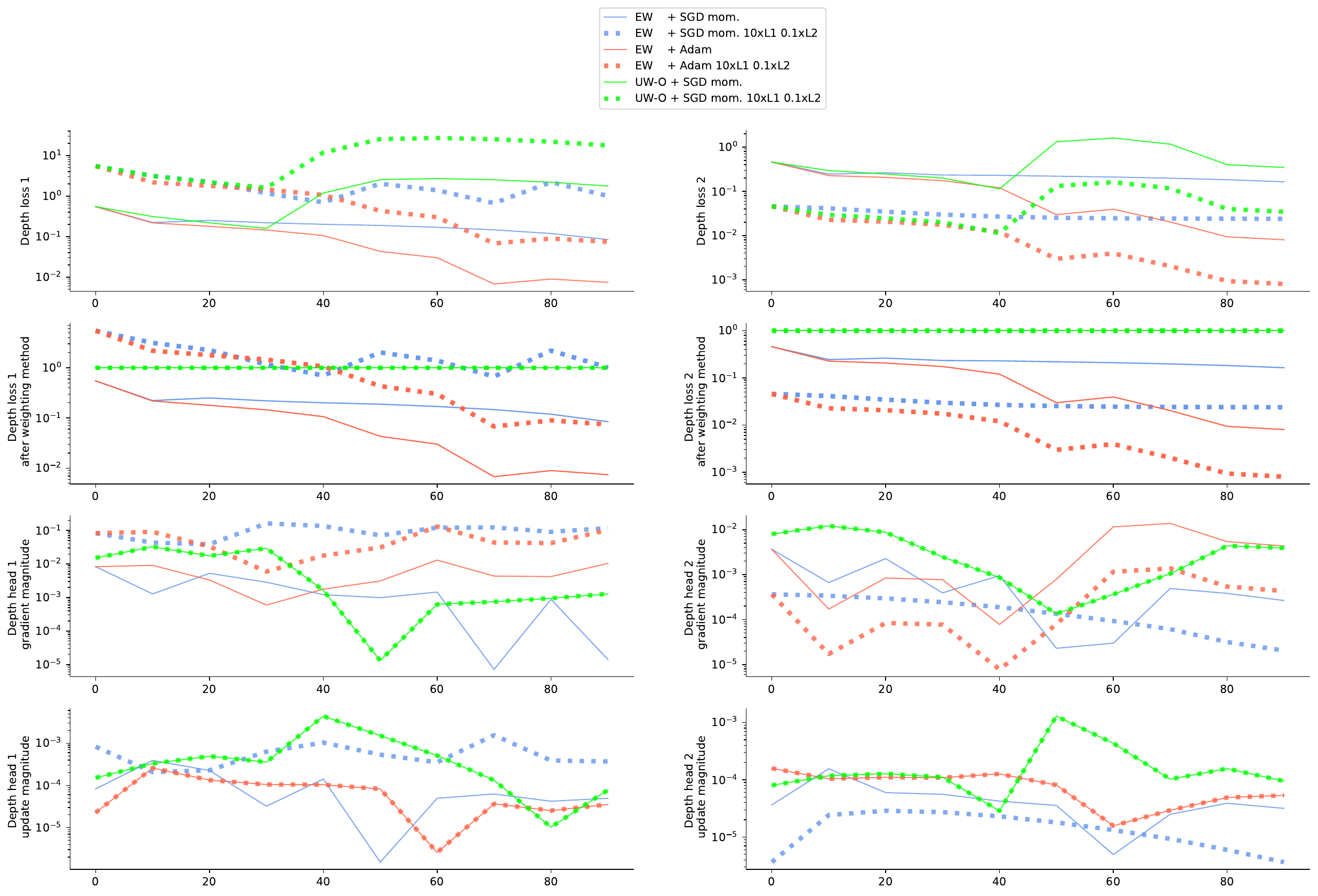}
    \caption{\textbf{Invariances within the neural network for a frozen backbone. } Comparing the effect of loss-scalings in a toy experiment with two tasks. For each optimizer and loss weighting combination, we run two settings with a) loss L1 and loss L2 are equally weighted or b) L1 is scaled by 10x and L2 by 0.1. For each setting, we measure the SGD + momentum and Adam optimizer with no post weighting (EW) and SGD + momentum with optimimal uncertainty weighting. We show the scaled losses, gradient magnitudes, and gradient update magnitudes in the the two task heads and keep the backbone frozen. While SGD does not offer any loss-scaling invariance, Adam makes the gradient updates of the head parameters invariant to scales confirming our derivation (red lines overlap in lowest row). Equivalently, for UW-O we also observe the theoretically derived invariances (green lines overlap in lowest row)}
    \label{fig:scale_invariance_fixed}
\end{figure}

\clearpage 
\paragraph{Results for fixed backbone}
Figure \ref{fig:scale_invariance_fixed} shows the losses, the scaled losses (by loss weighting method), the gradient magnitudes as well as the gradient update magnitudes for both heads along the 100 epochs of training with a fixed backbone. 
Regarding SGD, we can observe that the equal weighting experiment differs from its scaled variant along all 8 dimensions. This is because SGD does not offer any loss-scaling invariance. As expected, at the beginning of the training the gradient update magnitude of the first depth head parameters with the scaled loss ({\color{blue}dotted line}) is by a factor of $10$  higher than the unscaled ({\color{blue}solid line}) one. The same effect applies to the gradient update magnitude of the second depth head parameters, but with a factor of $0.01$. 

In contrast, Adam is loss-scale invariant. We can observe that the unscaled ({\color{red}solid line}) and the scaled version ({\color{red} dotted line}) have equal gradient update magnitudes in the last row. Note that practically due to an $epsilon=10^{-8}$ parameter in the denominator and float precision a slight divergence would occur with larger number of epochs.
This result confirms our theoretical finding in equation \ref{eq:adam_hist_invariance}.
We skip the experiment of separated Adam optimizers per task because it would be equivalent to this version given a fixed backbone. 

Lastly, we want to investigate the invariance properties of UW-O. We compare the scaled ({\color{green}dotted line}) and unscaled ({\color{green}solid line}) version of UW-O with the SGD optimizer. As expected, the gradients, as well as the gradient updates, match in both heads.

In the following, let's investigate whether the observed results still hold if we also consider the update of the backbone parameters.

\begin{figure}
    \centering
    \includegraphics[trim = 0mm 0mm 0mm 0mm, clip, width=\columnwidth]{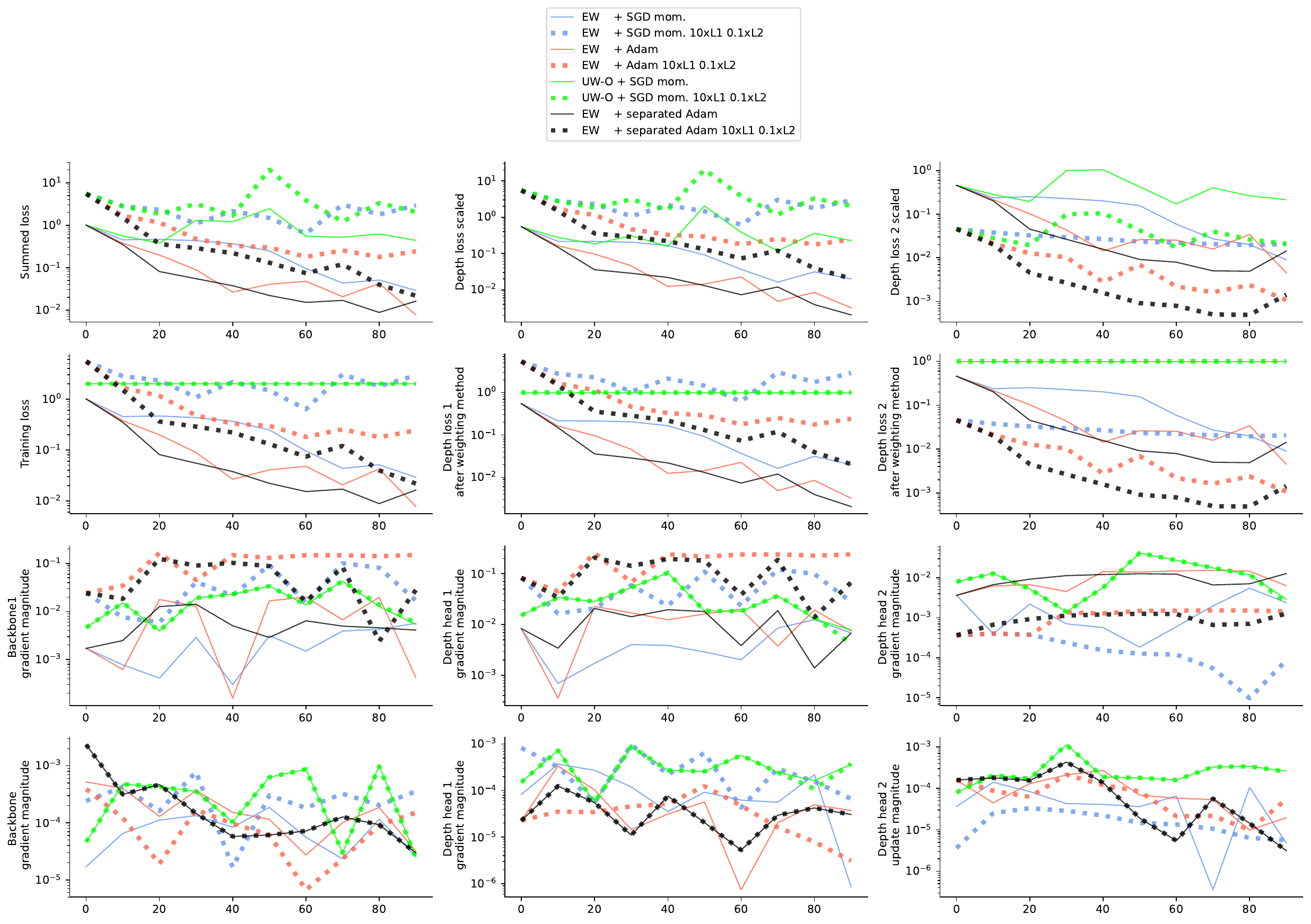}
    \caption{\textbf{Invariances within the neural network for a learnable backbone. } Comparing the effect of loss-scalings in a toy experiment with two tasks. For each optimizer and loss weighting combination, we run two settings with a) loss L1 and loss L2 are equally weighted or b) L1 is scaled by 10x and L2 by 0.1. For each setting, we measure the SGD + momentum and Adam optimizer with no post weighting (EW) and SGD + momentum with optimimal uncertainty weighting. Additionally, we implement independent Adam optimizer per task. We show the scaled losses, gradient magnitudes, and gradient update magnitudes in the backbone(first row) and the the two task heads (2nd and 3rd row). Neither Adam, nor SGD show invariances if the backbone is trained as well. UW-O is still invriant (green lines are overlapping). We revoke Adam's inveriance by implementing separate optimizers per task (lowerst black lines are overlapping).}
\label{fig:scale_invariance_free}
\end{figure}

\paragraph{Results for free backbone}
Figure \ref{fig:scale_invariance_free} shows the scaled losses, the gradient magnitudes as well as the gradient update magnitudes in the backbone and the depth heads along the 100 epochs of training with a free backbone. Again, the loss-scalings affect the gradient magnitudes using SGD. This applies to both backbone and heads.

When looking at the Adam experiments, we can observe that it is partly loss-scale invariant by looking at the first iteration in the heads. However, due to different updates in the backbone, the networks behave different in both settings (scaled and unscaled loses). 
Furthermore, when implementing task-specific optimizers, we can observe that not only the gradient update magnitudes in the task heads, but also in the backbone match between the scaled (\textbf{{\color{black}dotted line}}) and the unscaled (\textbf{{\color{black}solid line}}) variant.
Thus, all network parameters are invariant to loss-scalings when using separate Adam optimizers. This confirms our theoretical results.

Along the lines of our theoretical findings, we can observe that UW-O offers scaling-invariance across the whole network as the gradients as well as the gradient updates match among the two variants in the backbone and in both heads. This empirical observation matches our theoretical derivation in equation \ref{eq:uw_o_invariance}.

\clearpage
\section{Implementation  Details}
\label{sec:appx_impl_details}

In this section, we explain the applied settings used for the reported experiments in more detail.
In particular, we describe the handling of the different datasets in \cref{sec:appx_dataset_details} and provide further information on the applied training procedures in \cref{sec:appx_training_details}.
Our chosen experimental setups are designed to follow previous work and mainly inspirited by \cite{liu_cagrad_2021,xin_mto-even-help_2022,lin_libmtl_2022}.
However, we found that the experimental setup would vary widely across different works in the field of multi-task learning as can be seen in \Cref{tab:rel_work}.
We use a uniform setup for each dataset independent of the choice of network and MTO.

\subsection{Datasets}
\label{sec:appx_dataset_details}

\paragraph{Cityscapes}\cite{cordts_cityscapes_2016}
We make use of the official split of the dataset which consists of 2975 training and 500 validation scenes.
Similar to ~\cite{xin_mto-even-help_2022}, we denote 595 random samples from the training split as validation data and report test results on the original validation split.
We further follow the pre-processing scheme from ~\cite{liu_mtan_2019} of re-scaling images to 128x256 pixels and use inverse depth labels.
During training, we apply random scaling and cropping for data augmentation\footnote{\label{fn:cagrad_code}https://github.com/Cranial-XIX/CAGrad}.
Following previous work~\cite{liu_cagrad_2021} for number of epochs and learning rate schedule, we train for 300 epochs and decrease the learning rate by a factor of 0.5 every 100 epochs.
The batch size is setto 64, similar to \cite{xin_mto-even-help_2022}.
We only consider a fixed weight decay of $10^{-5}$ for all datasets and experiments as we found varying this parameter had only little influence in initial experiments.

\paragraph{NYUv2}~\cite{silberman_nyuv2_2012}
From the 795 official training images we use 159 for our validation split as in \cite{lin_libmtl_2022} and report test performance on the official 654 test images.
Similar to~\cite{liu_mtan_2019}, we re-size the images to 288x384 pixels.
Training is run for 200 epochs with a batch size of 8.
We apply the same data augmentation and learning rate schedule as for Cityscapes.

\paragraph{CelebA}~\cite{liu_celeba_2015}
We re-size images to 64x64 pixels as done in \cite{lin_rlw_2022} and consider the original split of 162,770/19,867/19,962 for training, validation, and testing.
We set the batch size to 512, train for 100 epochs, and halve the learning rates every 30 epochs.

\begin{table}[ht]
    \centering
    \tiny
    \setlength{\tabcolsep}{1pt}  
    \setlength{\extrarowheight}{.5em}
    \caption{
    \textbf{Original experiment setup as reported in respective papers.}
    We note a high variation regarding the choice of network, optimizer, and other hyper-parameters among the different works.
    }
    \begin{tabular}{cp{1.4cm}p{2.3cm}p{1.8cm}p{2,7cm}p{1.0cm}p{0.7cm}p{1.2cm}}
    \toprule    
        Data & MTO & Network & Optimizer & learning rate & weight decay & batch size & \#train. \newline iterations \\
    \midrule
        \rotatebox{90}{\hspace{-53pt}Cityscapes~\cite{cordts_cityscapes_2016}} & 
        UW \newline~\cite{kendall_uw_2018} & DeepLabV3~\cite{chen_deeplabv3_2018}\newline with ResNet101~\cite{yu_dilated-resnet_2017} & SGD + Nesterov updates, Mom. & init.: $2.5\cdot10^{-3}$; \newline polynomial lr decay & $1\cdot 10^{-4}$ & 8 & 100k iter. 
        \\
        & RLW \newline~\cite{lin_rlw_2022} & DeepLabV3~\cite{chen_deeplabv3_2018}\newline with ResNet50~\cite{yu_dilated-resnet_2017} & Adam & $1\cdot 10^{-4}$ & $1\cdot 10^{-5}$ & 64 & \\
        & IMTL \newline~\cite{liu_imtl_2021} & ResNet50~\cite{yu_dilated-resnet_2017}\newline + PSPNet~\cite{zhao_pspnet_17} heads & SGD+Mom. &  init.: $0.02$; \newline polynomial lr decay & $1\cdot 10^{-4}$& 32 & 200 epochs
        \\
        & PCGrad \newline~\cite{yu_pcgrad_2020} & MTAN~\cite{liu_mtan_2019} & Adam & init.: $1\cdot10{-4}$; \newline halve lr after 40k iter. & - & 8 & 80k iter \\
        & CAGrad \newline~\cite{liu_cagrad_2021} & MTAN~\cite{liu_mtan_2019} &Adam & init.: $1\cdot10{-4}$;\newline halve lr every 100 epochs & - & 8 & 200 epochs\\
        & AlignedMTL ~\cite{senushkin_alignedmtl_2023} & MTAN~\cite{liu_mtan_2019} / \newline PSPNet~\cite{zhao_pspnet_17} &Adam & init.: $1\cdot10{-4}$;\newline halve lr every 100 epochs & - & 8 & 200 epochs\\
    \midrule
        \rotatebox{90}{\hspace{-40pt}NYUv2~\cite{silberman_nyuv2_2012}}
        & RLW \newline~\cite{lin_rlw_2022} & DeepLabV3~\cite{chen_deeplabv3_2018}\newline with ResNet50~\cite{yu_dilated-resnet_2017} & Adam & $1\cdot 10^{-4}$ & $1\cdot 10^{-5}$ & 8 & \\
        & IMTL \newline~\cite{liu_imtl_2021} & ResNet50~\cite{yu_dilated-resnet_2017}\newline + PSPNet~\cite{zhao_pspnet_17} heads & SGD+Mom. &  init.: $0.03$ & - & 48 & 200 epochs \\
        & PCGrad \newline~\cite{yu_pcgrad_2020} & MTAN~\cite{liu_mtan_2019} & Adam & init.: $1\cdot10{-4}$; \newline halve lr after 40k iter. & - & 2 & 80k iter \\
        & CAGrad \newline~\cite{liu_cagrad_2021} & MTAN~\cite{liu_mtan_2019} & Adam & init.: $1\cdot10{-4}$; \newline halve lr after 100 epochs & - & 2 & 200 epochs\\
        & AlignedMTL \newline~\cite{senushkin_alignedmtl_2023}  & MTAN~\cite{liu_mtan_2019} / \newline PSPNet~\cite{zhao_pspnet_17} & Adam & init.: $1\cdot10{-4}$; \newline halve lr after 100 epochs & - & 2 & 200 epochs\\
    \midrule
        \rotatebox{90}{\hspace{-30pt}CelebA~\cite{liu_celeba_2015}}
        & RLW \newline~\cite{lin_rlw_2022} & ResNet17~\cite{he_resnet_16}\newline + lin. classifier & Adam & $1\cdot 10^{-3}$ & - & 512\\
        & IMTL \newline~\cite{liu_imtl_2021} & ResNet17~\cite{he_resnet_16}\newline + lin. classifier & Adam & $0.003$ & - & 256 & 100 epochs \\
        & PCGrad \newline~\cite{yu_pcgrad_2020} & ResNet17~\cite{he_resnet_16}\newline + lin. classifier  & Adam & init. from $\{10^{-4}, ..., 5\cdot10^{-2}\}$; \newline halve lr every 30 epochs & - & 256 & 100 epochs\\
    \bottomrule
    \end{tabular}
\label{tab:rel_work}
\end{table}

\subsection{Training}
\label{sec:appx_training_details}

\paragraph{Effectiveness of Adam in MTL.}
\label{sec:appx_impl_details_adamsgd}

All presented results are based on performing early stopping w.r.t. $\Delta_m$ metric on the validation set.
For this, we further trained single-task learning (STL) models for each experiment combination (dataset and network) using the respective network architecture except for the missing head(s).
We trained the models using Adam and any learning rate from $\{0.01, 0.005, ..., 0.00005\}$. 
The training was stopped early based on the validation loss.
Reported scores in \cref{tab:res_nyu_segnet,tab:res_nyu_deeplab,tab:res_cs_segnet,tab:res_cs_deeplab} are computed as the mean of the models' performance that were initialized with the three different seeds. 

Our implementation for all experiments is based on the LibMTL library \cite{lin_libmtl_2022}.

\paragraph{Gradient Similarity.}
\label{sec:appx_impl_details_gc}
Our gradient similarity experiments were conducted on the best performing hyper-parameter configuration for EW from the previous extensive evaluation.
Over the full training, gradient similarity measures are computed every five iteration steps and summarized per epoch.
To make the computation effort more feasible in case of settings with large batch size or high number of tasks, we randomly select eight samples or tasks respectively and consider corresponding gradients in these cases.

\subsection{Loss functions}

\paragraph{Cityscapes}
For the task of semantic segmentation, we employ a pixel-wise cross-entropy loss:
\begin{equation}
    \mathcal{L}_{CE} = \sum_{c=1}^C y^c\cdot \log (p^c)
\end{equation}
where $C$ is the number of classes, $y^c\in \{0,1\}$ indicates the ground truth class, and $p^c$ is the predicted probability  for class $c$ which results from computing the softmax for output logits $z^c$, $p^c=\frac{\exp(z^c)}{\sum_{c=1}^C \exp(z^c)}$. This loss is averaged over the image.

For depth estimation, we utilize the $L_1$ loss:
\begin{equation}
    \mathcal{L}_{depth} = \|\bm{y}-\hat{\bm{y}}\|_1
\end{equation}
where $\bm{y}, \hat{\bm{y}}$ indicate ground truth and prediction, respectively. Pixels with invalid depth value in the ground truth data are ignored.
It is noteworthy that these two types of losses are not balanced when used directly without modification.

\paragraph{NYUv2}
The tasks of semantic segmentation and depth estimation are trained using the same loss functions as described for Cityscapes.
In addition, we compute the cosine loss on the (normalized) surface normal maps:
\begin{equation}
    \mathcal{L}_{normal} = 1-\cos{\theta} = 1-\frac{\bm{y}\cdot\hat{\bm{y}}}{\|\bm{y}\|\|\hat{\bm{y}}\|}
\end{equation}
where $\bm{y}, \hat{\bm{y}}$ are the ground truth and predicted normal maps.
Similar to the Cityscapes setup, the combination of these loss functions is not balanced per default.

\paragraph{CelebA}
To learn to predict multiple attributes, we use a binary cross entropy loss for the individual classes:
\begin{equation}
    \mathcal{L}_{CE,bin} = -[y\log (p) +(1-y)\log (1-p)]
\end{equation}
Although all these losses have a similar scale, their impact varies based on the difficulty of the individual tasks and the number of available samples displaying the respective attribute.

\subsection{Evaluation criteria}

In this study, we primarily focus on Pareto optimal solution to acknowledge that different configurations may lead to varying preferences for the learned tasks.
However, it is important to note that not every point on the Pareto front is relevant in practice, especially when one metric significantly dominates while others are close to chance level. 
Moreover, specific real-world applications can have a stronger, pre-defined prioritization of one or a few sub-tasks which requires a relative weighting of the tasks' performances.

Additionally, we further employ the $\Delta_m$ metric which offers a simple option to directly compare the performance of two models using a single scalar. This metric further indicates the relative performance compared to the single-tasks models.

Note that for our initial toy task experiment (\cref{sec:exp_adam_sgd}), we consider the original setting from \cite{liu_cagrad_2021} which optimzes the global minimum of the two loss functions.

\section{Additional results on comparison between Adam and SGD}
\label{sec:additional_adam_sgd}

\begin{figure}[ht]
\centering
\scriptsize
\begin{tabular}{cccc}
    \includegraphics[width=0.24\textwidth]{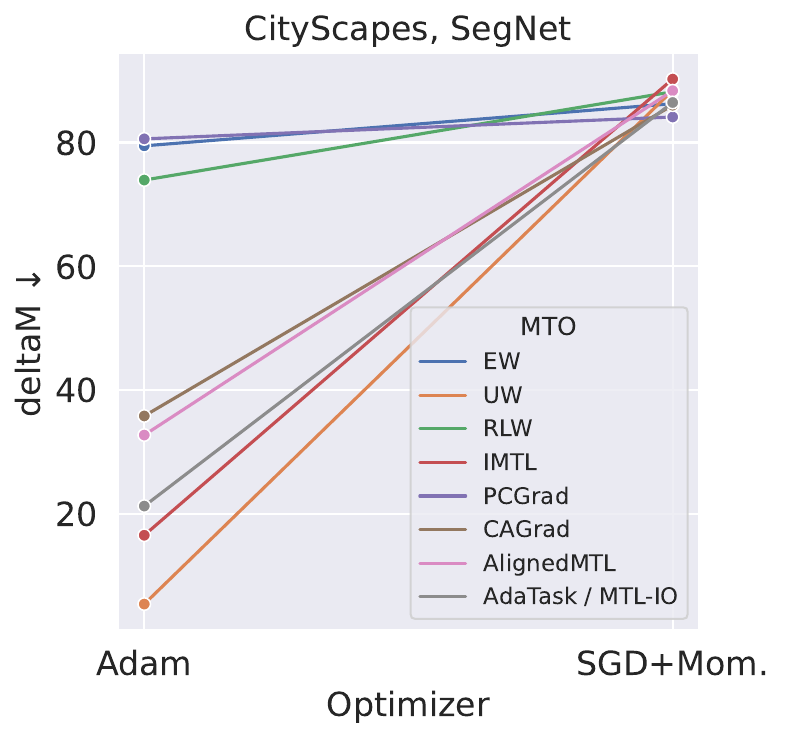} & 
    \includegraphics[width=0.24\textwidth]{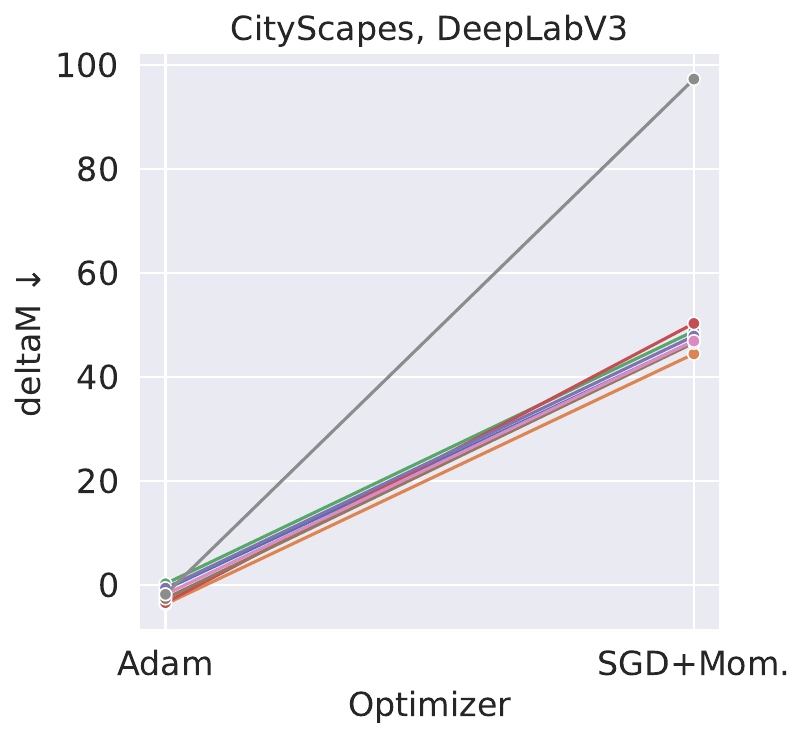} & 
    \includegraphics[width=0.24\textwidth]{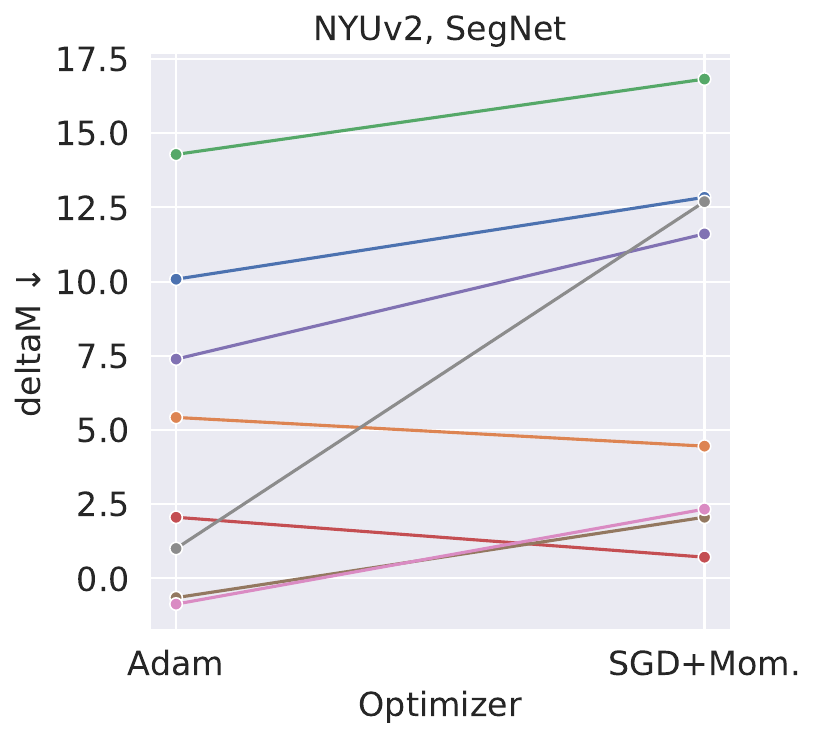} & 
    \includegraphics[width=0.24\textwidth]{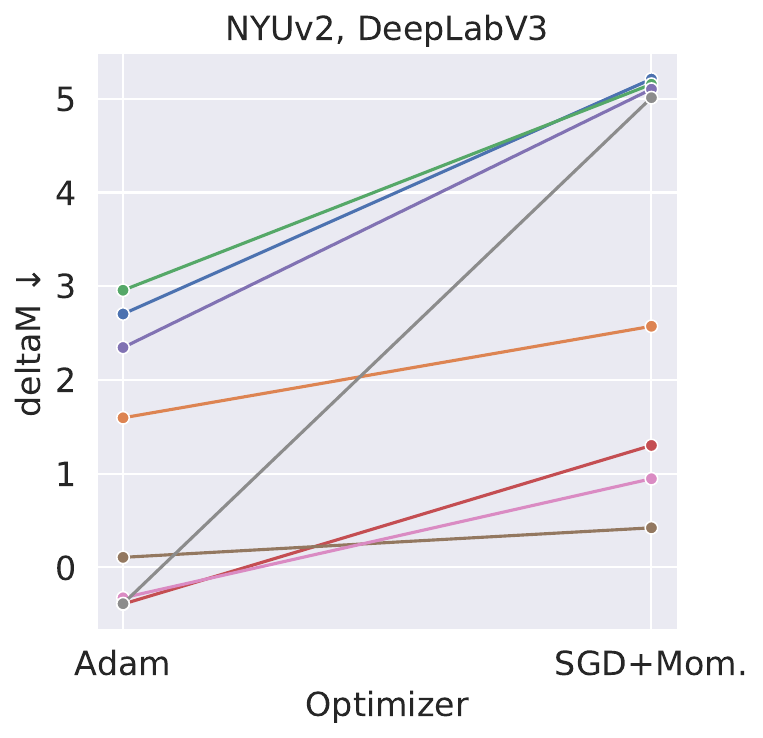} \\
\end{tabular}
\caption{ 
    \textbf{Mean $\Delta_m$ metric for experiments run on Cityscapes and NYUv2 with SegNet and DeepLabV3.}
    We compare the performance of the best hyperparameter setting for every MTO method using either Adam (left) or SGD+Momentum (right) (lower is better).
    Every MTO is associated with a different line color/style.
    On Cityscapes, there is a large difference for the $\Delta_m$ score for Adam compared to SGD+Momentum, especially for UW, IMTL, and CAGrad. Therefore, for this setup, the result depends more on the optimizer than on the MTO method. 
    On the NYUv2 dataset this observation weakens. Adam still achieves the lowest $\Delta_m$ scores across different MTO methods (except for SegNet with UW and IMTL), though, besides chosing Adam, it is also important to select the appropriate MTO method. 
    }
    \label{fig:deltam_over_datasets_networks}
\end{figure}

\begin{figure*}[ht]
     \centering
     \begin{subfigure}[b]{\textwidth}
         \centering
         \setlength{\belowcaptionskip}{0pt}
         \includegraphics[trim = 10mm 19mm 10mm 10mm,clip, width=\textwidth]{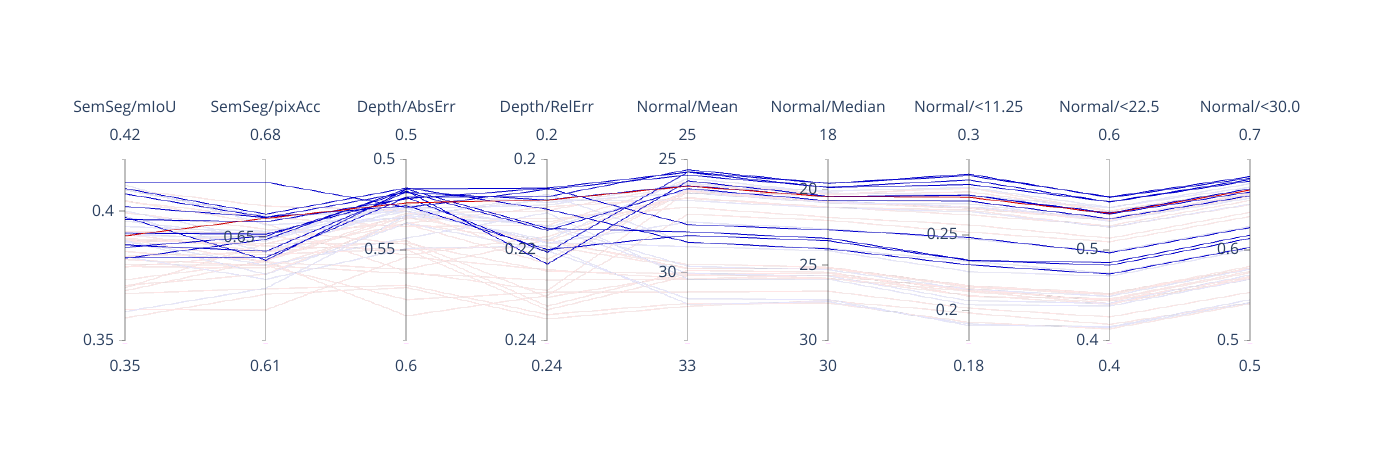}
         \caption{SegNet}
         \label{fig:pcp_nyu_segnet}
     \end{subfigure}   
    \hfill
     \\
     \begin{subfigure}[b]{\textwidth}
         \centering\setlength{\belowcaptionskip}{0pt}
         \includegraphics[trim = 10mm 19mm 10mm 10mm,clip, width=\textwidth]{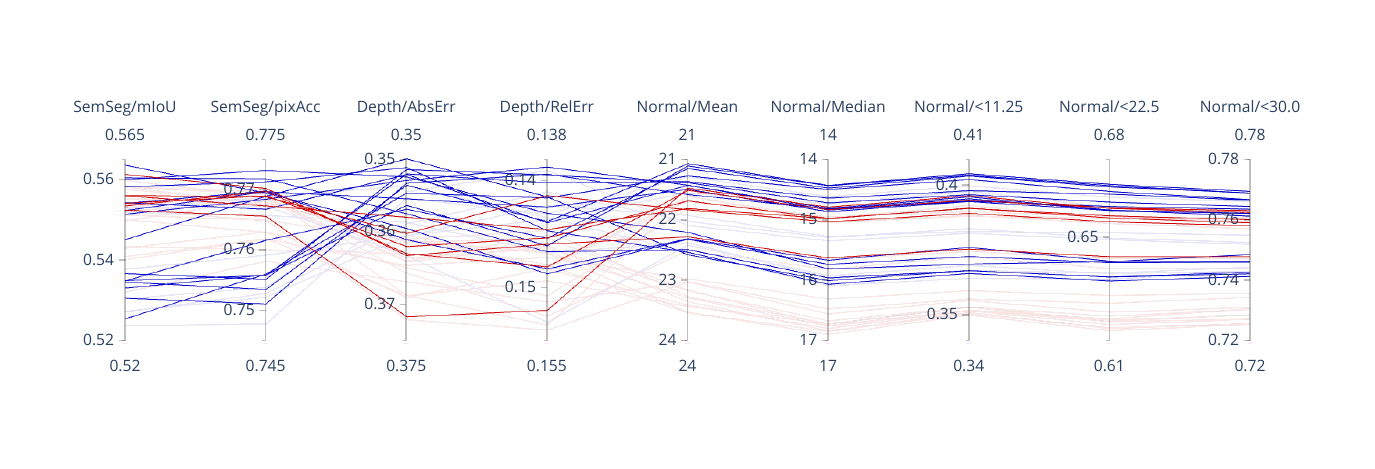}
         \caption{DeepLabV3}
         \label{fig:pcp_nyu_deeplabv3}
     \end{subfigure}
     \\
    \caption{
    \textbf{Parallel coordinate plot over all experiments on NYUv2}
    We distinguish between experiments using {\color{red}SGD+mom} and {\color{blue}Adam} optimizer. 
    Experiments that reached Pareto front performance are drawn with higher saturation.
    \\
    Similiar to results on Cityscapes in the main paper (\cref{fig:adam_vs_sgd_pcp}) we observe a dominance of Adam albeit, here, we also have some experiments using SGD+Mom. on the overall Pareto front.
    }
    \label{fig:adam_vs_sgd_pcp_v2}
\end{figure*}

We present additional evaluation results for our comparison between optimizers for MTL.
In \Cref{fig:deltam_over_datasets_networks}, we compare the $\Delta_m$ metric performance between the usage of Adam and SGD+mom.
\cref{fig:adam_vs_sgd_pcp_v2} shows additional parallel coordinate plots for NYUv2 and both choices of networks.
In \Cref{tab:adam_vs_sgd_po_exp_count}, we count for each used MTO method the number of experiment runs that are located on the Pareto front w.r.t. each setup.
Best performing quantitative results for all MTOs can be found in \cref{tab:res_cs_segnet,tab:res_cs_deeplab,tab:res_nyu_segnet,tab:res_nyu_deeplab}.

We further show extended results on the toy task by Liu et al.~\cite{liu_cagrad_2021} for more learning rates in \cref{tab:cagrad_toytask_niter_full}.


\begin{table*}[ht]
    \centering
    \tiny
    \newcommand{\gcs}{\hspace{10pt}}
    \newcommand{\customdashline}{\noalign{\vskip 0.5ex}\hdashline\noalign{\vskip 0.5ex}}
    \caption{
        \textbf{Count of Pareto optimal experiments for each MTO method.}
        We found no single MTO method to be clearly superior over all combinations of dataset and networks.
        Total numbers can be compared to \Cref{tab:adam_vs_sgd_pareto}
    }
    \begin{tabular}{cccc@{\gcs}ccccccccc@{\gcs}c}
        \toprule
        Data & Network & Optimizer && EW & UW & RLW & IMTL & PCGrad & CAGrad & AlignedMTL & AdaTask && Total\\
        \midrule
        Cityscapes & SegNet & Adam     && - & 3 & - & - & - & 1 & 1 & - && 5\\
        Cityscapes & SegNet & SGD+Mom. && - & - & - & - & - & - & - & - && -\\
        Cityscapes & DeepLabV3 & Adam     && 2 & 2 & 1 & 2 & 2 & 1 & - & - && 15\\
        Cityscapes & DeepLabV3 & SGD+Mom. && - & - & - & - & - & - & - & - && -\\
        \midrule
        NYUv2 & SegNet & Adam     && 1 & 2 & - & 1 & 1 & 1 & 3 & 2 && 11\\
        NYUv2 & SegNet & SGD+Mom. && - & - & - & 1 & - & - & - & - && 1\\
        NYUv2 & DeepLabV3 & Adam     && 2 & - & 1 & 3 & 3 & 2 & 2 & 3 && 16\\
        NYUv2 & DeepLabV3 & SGD+Mom. && - & 1 & - & - & - & 3 & 2 & - && 6\\
        \bottomrule
    \end{tabular}
    \label{tab:adam_vs_sgd_po_exp_count}
\end{table*}

\begin{table}
\centering
\scriptsize
\setlength{\tabcolsep}{2pt}  
\newcommand{\gcs}{\hspace{4pt}}
\caption{
\textbf{Number of iterations after which all seeds in toy task experiment from CAGrad~\cite{liu_cagrad_2021} have reached the global minimum} for different learning rates and optimizer.
We show results for additional learning rates compared to the main paper.
The maximum iteration number over all three seeds for each MTO method / learning rate / optimizer combination is reported. 
If not all seeds converged to the global minimum within 100k iteration steps, we denote it as '-'.
In several setups, EW+Adam converges fastest to the global minimum.
Especially for small learning rates, CAGrad performs advantageous compared to EW.
As reported in previous work, we found that PCGrad often would converge only to some point on the Pareto Front.
The \textbf{best} and \textit{second best} run for each learning rate over all MTO methods are indicated via font type. 
}
\begin{tabular}{cc@{\gcs}cc@{\gcs}ccccccccc}
    \toprule
   &&&& \multicolumn{9}{c}{learning rate}\\
   \cmidrule{5-13}
    && method && 10.0 & 5.0 & 1.0 & 0.5 & 0.1 & 0.05 & 0.01 & 0.005 & 0.001* \\
    \midrule
    {\multirow{3}{*}{\rotatebox{90}{\hspace{0pt} GD}}} && EW && - & 103 & - & - & - & - & - & - & -\\
    && PCGrad && - & - & - & - & - & - & - & - & -\\
    && CAGrad && 644 & - &213 &  621 & 8,069 & 5,732 & 20,418 & 34,405 & - \\
    \midrule
    {\multirow{3}{*}{\rotatebox{90}{\hspace{0pt} Adam}}} && EW && \textit{26} &  \textit{37} & \textbf{22} &\textbf{58} & \textbf{709} & \textbf{2,135} & \textbf{9,015} & \textit{16,005} & - \\
    && PCGrad && \textbf{25} & 4,960& 56 & 15,741 & 34,175 & 41,438 & - & - & - \\
    && CAGrad &&\cellcolor{green!25}  27 & \textbf{30} & \textit{32} &  \textit{106} & \textit{802} & \textit{7,109} & \textit{11,239} & \textbf{14,323} & \textbf{57,700} \\
    \bottomrule\noalign{\vskip 0.5ex}
    \multicolumn{9}{l}{\footnotesize{*LR used for results in~\cite{liu_cagrad_2021} with Adam}}
\end{tabular}
\label{tab:cagrad_toytask_niter_full}
\end{table}

\clearpage
\begin{table}[!ht]
    \centering
    \tiny
    \newcommand{\gcs}{\hspace{12pt}}
    \newcommand{\customdashline}{\noalign{\vskip 0.5ex}\hdashline\noalign{\vskip 0.5ex}}
    \caption{
    \textbf{Results for different MTO methods and optimizers on Cityscapes\cite{cordts_cityscapes_2016} using SegNet~\cite{badrinarayanan_segnet_2017}.}
    The best score per metric is highlighted \textbf{for each MTO method} as well as \underline{over all methods and optimizers}.
    While different MTO methods perform best over the distinct metrics, models trained with Adam outperform those based on SGD+mom in most direct comparisons.
    On the overall $\Delta_m$ metric, Adam shows superior performance for all MTO methods, in some cases even with a high margin.
    Best performance for each metric was also achieved by using Adam.
    }
    \begin{tabular}{cccc@{\gcs}ccc@{\gcs}ccc@{\gcs}c}
    \toprule    
        &&&& \multicolumn{2}{c}{Sem.Seg.} && \multicolumn{2}{c}{Depth}\\
        \cmidrule(r){5-6} \cmidrule(r){8-9}
        MTO & Optimizer & lr && mIoU $\uparrow$ & pixAcc $\uparrow$  && AbsErr $\downarrow$  & RelErr $\downarrow$ && DeltaM $\downarrow$\\
    \midrule
        STL & adam & && 0.7122 & 0.9221 && 0.0134 & 29.88 \\
    \midrule
        EW & adam & 0.005 &&     0.6898 & 0.9165 && \textbf{0.0196} & \textbf{109.84} && \textbf{79.43} \std{3.68}\\ 
        EW & signSGD & 0.001  && \textbf{0.7013} & 0.9174 && 0.0210 & 115.67 && 86.52 \std{6.06}\\
        EW & sgd & 0.1 &&        0.6967 & \textbf{0.9179} && 0.0216 & 113.82 && 86.24 \std{1.97}\\ \customdashline
        UW & adam & 0.001 &&     \textbf{0.7052} & \textbf{0.9202} && \textbf{\underline{0.0136}} & \textbf{\underline{35.69}} && ~\textbf{\underline{5.44}} \std{2.38}\\ 
        UW & signSGD & 0.001  && 0.6506 & 0.8997 && 0.0166 & 53.61 && 28.54 \std{7.83}\\
        UW & sgd & 0.01 &&       0.6750 & 0.9110 && 0.0219 & 114.67 && 88.39 \std{1.35}\\ \customdashline
        RLW & adam & 0.001 &&    \textbf{0.7013} & \textbf{0.9196} && \textbf{0.0197} & \textbf{103.61} && \textbf{73.91} \std{7.63}\\ 
        RLW & signSGD & 0.0005  && 0.6962 & 0.9169 && 0.0204 & 112.46 && 82.90 \std{5.32}\\
        RLW & sgd & 0.1 &&       0.6918 & 0.9156 && 0.0227 & 113.59 && 88.16 \std{0.67}\\ \customdashline
        IMTL & adam & 0.005 &&   \textbf{0.6963} & \textbf{0.9170} && \textbf{0.0148} & \textbf{45.63} && \textbf{16.55} \std{1.52}\\ 
        IMTL & signSGD & 0.005  && 0.6659 & 0.9070 && 0.0197 & 101.90 && 74.03 \std{14.62}\\
        IMTL & sgd & 0.01 &&     0.6716 & 0.9107 && 0.0230 & 114.38 && 90.21 \std{0.74}\\ \customdashline
        PCGrad & adam & 0.01 &&  0.6770 & 0.9135 && 0.0226 & \textbf{103.88} && \textbf{80.56} \std{3.70}\\ 
        PCGrad & signSGD & 0.001  && 0.6929 & 0.9170 && \textbf{0.0210} & 113.32 && 84.74 \std{3.46}\\
        PCGrad & sgd & 0.1 &&    \textbf{0.6972} & \textbf{0.9176} && 0.0235 & 107.06 && 84.09 \std{0.93}\\ \customdashline
        CAGrad & adam & 0.001 && \textbf{0.7088} & \textbf{0.9208} && \textbf{0.0162} & \textbf{66.39} && \textbf{35.81} \std{14.91}\\ 
        CAGrad & signSGD & 0.001  && 0.6883 & 0.9138 && 0.0182 & 107.06 && 74.68 \std{18.91}\\
        CAGrad & sgd & 0.1 &&    0.6896 & 0.9156 && 0.0205 & 115.52 && 85.88 \std{0.31} \\ \customdashline
        AlignedMTL & adam & 0.0005 && \textbf{\underline{0.7164}} & \textbf{\underline{0.9246}} && \textbf{0.0154} & \textbf{64.89} && \textbf{32.76} \std{5.27}\\ 
        AlignedMTL & signSGD & 0.001 && 0.7010 & 0.9189 && 0.0180 & 102.41 && 69.77 \std{24.91} \\
        AlignedMTL & sgd & 0.1 && 0.6684 & 0.9103 && 0.0224 & 113.11 && 88.37 \std{0.53} \\ \customdashline
        Adatask & adam & 0.0005 && \textbf{0.7039} & \textbf{0.9196} && \textbf{0.0146} & \textbf{52.17} && \textbf{21.28} \std{0.35}\\ 
        MTL-IO & sgd & 0.1 &&    0.6957 & 0.9172 && 0.0233 & 110.19 && 86.46 \std{1.63} \\ 
    \bottomrule
    \end{tabular}
\label{tab:res_cs_segnet}

    \smallskip
    
    \caption{
    \textbf{Results for different MTO methods and optimizers on Cityscapes~\cite{cordts_cityscapes_2016} using DeepLabV3+~\cite{chen_deeplabv3_2018}.}
    The best score per metric is highlighted \textbf{for each MTO method} as well as \underline{over all methods and optimizers}.
    Adam is Pareto dominant over SGD+mom in a direct pairwise comparison across all MTO methods. Here, we would also like to highlight that MTL can outperform STL as suggested by \cite{caruana1997multitask}. 
    }
    \begin{tabular}{cccc@{\gcs}ccc@{\gcs}ccc@{\gcs}c}
    \toprule    
        &&&& \multicolumn{2}{c}{Sem.Seg.} && \multicolumn{2}{c}{Depth}\\
        \cmidrule(r){5-6} \cmidrule(r){8-9}
        MTO & Optimizer & lr && mIoU $\uparrow$ & pixAcc $\uparrow$  && AbsErr $\downarrow$  & RelErr $\downarrow$ && DeltaM $\downarrow$\\
    \midrule
        STL & adam & && 0.7203 & 0.9253 && 0.0132 & 47.37 \\
    \midrule
        EW & adam & 0.001 &&     \textbf{\underline{0.7247}} & \textbf{0.9268} && \textbf{0.0128} & \textbf{47.73}  && \textbf{-0.80} \std{0.69}\\ 
        EW & signSGD & 0.0005  && 0.7188 & 0.9249 && 0.0134 & 52.47 && 3.07 \std{2.70}\\
        EW & sgd & 0.05 &&       0.7100 & 0.9217 && 0.0174 & 120.34 && 46.88 \std{2.02}\\ \customdashline
        UW & adam & 0.001 &&     \textbf{0.7224} & \textbf{0.9259} && \textbf{0.0122} & \textbf{\underline{44.37}} && \textbf{\underline{-3.65}}\std{0.61}\\ 
        UW & signSGD & 0.001  && 0.7172 & 0.9244 && 0.0127 & 48.31 && -0.35 \std{2.23}\\
        UW & sgd & 0.005 & &     0.7003 & 0.9187 && 0.0171 & 116.09 && 44.47 \std{1.91}\\ \customdashline
        RLW & adam & 0.001 &&    \textbf{0.7230} & \textbf{0.9263} && \textbf{0.0133} & \textbf{47.76} && \textbf{0.26} \std{1.08} \\ 
        RLW & signSGD & 0.001  && 0.7185 & 0.9244 && \textbf{0.0133} & 54.75 && 4.09 \std{0.39}\\
        RLW & sgd & 0.05 &&         0.7070 & 0.9205 && 0.0176 & 123.13 && 48.84 \std{1.79} \\ \customdashline
        IMTL & adam & 0.001 &&   \textbf{0.7226} & \textbf{0.9259} && \textbf{\underline{0.0121}} & \textbf{45.25} && \textbf{-3.38} \std{0.92}\\ 
        IMTL & signSGD & 0.001  && 0.7177 & 0.9243 && 0.0128 & 49.61 && 0.57 \std{3.75}\\
        IMTL & sgd & 0.005 &&    0.7027 & 0.9192 && 0.0185 & 122.37 && 50.33 \std{3.30}\\ \customdashline
        PCGrad & adam & 0.001 && \textbf{\underline{0.7247}} & \textbf{\underline{0.9272}} && \textbf{0.0130} & \textbf{47.24} && \textbf{-0.58} \std{0.56}\\ 
        PCGrad & signSGD & 0.001  && 0.7187 & 0.9248 && 0.0132 & 53.35 && 3.28 \std{1.41}\\
        PCGrad & sgd & 0.05 &&   0.7083 & 0.9212 && 0.0173 & 122.52 && 47.90 \std{3.68}\\ \customdashline
        CAGrad & adam & 0.001 && \textbf{0.7245} & \textbf{0.9264} && \textbf{0.0124} & \textbf{45.56} && \textbf{-2.59} \std{0.68}\\ 
        CAGrad & signSGD & 0.001  && 0.7154 & 0.9243 && 0.0128 & 49.06 && 0.37 \std{0.41}\\
        CAGrad & sgd & 0.1 &&    0.7096 & 0.9220 && 0.0172 & 120.24 && 46.58 \std{2.21}\\ \customdashline
        AlignedMTL & adam & 0.0005 && \textbf{0.7225} & \textbf{0.9259} && \textbf{0.0122} & 47.70 && \textbf{-1.79} \std{0.70} \\ 
        AlignedMTL & signsgd & 0.001 && 0.7169 & 0.9241 && 0.0127 & \textbf{47.56} && -0.70 \std{0.61}
 \\
        AlignedMTL & sgd & 0.1 && 0.7096 & 0.9218 && 0.0174 & 120.38 && 46.94 \std{1.19}
 \\ \customdashline
        
        Adatask & adam & 0.001 && \textbf{0.7234} & \textbf{0.9260 } && \textbf{0.0124} &\textbf{47.21} && \textbf{--1.77} \std{0.86}\\
        MTL-IO & sgd & 0.05 && 0.7098 & 0.9217 && 0.0187 & 120.61 && 49.57  \std{2.28}\\
    \bottomrule
    \end{tabular}
\label{tab:res_cs_deeplab}
\end{table}

\begin{table}[!ht]
    \centering
    \tiny
    \setlength{\tabcolsep}{0pt}  
    \newcommand{\gcs}{\hspace{2pt}}
    \newcommand{\customdashline}{\noalign{\vskip 0.5ex}\hdashline\noalign{\vskip 0.5ex}}
        \caption{
    \textbf{Results for different MTO methods and optimizers on NYUv2\cite{silberman_nyuv2_2012} using SegNet~\cite{badrinarayanan_segnet_2017}.}
    The best score per metric is highlighted \textbf{for each MTO method} as well as \underline{over all methods and optimizers}.
    Using Adam yields in superior performance in the majority of cases, both when considering the individual metrics and the overall $\Delta_m$ metric.
    We note that $\Delta_m$ is more effected by the normal task due to the higher number of corresponding metrics as can be observed in the case of UW.
    }
    \begin{tabular}{cccc@{\gcs}ccc@{\gcs}ccc@{\gcs}cccccc@{\gcs}c}
    \toprule    
        &&&& \multicolumn{2}{c}{Sem.Seg.} && \multicolumn{2}{c}{Depth}  && \multicolumn{5}{c}{Normal}\\
        \cmidrule(r){5-6} \cmidrule(r){8-9} \cmidrule(r){11-15}
        MTO & Optimizer & lr && mIoU$\uparrow$ & pixAcc$\uparrow$  && AbsErr$\downarrow$  & RelErr$\downarrow$ && Mean$\downarrow$ & Median$\downarrow$ & $<11.25\uparrow$ & $<22.5\uparrow$ & $<30.0\uparrow$ && DeltaM$\downarrow$\\
    \midrule
        STL & adam &  &&       0.392  & 0.646 && 0.607 & 0.258 && 24.74 & 18.49 & 0.308 & 0.582 & 0.700\\ 
    \midrule
        EW & adam & 0.0001 &&        \textbf{0.398} & \textbf{0.650} && \textbf{0.530} & \textbf{0.212} && \textbf{29.53} & \textbf{25.02} & \textbf{0.216} & \textbf{0.454} & \textbf{0.584} && \textbf{10.08} \std{2.84}\\ 
        EW & signSGD & 0.0001    && 0.389 & 0.647 && 0.543 & 0.220 && 30.18 & 25.93 & 0.204 & 0.438 & 0.569 && 12.75 \std{2.24}\\
        EW & sgd & 0.01 &&           0.384 & 0.644 && 0.551 & 0.227 && 30.12 & 25.72 & 0.212 & 0.443 & 0.571 && 12.84 \std{0.65}\\ \hdashline
        UW & adam & 0.0001 &&        \textbf{0.401}& \textbf{0.652} && \textbf{0.522} & \textbf{0.213} && 27.93 & 22.80 & 0.243 & 0.494 & 0.623 && 5.42 \std{1.04}\\ 
        UW & signSGD & 0.0001    && 0.384 & 0.634 && 0.535 & 0.218 && 28.36 & 23.45 & 0.233 & 0.482 & 0.612 && 8.05 \std{1.18}\\
        UW & sgd & 0.05 &&           0.383 & 0.642 && 0.551 & 0.218 && \textbf{27.07} & \textbf{21.66} & \textbf{0.259} & \textbf{0.516} & \textbf{0.644} && \textbf{4.46} \std{2.13}\\ \hdashline
        RLW & adam & 0.0001 &&       \textbf{0.390} & \textbf{0.638} && \textbf{0.537} & \textbf{0.218} && 30.83 & 26.78 & 0.196 & 0.424 & 0.554 && 14.28 \std{2.38}\\ 
        RLW & signSGD & 0.0001    && 0.375 & 0.636 && 0.544 & 0.226 && \textbf{29.38} & \textbf{24.77} & \textbf{0.217} & \textbf{0.458} & \textbf{0.589} && \textbf{11.42} \std{1.02}\\
        RLW & sgd & 0.05 &&          0.369 & 0.633 && 0.572 & 0.233 && 31.21 & 27.17 & 0.198 & 0.420 & 0.546 && 16.82 \std{1.16}\\ \hdashline
        IMTL & adam & 0.0001 &&      0.380 & 0.644 && \textbf{0.524} & 0.216 && 26.35 & 20.77 & 0.270 & 0.534 & 0.660 && 2.06 \std{1.31}\\ 
        IMTL & signSGD & 0.0001    && 0.374 & 0.639 && 0.527 & \textbf{0.212} && 26.26 & 20.65 & 0.272 & 0.537 & 0.663 && 1.97 \std{0.832}\\
        IMTL & sgd & 0.05 &&         \textbf{0.396} & \textbf{0.656} && 0.532 & 0.215 && \textbf{26.16} & \textbf{20.38} & \textbf{0.277} & \textbf{0.542} & \textbf{0.666} && \textbf{0.72} \std{0.68}\\ \hdashline
        PCGrad & adam & 0.0001 &&    \textbf{\underline{0.406}} & \textbf{0.654} && \textbf{0.529} & \textbf{0.215} && \textbf{28.60} & \textbf{23.75} & \textbf{0.231} & \textbf{0.477} & \textbf{0.606} && \textbf{7.39}\std{0.86}\\ 
        PCGrad & signSGD & 0.0001    && 0.394 & 0.649 && 0.556 & 0.219 && 28.98 & 24.25 & 0.223 & 0.467 & 0.598 && 9.59 \std{1.02}\\
        PCGrad & sgd & 0.01 &&       0.389 & 0.644 && 0.547 & 0.223 && 29.76 & 25.22 & 0.214 & 0.451 & 0.580 && 11.61\std{0.36}\\ \hdashline
        CAGrad & adam & 0.0001 &&    \textbf{0.405} & \textbf{\underline{0.661}} && \textbf{0.527} & \textbf{0.211} && \textbf{25.85} & \textbf{20.02} & \textbf{0.282} & \textbf{0.550} & \textbf{0.674} && \textbf{-0.65} \std{0.96}\\ 
        CAGrad & signSGD & 0.0001    && 0.393 & 0.648 && 0.546 & 0.217 && 26.21 & 20.58 & 0.272 & 0.538 & 0.665 && 1.67 \std{0.61}\\
        CAGrad & sgd & 0.05 &&       0.400 & 0.656 && 0.547 & 0.223 && 26.39 & 20.78 & 0.268 & 0.534 & 0.662 && 2.06 \std{0.65}\\   \hdashline
        AlignedMTL & adam & 0.0001 && \textbf{0.385} & \textbf{0.648} && \textbf{0.519} & 0.212 && \textbf{\underline{25.51}} & \textbf{\underline{19.66}} & \textbf{\underline{0.288}} & \textbf{\underline{0.557}}	& \textbf{\underline{0.680}} && \textbf{\underline{-0.86}} \std{0.14} \\ 
        AlignedMTL & signSGD & 0.0001
 && 0.363 & 0.637 && 0.528 & \textbf{\underline{0.211}} && 25.86 & 20.02 & 0.282 & 0.550 & 0.674 && 0.90 \std{0.54} \\ 
        AlignedMTL & sgd & 0.05 && 0.371 & 0.642 && 0.531 & 0.214 && 26.41 & 20.71 & 0.272 & 0.535 & 0.661 && 2.33 \std{1.00} \\ \hdashline
        adatask & adam & 0.0001 &&       \textbf{0.394} & \textbf{0.650} && \textbf{\underline{0.518}} & \textbf{0.211} && \textbf{26.29} &  \textbf{20.67} & \textbf{0.273} & \textbf{0.536} & \textbf{0.661} && \textbf{1.00}\std{0.8} \\ 
        MTL-IO & sgd & 0.01 &&          0.377 & 0.637 && 0.551 & 0.225 && 30.05 & 25.50 & 0.215 & 0.447 & 0.574 && 12.69\std{1.21} \\
    \bottomrule
    \end{tabular}
\label{tab:res_nyu_segnet}
\end{table}
\begin{table}[!ht]
    \centering
    \tiny
    \setlength{\tabcolsep}{0pt}  
    \newcommand{\gcs}{\hspace{2pt}}
    \newcommand{\customdashline}{\noalign{\vskip 0.5ex}\hdashline\noalign{\vskip 0.5ex}}
    \caption{
    \textbf{Results for different MTO methods and optimizers on NYUv2\cite{silberman_nyuv2_2012} using DeepLabV3+~\cite{chen_deeplabv3_2018}.}
    The best score per metric is highlighted \textbf{for each MTO method} as well as \underline{over all methods and optimizers}.
    We note a full dominance of Adam over SGD+mom on both the depth and normal tasks as well as on the $\Delta_m$ metric.
    Overall, best results for all metrics were also achieved using Adam as optimizer.
    }
    \begin{tabular}{cccc@{\gcs}ccc@{\gcs}ccc@{\gcs}cccccc@{\gcs}c}
    \toprule    
        &&&& \multicolumn{2}{c}{Sem.Seg.} && \multicolumn{2}{c}{Depth}  && \multicolumn{5}{c}{Normal}\\
        \cmidrule(r){5-6} \cmidrule(r){8-9} \cmidrule(r){11-15}
        MTO & Optimizer & lr && mIoU$\uparrow$ & pixAcc$\uparrow$  && AbsErr$\downarrow$  & RelErr$\downarrow$ && Mean$\downarrow$ & Median$\downarrow$ & $<11.25\uparrow$ & $<22.5\uparrow$ & $<30.0\uparrow$ && DeltaM$\downarrow$\\
    \midrule
        STL & adam &  &&       0.552 & 0.767 && 0.365 & 0.152 && 21.16 & 14.52 & 0.402 & 0.668 & 0.768\\ 
    \midrule
        EW & adam & 0.0001 &&        0.552 & \textbf{0.770} && \textbf{\underline{0.352}} & \textbf{0.143} && \textbf{22.54} & \textbf{16.05} & \textbf{0.366} & \textbf{0.634} & \textbf{0.742} && \textbf{2.70} \std{0.12}\\ 
        EW & signSGD & 0.0001    && 0.554 & 0.766 && 0.362 & 0.147 && 22.69 & 16.15 & 0.364 & 0.630 & 0.739 && 3.60 \std{0.38}\\
        EW & sgd & 0.01 &&           \textbf{0.556} & 0.769 && 0.365 & 0.147 && 23.35 & 16.76 & 0.352 & 0.617 & 0.727 && 5.21 \std{0.18}\\ \hdashline
        UW & adam & 0.0005 &&        0.532 & 0.755 && \textbf{0.358} & \textbf{0.141} && \textbf{22.01} & \textbf{15.30} & \textbf{0.383} & \textbf{0.650} & \textbf{0.753} && \textbf{1.59} \std{0.14}\\ 
        UW & signSGD & 0.0005    && 0.526 & 0.748 && 0.366 & 0.146 && 22.18 & 15.46 & 0.380 & 0.645 & 0.749 && 2.90 \std{0.79}\\
        UW & sgd & 0.01 &&           \textbf{0.557} & \textbf{0.769} && 0.364 & 0.148 && 22.37 & 15.69 & 0.374 & 0.641 & 0.746 && 2.57 \std{0.56}\\ \hdashline
        RLW & adam & 0.0001 &&       0.556 & \textbf{0.768} && \textbf{0.360} & 0.150 && \textbf{22.37} & 15.87 & 0.370 & \textbf{0.638} & \textbf{0.745} && 2.96 \std{0.55}\\ 
        RLW & signSGD & 0.0001    && \textbf{0.557} & \textbf{0.768} && 0.363 & \textbf{0.146} && 22.40 & \textbf{15.84} & \textbf{0.371} & \textbf{0.638} & \textbf{0.745} && \textbf{2.74} \std{0.30}\\
        RLW & sgd & 0.05 &&          0.541 & 0.762 && 0.368 & 0.151 && 23.00 & 16.44 & 0.357 & 0.625 & 0.734 && 5.15 \std{0.54}\\ \hdashline
        IMTL & adam & 0.0005 &&      0.532 & 0.756 && \textbf{0.353} & \textbf{\underline{0.140}} && \textbf{21.34} & \textbf{14.61} & \textbf{0.399} & \textbf{0.666} & \textbf{0.766} && \textbf{\underline{-0.39}} \std{0.05}\\ 
        IMTL & signSGD & 0.0001    && \textbf{0.552} & \textbf{0.767} && 0.359 & 0.146 && 21.81 & 15.22 & 0.385 & 0.652 & 0.756 && 1.12 \std{0.16}\\
        IMTL & sgd & 0.05 &&         0.545 & 0.762 && 0.367 & 0.150 && 21.63 & 14.98 & 0.390 & 0.658 & 0.760 && 1.30 \std{0.40}\\ \hdashline
        PCGrad & adam & 0.0001 &&    \textbf{\underline{0.558}} & \textbf{\underline{0.772}} && \textbf{0.357} & \textbf{0.146} && \textbf{22.33} & \textbf{15.81} & \textbf{0.371} & \textbf{0.639} & \textbf{0.746} && \textbf{2.34} \std{0.48}\\ 
        PCGrad & signSGD & 0.0001    && 0.553 & 0.766 && 0.360 & 0.148 && 22.61 & 16.11 & 0.365 & 0.631 & 0.739 && 3.55 \std{0.13}\\
        PCGrad & sgd & 0.01 &&       0.556 & 0.769 && 0.361 & 0.147 && 23.36 & 16.79 & 0.351 & 0.617 & 0.728 && 5.10 \std{0.28}\\ \hdashline
        CAGrad & adam & 0.0005 &&    0.531 & 0.755 && \textbf{0.357} & \textbf{0.143} && \textbf{21.42} & \textbf{14.67} & \textbf{0.399} & \textbf{0.664} & \textbf{0.764} && \textbf{0.10} \std{0.55}\\ 
        CAGrad & signSGD & 0.0001    && 0.550 & 0.766 && 0.361 & 0.146 && 21.94 & 15.27 & 0.383 & 0.650 & 0.754 && 1.49 \std{0.12}\\
        CAGrad & sgd & 0.05 &&       \textbf{0.556} & \textbf{0.768} && 0.360 & 0.144 && 21.77 & 14.99 & 0.391 & 0.657 & 0.759 && 0.42 \std{0.07}\\   \hdashline
        AlignedMTL & adam & 0.0005 && 0.531 & 0.752 && \textbf{0.355} & \textbf{0.144} && \textbf{\underline{21.15}} & \textbf{\underline{14.49}} & \textbf{\underline{0.403}} & \textbf{\underline{0.669}} & \textbf{\underline{0.768}} && \textbf{-0.33} \std{0.41}\\
        AlignedMTL & signSGD & 0.0005 && 0.527 & 0.751 && 0.360 & \textbf{0.144} && 21.35 & 14.60 & 0.400 & 0.666 & 0.765 && 0.25 \std{0.21}\\
        AlignedMTL & sgd & 0.1 && \textbf{0.550} & \textbf{0.766} && 0.367 & 0.150 && 21.59 & 14.88 & 0.393 & 0.659 & 0.761 && 0.95 \std{0.62}
        \\  \hdashline
        adatask & adam & 0.0001 &&       0.553 & \textbf{0.768} && \textbf{0.353}& \textbf{0.142} && \textbf{21.44} & \textbf{14.83} & \textbf{0.395} & \textbf{0.661} & \textbf{0.763} &&\textbf{\underline{-0.39}} \std{0.21} \\
        MTL-IO & sgd & 0.01 &&          \textbf{0.554} & 0.768 && 0.362 & 0.146 && 23.31 & 16.75 & 0.352 & 0.617 & 0.728 && 5.01 \std{0.38} \\
    \bottomrule
    \end{tabular}
\label{tab:res_nyu_deeplab}
\end{table}

\clearpage

\section{Additional gradient alignment results}
\label{sec:additional_gradient}

We report extended evaluation on gradient simlarity in MTL and STL.

\begin{figure}
     \centering
     \begin{subfigure}[t]{0.49\textwidth}
         \centering
         \includegraphics[trim = 0mm 10mm 35mm 20mm, clip, width=\textwidth]{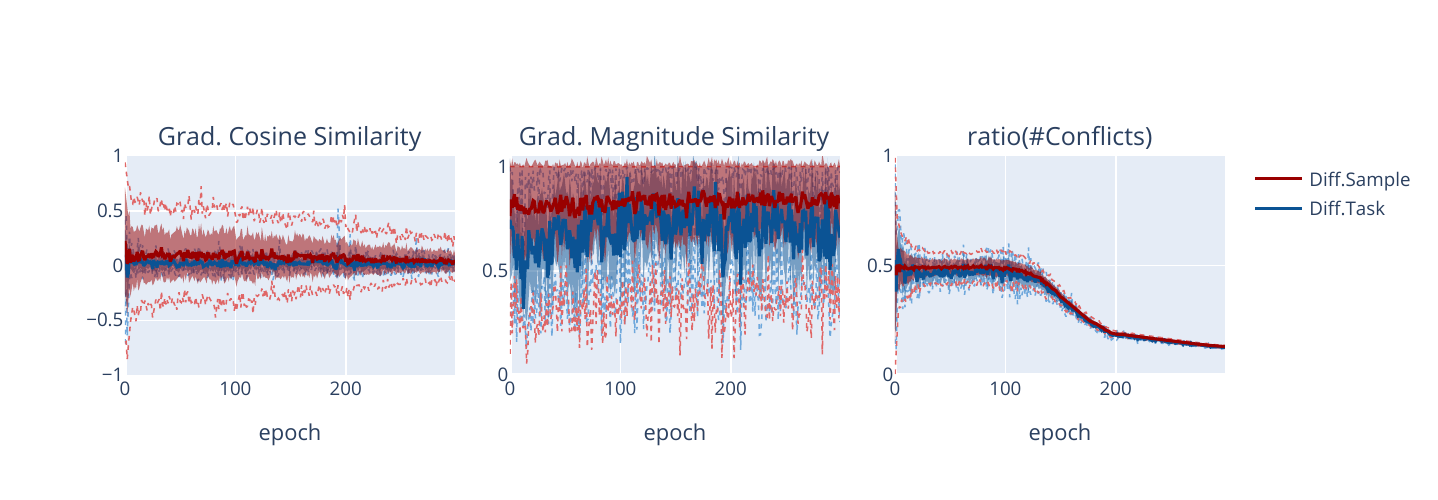}
         \caption{Cityscapes, \\ SegNet}
     \end{subfigure}
     \hfill
     \begin{subfigure}[t]{0.49\textwidth}
         \centering
         \includegraphics[trim = 0mm 10mm 35mm 20mm, clip, width=\textwidth]{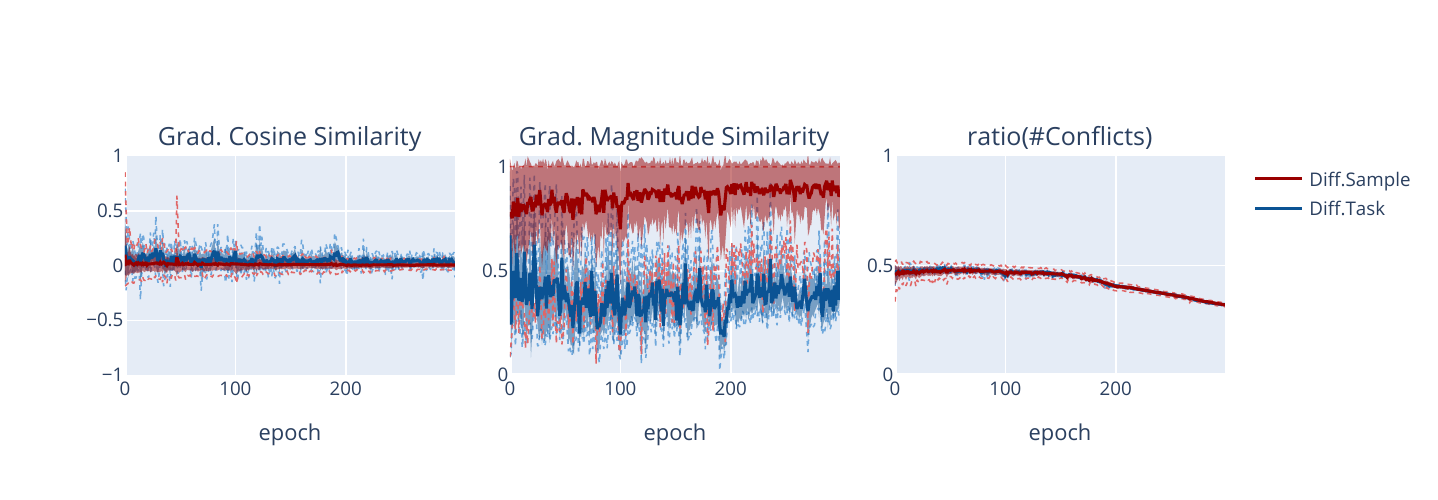}
         \caption{Cityscapes, \\ \hspace{14pt}DeepLabV3}
     \end{subfigure}
     \hfill
     \\
    \begin{subfigure}[t]{0.49\textwidth}
         \centering
         \includegraphics[trim = 0mm 10mm 35mm 20mm, clip, width=\textwidth]{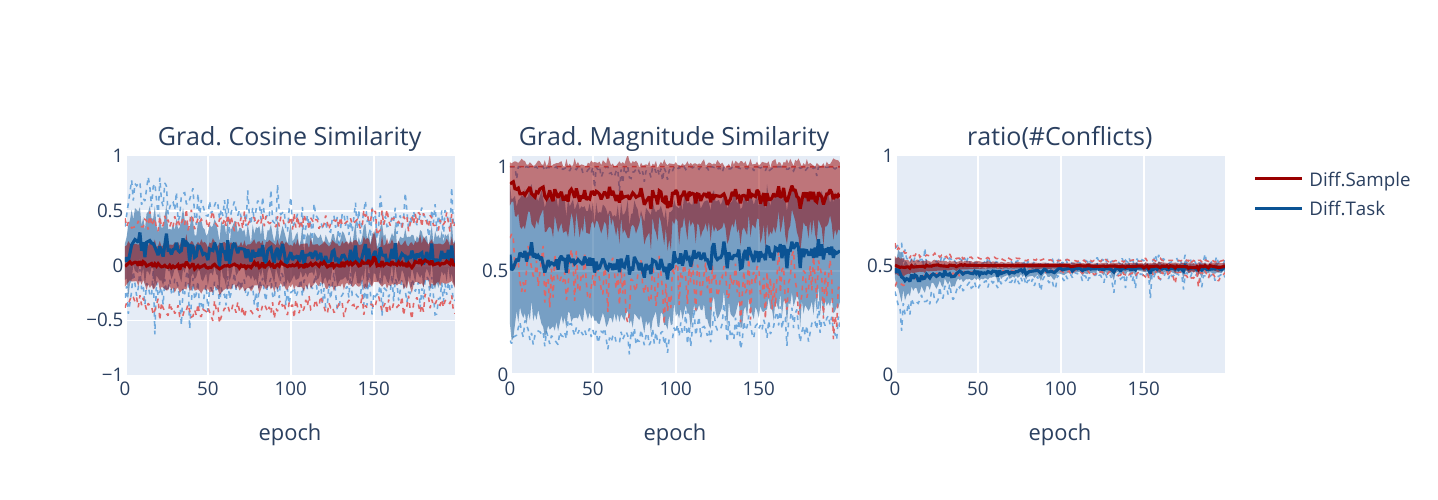}
         \caption{NYUv2, \\ \hspace{11pt}SegNet}
     \end{subfigure}
     \hfill
     \begin{subfigure}[t]{0.49\textwidth}
         \centering
         \includegraphics[trim = 0mm 10mm 35mm 20mm, clip, width=\textwidth]{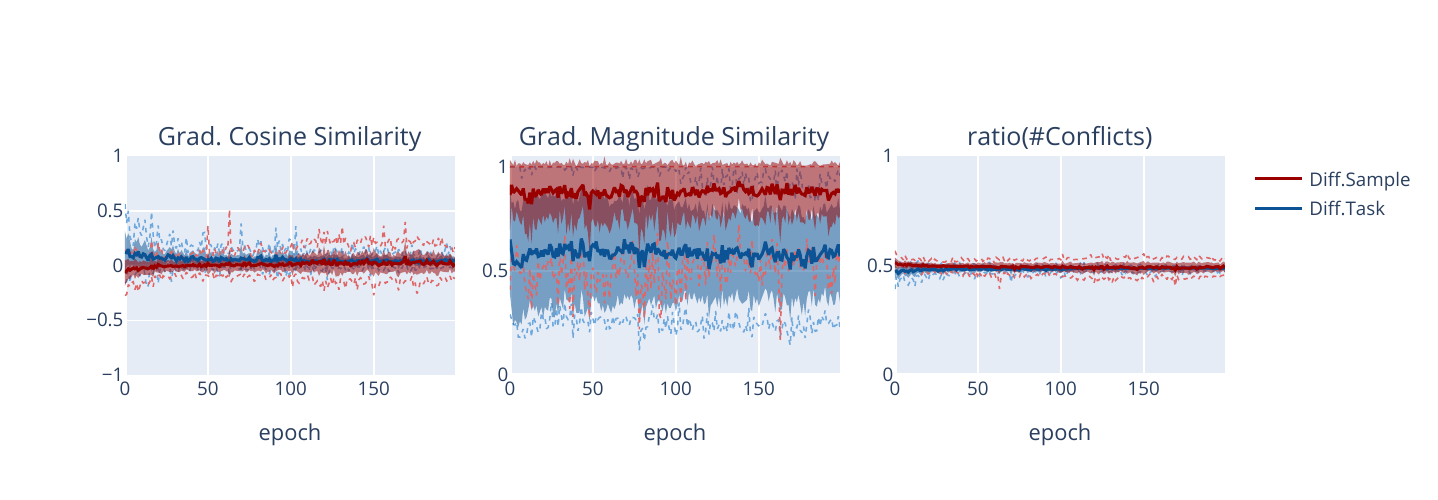}
         \caption{NYUv2, \\ \hspace{25pt}DeepLabV3}
     \end{subfigure}
     \hfill
    \caption{
    \textbf{Gradient similarities when averaging over batch/ losses.}
    In contrast to results shown in \cref{fig:gradsim}, we compute either the average gradient over all tasks when comparing {\color{red} inter-samples} or the average gradient over all samples within the batch for the comparison between {\color{blue} inter-tasks}.    
     We observed a lower variance in some cases (e.g. Cityscapes+Segnet, grad. cosine similarity) which we trace back on noisy gradients being averaged out.
     Overall, we obtain the same findings as for a direct gradient comparison.
    }
    \label{fig:gradsim_batch}
\end{figure}

\begin{figure}[ht]
     \centering
     \begin{subfigure}[t]{0.023\textwidth}
         \includegraphics[trim = 0mm 10mm 235mm 15mm, clip, width=\textwidth]{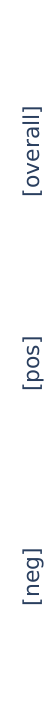}
     \end{subfigure}
     \begin{subfigure}[t]{0.185\textwidth}
         \includegraphics[trim = 10mm 10mm 165mm 15mm, clip, width=\textwidth]{imgs/gradsim/gradsim__combi_posneg__CityScapes+aug__EW__HPS-SegNet__adam-lr0.001-wd1e-05__bs-64__seed-0.pdf}
         \caption{Cityscapes, \\ SegNet}
     \end{subfigure}
     \hfill
     \begin{subfigure}[t]{0.185\textwidth}
         \centering
         \includegraphics[trim = 10mm 10mm 165mm 15mm, clip, width=\textwidth]{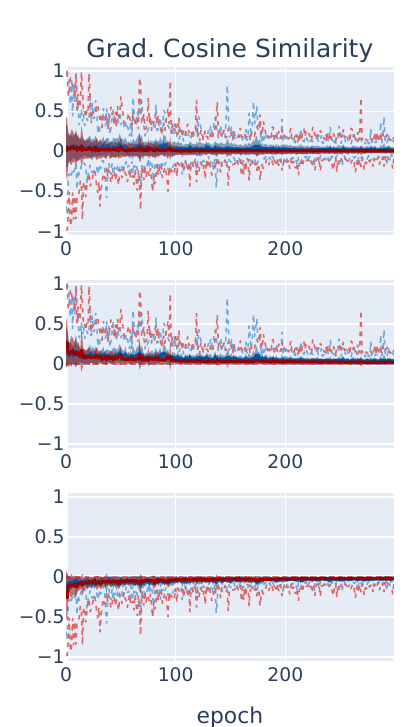}
         \caption{Cityscapes, \\ DeepLabV3}
     \end{subfigure}
     \hfill
    \begin{subfigure}[t]{0.185\textwidth}
         \centering
         \includegraphics[trim = 10mm 10mm 165mm 15mm, clip, width=\textwidth]{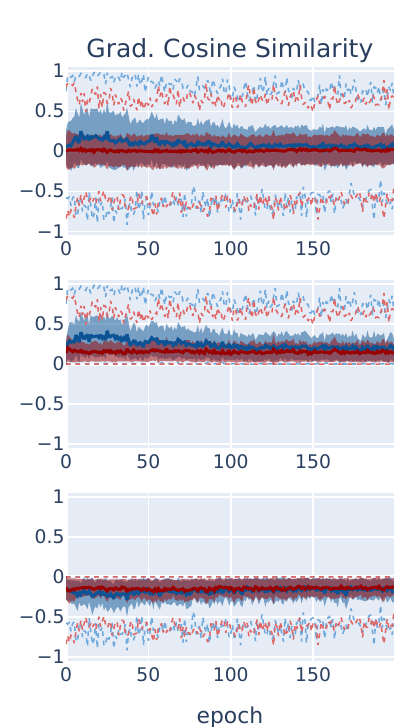}
         \caption{NYUv2, \\ SegNet}
     \end{subfigure}
     \hfill
     \begin{subfigure}[t]{0.185\textwidth}
         \centering
         \includegraphics[trim = 10mm 10mm 165mm 15mm, clip, width=\textwidth]{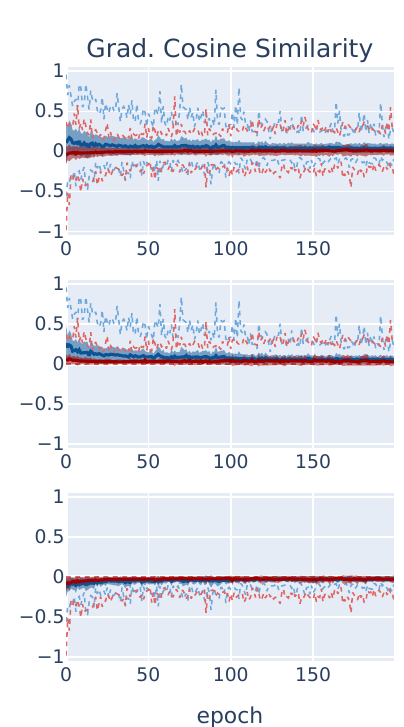}
         \caption{NYUv2, \\ DeepLabV3}
     \end{subfigure}
     \hfill
    \begin{subfigure}[t]{0.185\textwidth}
         \centering
         \includegraphics[trim = 10mm 10mm 165mm 15mm, clip, width=\textwidth]{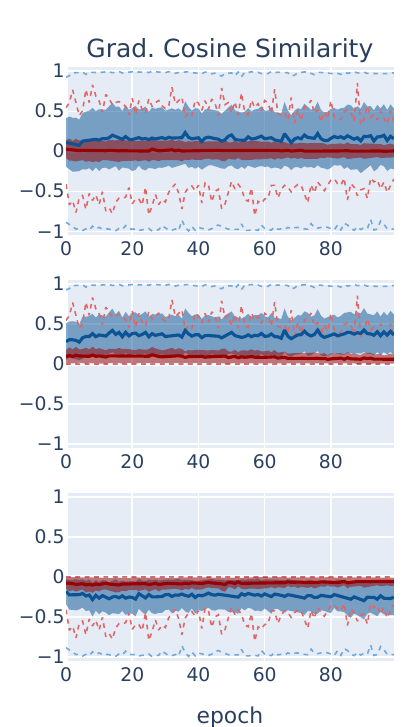}
         \caption{CelebA, \\ ResNet50}
     \end{subfigure}
     \hfill
    \caption{\textbf{Differentiation between conflicting and supportive gradients.}
    We report mean (solid line), standard deviation (shaded area), upper ($97.5\%$) and lower ($2.5\%$) percentile (dotted line) of the gradient cosine similarity between either {\color{red} inter-samples} gradients or {\color{blue} inter-tasks} gradients within an epoch.
    While showing overall results over all respective gradient pairs (Top) as can be also found in \Cref{fig:gradsim}, we also show the course of cosine similarity for either gradients that are conflicting ([neg], bottom) or those which have cosine similarity greater than zero ([pos], middle).
    }
    \label{fig:gradsim_posneg}
\end{figure}

\begin{figure}
     \centering
     \begin{subfigure}[t]{0.19\textwidth}
         \centering
         \includegraphics[trim = 13mm 9mm 35mm 20mm, clip, width=\textwidth]{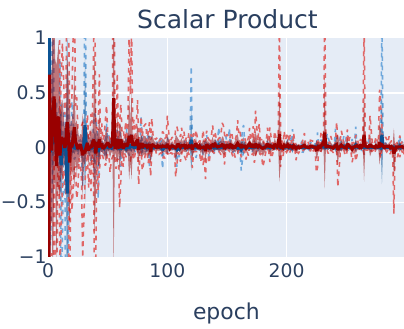}
         \caption{Cityscapes, \\ SegNet}
     \end{subfigure}
     \hfill
     \begin{subfigure}[t]{0.19\textwidth}
         \centering
         \includegraphics[trim = 13mm 9mm 35mm 20mm, clip, width=\textwidth]{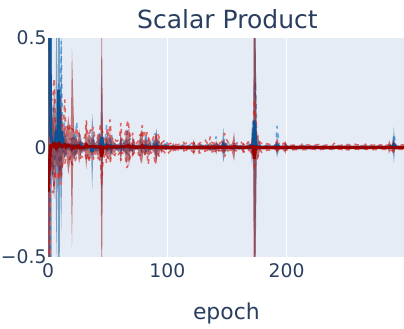}
         \caption{Cityscapes, \\ DeepLabV3}
     \end{subfigure}
     \hfill
    \begin{subfigure}[t]{0.19\textwidth}
         \centering
         \includegraphics[trim = 13mm 9mm 35mm 20mm, clip, width=\textwidth]{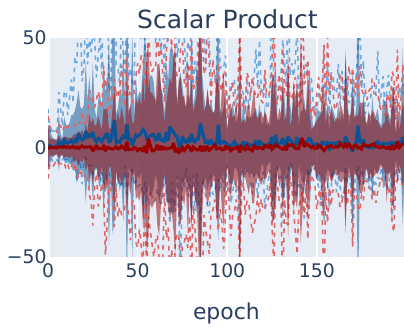}
         \caption{NYUv2, \\ SegNet}
     \end{subfigure}
     \hfill
     \begin{subfigure}[t]{0.19\textwidth}
         \centering
         \includegraphics[trim = 13mm 9mm 35mm 20mm, clip, width=\textwidth]{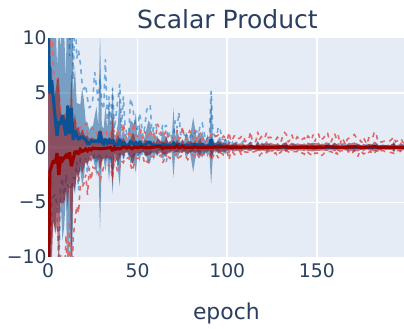}
         \caption{NYUv2, \\ DeepLabV3}
     \end{subfigure}
     \hfill
    \begin{subfigure}[t]{0.19\textwidth}
         \centering
         \includegraphics[trim = 13mm 9mm 35mm 20mm, clip, width=\textwidth]{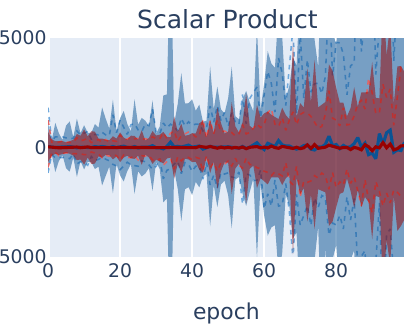}
         \caption{CelebA, \\ ResNet50}
     \end{subfigure}
     \hfill
    \caption{\textbf{Scalar product between pairs of gradients.}
    We report mean (solid line), standard deviation (shaded area), upper ($97.5\%$) and lower ($2.5\%$) percentile (dotted line) of the gradient cosine similarity between either gradients of {\color{red} inter-samples} or {\color{blue} inter-tasks} within an epoch.
    We observe an overall decrease of the variance of the scalar product for both Cityscapes setups and the NYUv2+DeepLabV3 experiment over the training which we explain with evenly smaller overall gradients.
    Surprisingly, this does not apply for NYUv2 with SegNet or CelebA.
    Similar to previous results, we do not see any indication for inter-sample gradients being better aligned than inter-tasks gradients.
    }
    \label{fig:gradsim_scalarproduct}
\end{figure}

Alternatively to a single sample/ task, we consider the average gradient in \cref{fig:gradsim_batch}.
In \cref{fig:gradsim_posneg}, we differentiate between conflicting and supporting gradient pairs when evaluating the cosine similarity.
\Cref{fig:gradsim_scalarproduct} shows the evaluation of the scalar product as an combined measure of similarity in gradient direction and magnitude.

\clearpage

\section{Robustness of multi-task representations on corrupted data}
\label{sec:app_ood}

In this additonal part of our analysis, we investigate whether features learned for multiple tasks generalize better to corrupted data compared to those learned for single tasks only.

\textit{Motivation:} 
In his seminal paper, Caruana gives preliminary evidence that MTL provides stronger features and avoids spurious correlations (referred to better \emph{attribute selection})~\cite{caruana1997multitask}. More recently, spurious correlations have often been directly connected with robustness \cite{geirhos2020shortcut, ilyas2019adversarial}. 
Results from current literature on the robustness of MTL features are mixed. While MTL is stated to increase the adversarial- and noise-robustness over STL\cite{klingner2020improved,mao2020multitask,yeo2021robustness}, others argue features selected by MTL could be more likely to be non-causal and, therefore, less robust~\cite{hu2022improving,beery2018recognition}.
Here, we further examine whether MTL features lead to better robustness. 
We would like to nuance that we do not consider the transferability of representations, e.g., to new tasks, but solely focus on the claim that the MTL trained features are more robust w.r.t.\ different inputs. 

\textit{Approach:} 
In our experiment, we treat the common corruptions~\cite{hendrycks_robustness_2019} as downstream task and compare the performance after fine-tuning the heads on corrupted data while freezing the pre-trained STL/MTL backbone.
While this differs from the typical OOD setup, here, it allows us to explore whether MTL or STL yield more robust representation for corrupted data.

We select models trained on clean data with the best performing hyperparameter configuration from previous experiments and fine-tune their heads on corrupted data.
Following this, we compare the test performance of models trained in the multi-task setup to those that were learned for a single-task only.
We use the perturbation modes proposed by Hendrycks et al.~\cite{hendrycks_robustness_2019} which include different variants of noise, blur, and weather conditions and apply five levels of severity.
We randomly select corruption and severity level for each data sample during fine-tuning and create a full corrupted version of the test data 
considering all proposed corruptions and perturbation levels. 

To quantify the robustness of single- and multi-task models, we first compute the individual task metrics $M$ (e.g., mIoU) per task $t$ for a STL and MTL network. Next, we compute the relative performance when each model is faced with corrupted data. Lastly, we calculate the difference of relative performances of the MTL compared to the STL model. 
In detail, over all corruption modes $C$ and levels of severity $S$ we have 
\begin{align}
    \label{eq:ood_eval}
    \delta_t &= \frac{1}{|C|\cdot |S|}\sum_{c\in C}\sum_{s\in S}(-1)^{p(t)}\delta_{t,c,s} \\ \nonumber
    & \text{with} \quad \delta_{t,c,s} = \frac{M^{MTL, corrupted}_{t,c,s}}{M^{MTL, clean}_{t}} - \frac{M^{STL, corrupted}_{t,c,s}}{M^{STL, clean}_{t}}
\end{align}
where $p(t)=1$ if a higher value on task $t$ corresponds to better performance and $p(t)=0$ otherwise.
This metric yields $\delta_t<0$ if the MTL model was able to handle data corruption better. If the STL model is less impacted, we get $\delta_t>0$.

\begin{figure*}[!ht]
     \centering
     \begin{subfigure}[b]{0.49\textwidth}
         \centering
         \includegraphics[trim = 0mm 0mm 0mm 0mm, clip, width=\textwidth]{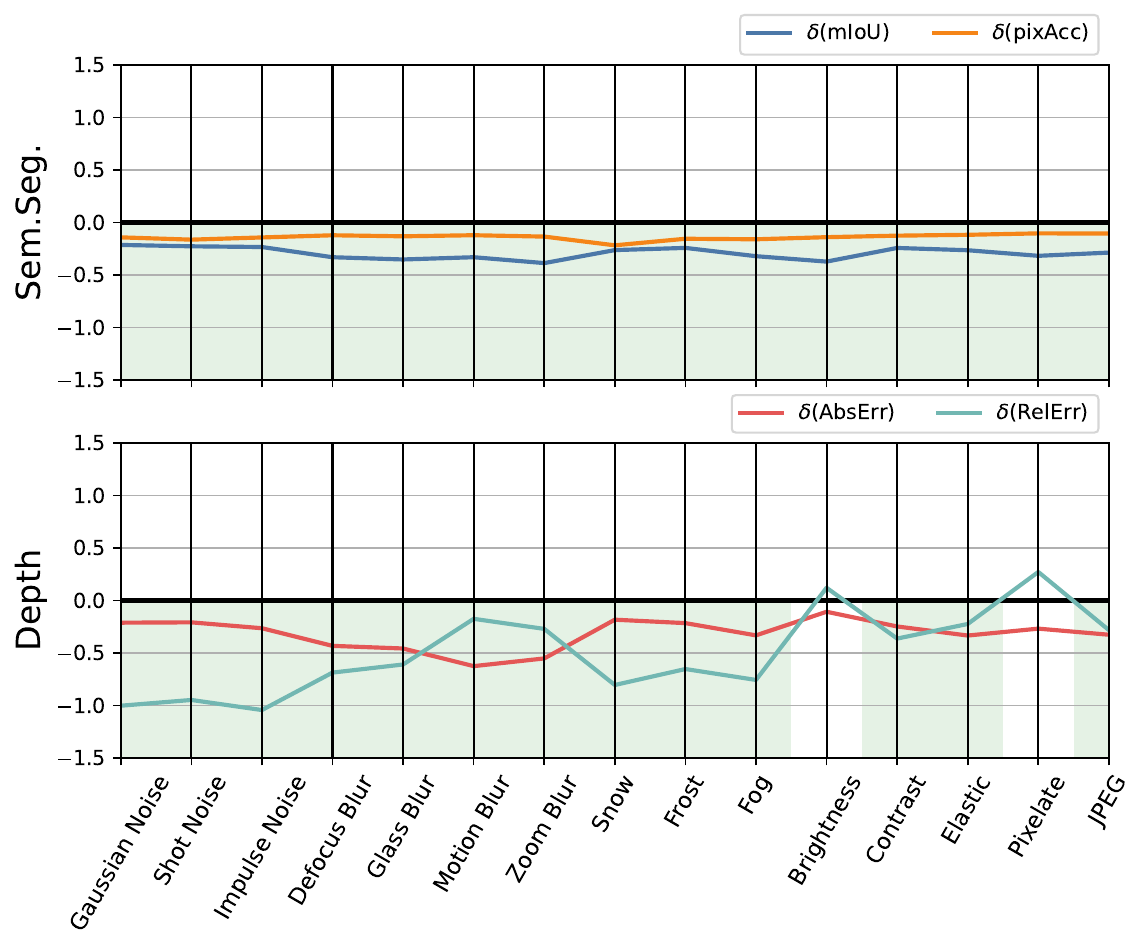}
         \caption{Cityscapes, SegNet}
     \end{subfigure}
     \hfill
    \begin{subfigure}[b]{0.49\textwidth}
         \centering
         \includegraphics[trim = 0mm 0mm 0mm 0mm, clip, width=\textwidth]{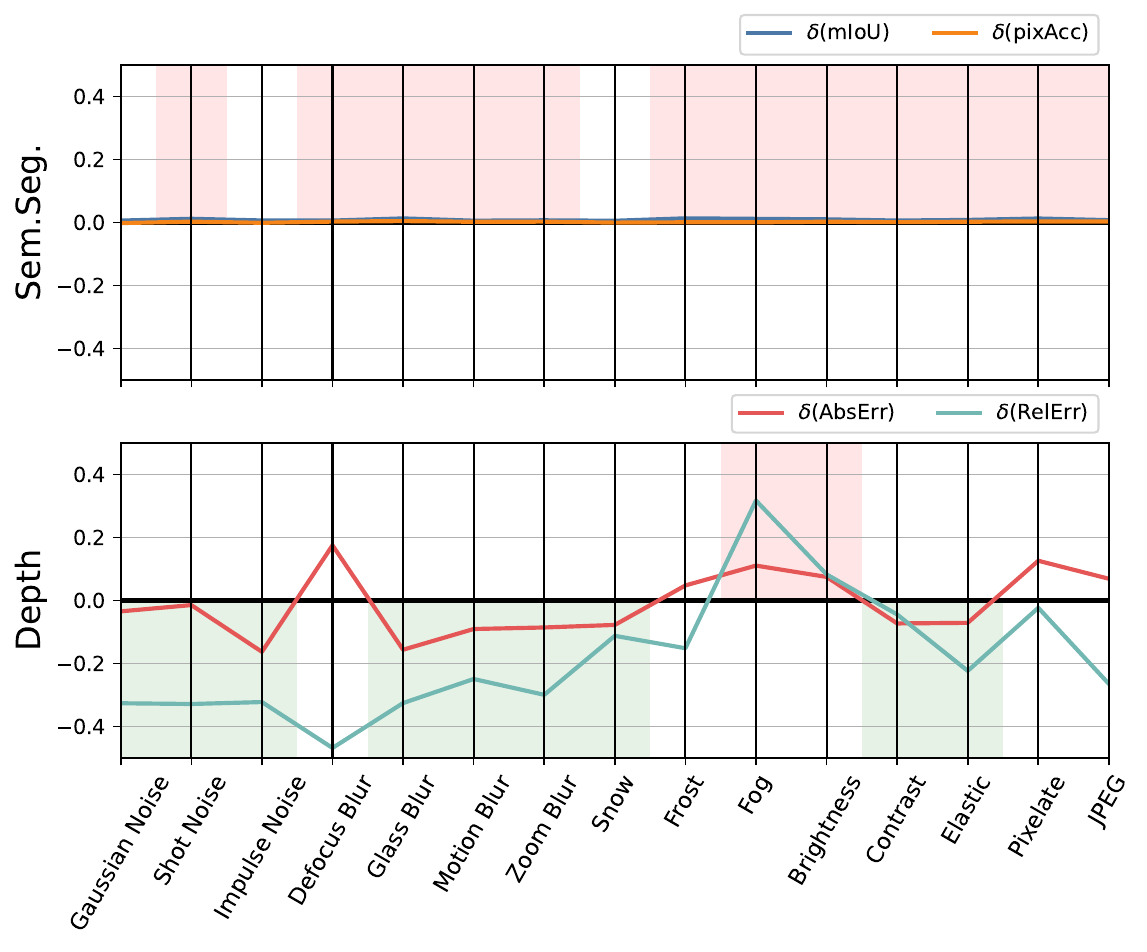}
         \caption{Cityscapes, DeepLabV3}
     \end{subfigure}
     \\
    \begin{subfigure}[b]{0.49\textwidth}
         \centering
         \includegraphics[trim = 0mm 0mm 0mm 0mm, clip, width=\textwidth]{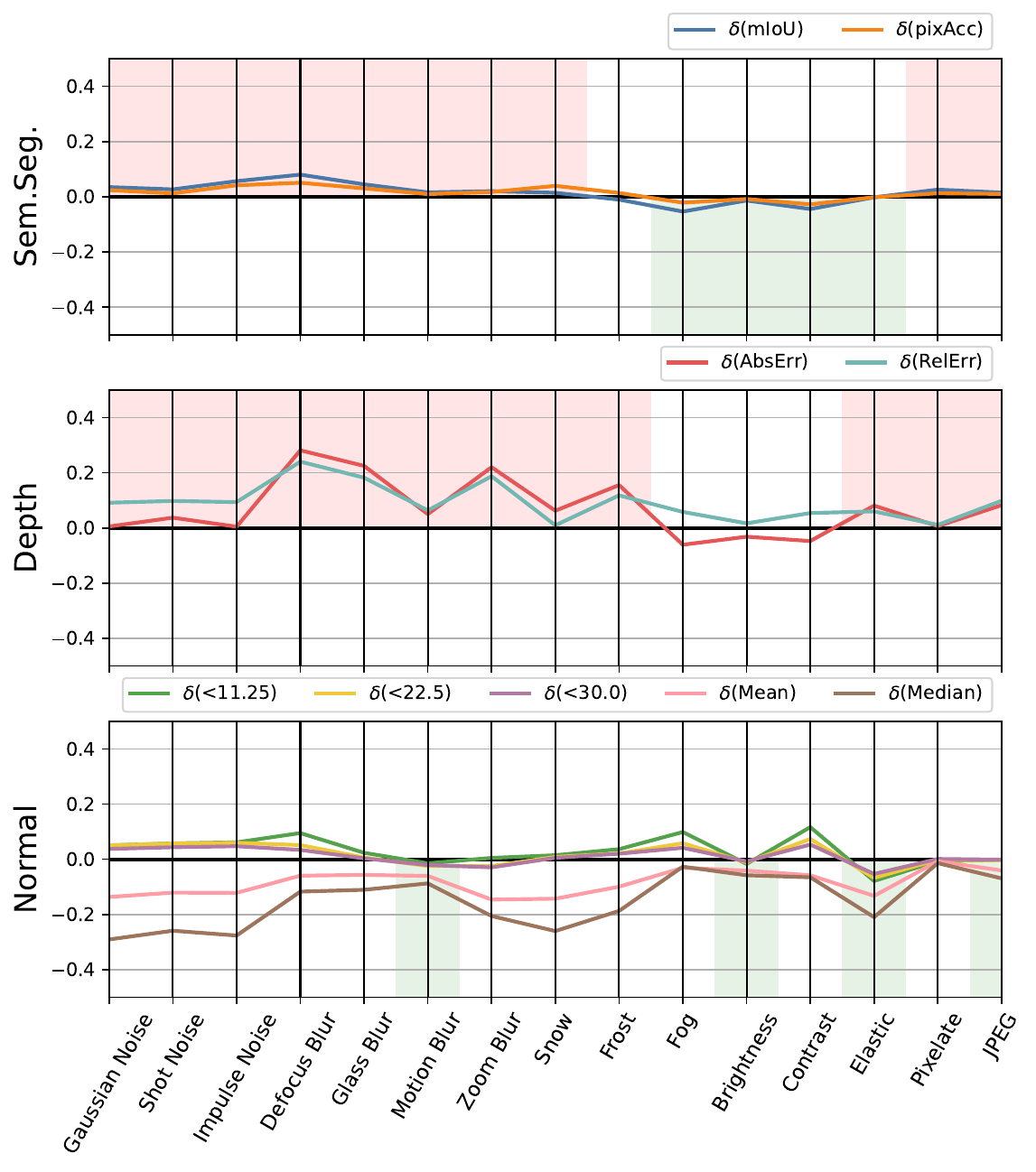}
         \caption{NYUv2, SegNet}
     \end{subfigure}
    \hfill
    \begin{subfigure}[b]{0.49\textwidth}
         \centering
         \includegraphics[trim = 0mm 0mm 0mm 0mm, clip, width=\textwidth]{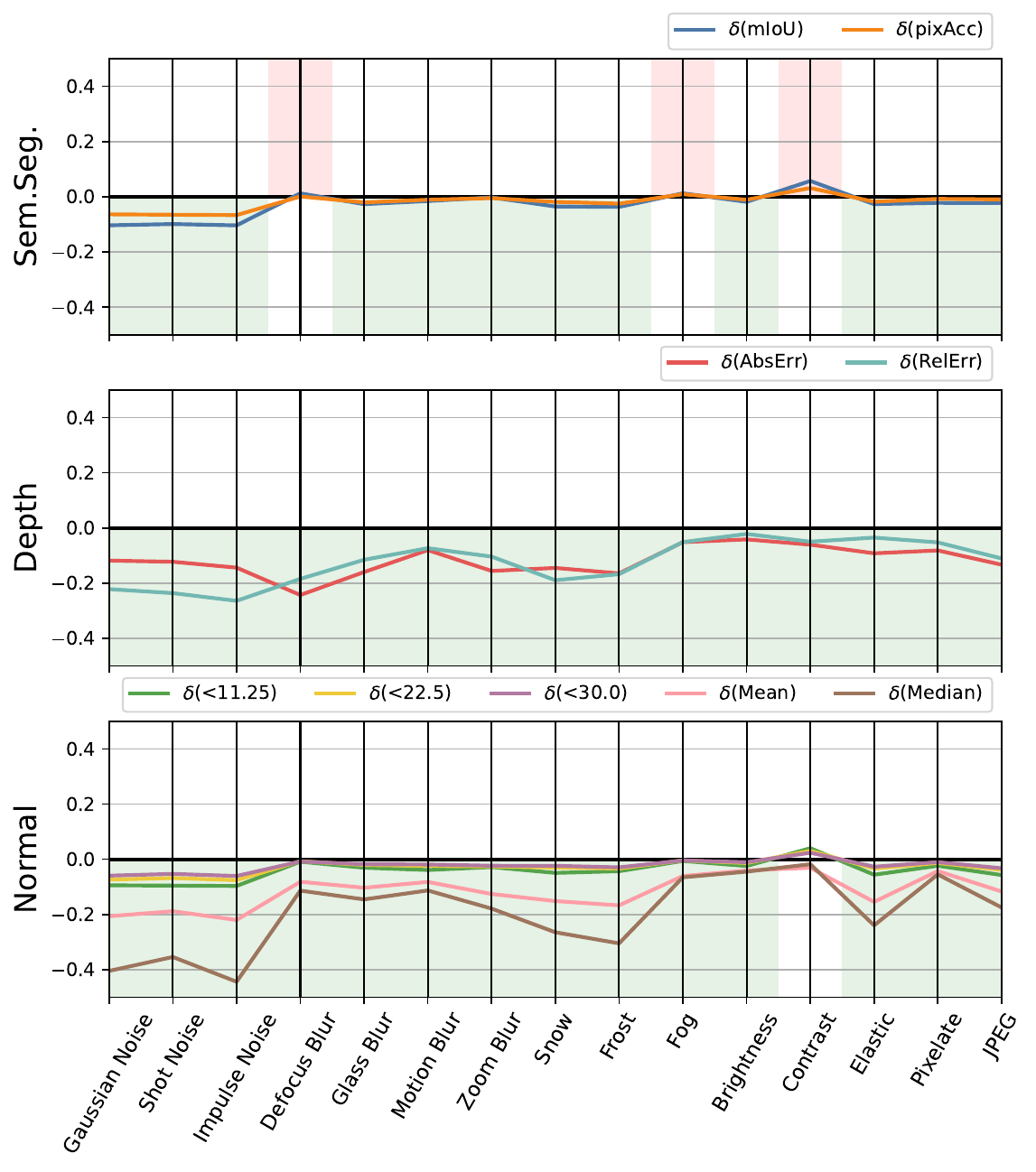}
         \caption{NYUv2, DeepLabV3}
    \end{subfigure}
    \caption{\textbf{Transfer to out-of-distribution data for MTL and STL.} 
    For every task and respective metrics, we show the difference over relative performance decrease over all corruption modes averaged over five levels of severity and three runs.
    EW was used to train the MTL model on uncorrupted data.
    We color blocks in case either {\color{red}{STL}} or {\color{ForestGreen}{MTL}} is able to handle the respective corruption better for all metrics of one task.
    Regarding the Cityscapes dataset, the performance on both task would strongly benefit in MTL setup.
    A similar behavior can be seen for NYUv2+DeepLabV3.
    Using SegNet on NYUv2, however, shows preferences towards STL features.
    Overall, we see a minor indication that MTL result in features that would generalize better to corrupted data.
    }
    \label{fig:ood_more}
\end{figure*}

\begin{table*}[ht]
    \centering
    \scriptsize
    \setlength{\tabcolsep}{4pt}  
    \newcommand{\gcs}{\hspace{12pt}}
    \newcommand{\customdashline}{\cdashline{2-10}\noalign{\vskip 0.5ex}}
    \caption{
        \textbf{Out-Of Distribution transfer on corrupted Cityscapes~\cite{cordts_cityscapes_2016} dataset for different networks and MTO methods.}
        We report difference between relative performance decrease for single-task and multi-task learning averaged over all modes of corruption and all levels of severity (cf. \Cref{eq:ood_eval}).
        A value lower than zero indicates a better generalization capability of the MTL backbone, a positive value displays that the STL backbone shows a lower decrease when evaluated on the corrupted data.
        Results are averaged over runs for three seeds for both multi-task and single-task models.
        Overall, we observe a slight benefit in performance for the depth task when training for multiple tasks.
    }
    \begin{tabular}{llc@{\gcs}ccc@{\gcs}ccc@{\gcs}c}
    \toprule
        &&& \multicolumn{2}{c}{Sem.Seg.} && \multicolumn{2}{c}{Depth}  && Mean\\
        \cmidrule(){4-5}\cmidrule(){7-8}\cmidrule(){10-10}
        Network & MTO  && $\delta_{\text{mIoU}}$  & $\delta_{\text{pixAcc}}$   && $\delta_{\text{AbsErr}}$  & $\delta_{\text{RelErr}}$  &&  \\
        \midrule
        SegNet & EW && -0.2922 & -0.1386 && -0.3175 & -0.4949 && -0.3108 \\
        \customdashline
        ~ & UW && 0.0854 & 0.0421 && 0.7799 & -0.2118 && 0.1739 \\
        ~ & RLW && 0.0699 & 0.0307 && 0.6553 & -0.5652 && 0.0477 \\
        ~ & IMTL && 0.0696 & 0.0243 && 0.5384 & 0.1108 && 0.1858 \\
        ~ & PCGrad && 0.0824 & 0.0316 && 0.4106 & -0.6392 && -0.0286 \\
        ~ & CAGrad && 0.0561 & 0.0213 && 0.3860 & -0.3100 && 0.0383 \\
        \midrule
        DeepLabV3 & EW && 0.0087 & 0.0017 && -0.0108 & -0.1826 && -0.0457 \\
        \customdashline
        ~ & UW && 0.0110 & 0.0018 && 0.1179 & 0.0861 && 0.0542 \\
        ~ & RLW && 0.0263 & 0.0084 && -0.0809 & -0.3595 && -0.1015 \\
        ~ & IMTL && 0.0172 & 0.0041 && 0.1747 & 0.0857 && 0.0704 \\
        ~ & PCGrad && 0.0090 & 0.0019 && 0.0394 & -0.2240 && -0.0434 \\
        ~ & CAGrad && 0.0224 & 0.0067 && 0.0898 & -0.2841 && -0.0413 \\
    \bottomrule
    \end{tabular}
    \label{tab:ood_cityscapes}
\end{table*}
\begin{table*}[ht]
    \centering
    \tiny
    \setlength{\tabcolsep}{1pt}  
    \newcommand{\gcs}{\hspace{4pt}}
    \newcommand{\customdashline}{\cdashline{2-16}\noalign{\vskip 0.5ex}}
    \caption{
        \textbf{Out-Of Distribution transfer on corrupted NYUv2~\cite{silberman_nyuv2_2012} dataset for different multi-task optimization methods.}
        We report difference between relative performance decrease for single-task and multi-task learning averaged over all modes of corruption and all levels of severity (cf. \Cref{eq:ood_eval}).
        A value lower than zero indicates a better generalization capability of the MTL backbone, a positive value displays that the STL backbone shows a lower decrease when evaluated on the corrupted data.
        Results are averaged over runs for three seeds for both multi-task and single-task models.
        While for DeepLabV3 largely benefits from MTL, this is not the case for SegNet.
        Over both networks, EW profits shows lowest relative performance decrease among all MTO methods.
        Interestingly we found that even for different metrics corresponding to the same task, either the multi-task or single-task learning model would show lower decrease in performance on the corrupted data.
    }
    \begin{tabular}{llc@{\gcs}ccc@{\gcs}ccc@{\gcs}cccccc@{\gcs}c}
    \toprule
        &&& \multicolumn{2}{c}{Sem.Seg.} && \multicolumn{2}{c}{Depth} && \multicolumn{5}{c}{Normal} && Mean\\
        \cmidrule(){4-5}\cmidrule(){7-8}\cmidrule(){10-14}\cmidrule(){16-16}
        Network & MTO  && $\delta_{\text{mIoU}}$  & $\delta_{\text{pixAcc}}$   && $\delta_{\text{AbsErr}}$  & $\delta_{\text{RelErr}}$  && $\delta_{\text{Mean}}$  & $\delta_{\text{Median}}$  & $\delta_{<11.25}$ & $\delta_{<22.5}$ & $\delta_{<30.0}$ &&  \\
        \midrule
        SegNet & EW && 0.0139 & 0.0134 && 0.0715 & 0.0922 && -0.0836 & -0.1492 & 0.0293 & 0.0168 & 0.0116 && 0.0018 \\
        \customdashline
 ~ & UW && 0.0166 & 0.0149 && 0.1146 & 0.1225 && -0.0564 & -0.0984 & 0.0313 & 0.0207 & 0.0164 && 0.0204 \\
 ~ & RLW && 0.0174 & 0.0223 && -0.0062 & 0.0281 && -0.1048 & -0.1857 & 0.0229 & 0.0089 & 0.0055 && -0.0213 \\
 ~ & IMTL && 0.0073 & 0.0123 && 0.1108 & 0.1247 && -0.0186 & -0.0390 & 0.0361 & 0.0274 & 0.0228 && 0.0315 \\
 ~ & PCGrad && 0.0186 & 0.0174 && 0.0720 & 0.0817 && -0.0702 & -0.1271 & 0.0264 & 0.0158 & 0.0121 && 0.0052 \\
 ~ & CAGrad && 0.0061 & 0.0116 && 0.1172 & 0.1428 && -0.0082 & -0.0202 & 0.0465 & 0.0286 & 0.0223 && 0.0385 \\
        \midrule
        DeepLabV3 & EW && -0.0289 & -0.0187 && -0.1194 & -0.1249 && -0.1182 & -0.1947 & -0.0409 & -0.0288 & -0.0238 && -0.0776 \\
        \customdashline
 ~ & UW && -0.0268 & -0.0165 && -0.0263 & -0.0219 && -0.0465 & -0.0701 & -0.0237 & -0.0139 & -0.0113 && -0.0286 \\
 ~ & RLW && -0.0084 & -0.0054 && -0.0075 & -0.0187 && -0.0390 & -0.0597 & -0.0051 & 0.0009 & 0.0005 && -0.0158 \\
 ~ & IMTL && -0.0211 & -0.0167 && 0.0130 & 0.0059 && -0.0153 & -0.0219 & -0.0072 & -0.0045 & -0.0038 && -0.0079 \\
 ~ & PCGrad && -0.0102 & -0.0081 && -0.0242 & -0.0327 && -0.0504 & -0.0800 & -0.0236 & -0.0107 & -0.0076 && -0.0275 \\
 ~ & CAGrad && -0.0139 & -0.0129 && 0.0022 & -0.0123 && -0.0096 & -0.0063 & 0.0023 & 0.0047 & 0.0034 && -0.0047 \\

    \bottomrule
    \end{tabular}
    \label{tab:ood_nyu_segnet_deeplab}
\end{table*}

\begin{table*}[ht]
    \centering
    \scriptsize
    \setlength{\tabcolsep}{4pt}  
    \newcommand{\gcs}{\hspace{12pt}}
    \newcommand{\customdashline}{\cdashline{2-8}\noalign{\vskip 0.5ex}}
    \begin{tabular}{llc@{\gcs}ccc@{\gcs}cc}
    \toprule
        &&& \multicolumn{2}{c}{Sem.Seg.} && \multicolumn{2}{c}{Depth}\\
        \cmidrule(){4-5}\cmidrule(){7-8}
        Network & MTO  && mIoU & pixAcc   && AbsErr  & RelErr \\
        \midrule
        SegNet & STL && 0.3774 & 0.7241 && 0.0592 & 87.06 \\
        \customdashline
         & EW && 0.3783 & 0.7399 && 0.0533 & 95.17 \\
        \customdashline
         & UW && 0.3689 & 0.7273 && 0.0503 & 77.86 \\
         & RLW && 0.3744 & 0.7321 && 0.0526 & 96.26 \\
         & IMTL && 0.3684 & 0.7363 && 0.0479 & 78.92 \\
         & PCGrad && 0.3678 & 0.7310 && 0.0515 & 94.44 \\
         & CAGrad && 0.3769 & 0.7397 && 0.0479 & 92.75 \\
        \midrule
        DeepLabV3 & STL && 0.5732 & 0.8593 && 0.0244 & 98.97 \\
        \customdashline
         & EW && 0.5769 & 0.8619 && 0.0253 & 104.37 \\
        \customdashline
         & UW && 0.5801 & 0.8632 && 0.0240 & 99.67 \\
         & RLW && 0.5634 & 0.8557 && 0.0270 & 103.22 \\
         & IMTL && 0.5769 & 0.8617 && 0.0244 & 101.26 \\
         & PCGrad && 0.5742 & 0.8609 && 0.0263 & 105.46 \\
         & CAGrad && 0.5690 & 0.8576 && 0.0259 & 104.90 \\
    \bottomrule
    \end{tabular}
    \caption{
        \textbf{Results for evaluating on corrupted Cityscapes~\cite{cordts_cityscapes_2016}.}
        Scores are averaged over all corruption modes, level of severity and three seeds.
    }
    \label{tab:ood_cityscapes_abs}
\end{table*}

\begin{table*}[ht]
    \centering
    \tiny
    \setlength{\tabcolsep}{2pt}  
    \newcommand{\gcs}{\hspace{6pt}}
    \newcommand{\customdashline}{\cdashline{2-14}\noalign{\vskip 0.5ex}}
    \begin{tabular}{llc@{\gcs}ccc@{\gcs}ccc@{\gcs}ccccc}
    \toprule
        &&& \multicolumn{2}{c}{Sem.Seg.} && \multicolumn{2}{c}{Depth} && \multicolumn{5}{c}{Normal} \\
        \cmidrule(){4-5}\cmidrule(){7-8}\cmidrule(){10-14}
        Network & MTO  && mIoU & pixAcc   && AbsErr  & RelErr && Mean  & Median  & $<11.25$ & $<22.5$ & $<30.0$ \\
        \midrule
        SegNet & STL && 0.2282 & 0.4783 && 0.8389 & 0.3156 && 32.05 & 26.60 & 0.1994 & 0.4390 & 0.5608 \\
        \customdashline
         & EW && 0.2253 & 0.4750 && 0.7702 & 0.2886 && 36.06 & 32.90 & 0.1306 & 0.3263 & 0.4529 \\
        \customdashline
         & UW && 0.2266 & 0.4750 && 0.7847 & 0.2913 && 34.90 & 31.25 & 0.1445 & 0.3538 & 0.4814 \\
         & RLW && 0.2182 & 0.4587 && 0.7718 & 0.2957 && 37.09 & 34.39 & 0.1165 & 0.3045 & 0.4298 \\
         & IMTL && 0.2190 & 0.4743 && 0.7804 & 0.2885 && 33.89 & 29.67 & 0.1609 & 0.3812 & 0.5085 \\
         & PCGrad && 0.2279 & 0.4751 && 0.7790 & 0.2900 && 35.34 & 31.90 & 0.1388 & 0.3423 & 0.4696 \\
         & CAGrad && 0.2324 & 0.4850 && 0.7822 & 0.2897 && 33.50 & 29.10 & 0.1648 & 0.3908 & 0.5183 \\
        \midrule
        DeepLabV3 & STL && 0.3793 & 0.6285 && 0.5342 & 0.2111 && 27.32 & 21.25 & 0.2853 & 0.5286 & 0.6414 \\
        \customdashline
         & EW && 0.3644 & 0.6132 && 0.5594 & 0.2181 && 29.88 & 24.80 & 0.2359 & 0.4657 & 0.5851 \\
        \customdashline
         & UW && 0.3836 & 0.6321 && 0.5137 & 0.2003 && 27.75 & 21.89 & 0.2738 & 0.5169 & 0.6324 \\
         & RLW && 0.3861 & 0.6338 && 0.5290 & 0.2081 && 28.33 & 23.00 & 0.2566 & 0.4964 & 0.6162 \\
         & IMTL && 0.3773 & 0.6323 && 0.5208 & 0.1991 && 27.27 & 21.34 & 0.2830 & 0.5273 & 0.6412 \\
         & PCGrad && 0.3932 & 0.6409 && 0.5168 & 0.2025 && 28.05 & 22.44 & 0.2659 & 0.5063 & 0.6238 \\
         & CAGrad && 0.3738 & 0.6289 && 0.5266 & 0.2022 && 27.68 & 21.86 & 0.2754 & 0.5177 & 0.6324 \\

    \bottomrule
    \end{tabular}
    \caption{
        \textbf{Results for evaluating on corrupted NYUv2~\cite{silberman_nyuv2_2012}.}
        Scores are averaged over all corruption modes, level of severity and three seeds.
    }
    \label{tab:ood_nyu_abs}
\end{table*}

\textit{Results:} \Cref{fig:ood_more} shows $\delta_t$ for different corruption types for all network architectures and datasets with EW averaged over five corruption levels and three seeds. 
Considering the combination DeepLabV3 and Cityscapes, on the semantic segmentation tasks, the STL models show a slightly lower decrease in performance on the corrupted data than MTL ($\delta_t>0$ more often; shaded in red), indicating that the features learned for these respective tasks can better generalize to corrupted data.
In contrast, the MTL model shows significant better relative performance on the depth task ($\delta_t<0$  more often; shaded green). 
Comparing these observations to other dataset+network combinations, we find a strong robustness of MTL features for both setups of CityScape+SegNet and NYUv2+DeepLabV3 over all tasks. However, evaluation on NYUv2+SegNet resulted on average in better performance on STL features, especially for the depth task.
Furthermore, we see little evidence for general higher robustness against certain types of corruption (e.g., higher robustness against weather conditions) for either MTL or STL across all setups.

The results of other MTO methods (\cref{tab:ood_cityscapes,tab:ood_nyu_segnet_deeplab}) indicate that it depends less on the used method but more on the choice of dataset and network architecture whether some tasks would benefit from MTL for learning more robust features. Averaged absolute scores can be found in \cref{tab:ood_cityscapes_abs,tab:ood_nyu_abs}.

\textit{Conclusion:}
Our experiments show that MTL \textit{can} result in learning more robust features, either for a subset of tasks or even all. However, we could not observe a uniform pattern whether certain tasks consistently benefit from MTL.
Instead, it depends on the task, the type of corruption, the network, and the dataset whether MTL or STL is superior towards corrupted data. Whether there is a general pattern, we leave to further research. 
We further cannot fully confirm the outcome of \cite{klingner2020improved} as only two of our setups have indicated that the segmentation task can be more robust in the MTL setting.
Controversial to the claim of \cite{mao2020multitask}, our evaluation shows that none of the MTL approaches, even IMTL, PCGrad or CAGrad which adjust the gradients, yields consistent values of $\delta_t<0$ which would have shown an advantage of certain MTO methods over STL.


\clearpage

\end{document}